\documentclass[twoside]{article}

\usepackage[accepted]{aistats2022}
\usepackage{graphicx}
\usepackage{subfigure}
\usepackage{amssymb}
\usepackage{amsthm}
\usepackage{amsmath}
\usepackage{mathtools}

\usepackage{algorithm}
\usepackage{algorithmic}

\usepackage{diagbox}

\usepackage{lipsum}

\newtheorem{theorem}{Theorem}
\newtheorem*{theorem*}{Theorem}
\newtheorem{Lemma}{Lemma}

\newtheorem{corollary*}{Corollary}
\newtheorem{property}{Proposition}

\newtheorem*{proposition*}{Proposition}

\DeclareMathOperator*{\argmin}{arg\,min}

\begin{document}

%

%

\twocolumn[

\aistatstitle{Neural Optimization Kernel: Towards Robust Deep Learning }


\aistatsauthor{ Yueming  Lyu \And Ivor Tsang }

\aistatsaddress{ Australian Artificial Intelligence Institute \\University of Technology Sydney \And Australian Artificial Intelligence Institute \\ University of Technology Sydney } ]

\begin{abstract}
Deep neural networks (NN) have achieved great success in many applications.  However, why do deep neural networks obtain good generalization at an over-parameterization regime is still unclear. To better understand deep NN,  we establish the connection between deep NN and a novel kernel family, i.e., Neural Optimization Kernel (NOK).   The architecture of structured approximation of NOK performs monotonic descent updates of implicit regularization problems.    We can implicitly choose the regularization problems by employing different activation functions, e.g., ReLU, max pooling, and soft-thresholding.   We further establish a new generalization bound of our deep structured approximated NOK architecture.   Our  unsupervised  structured approximated NOK block can serve as a simple plug-in of popular backbones for  a good generalization against input noise.
\end{abstract}

\section{Introduction}

Deep neural networks (DNNs) have obtained  great success  in many applications, including computer vision~\cite{he2016deep}, reinforcement learning~\cite{mnih2013playing}  and natural language processing~\cite{wolf2019huggingface}, etc.   However,  the theory of deep learning is much less explored compared with its great empirical success.  A key challenge of deep learning theory is that  deep neural networks are heavily overparameterized. Namely, the number of parameters is much larger than training samples. In practice, as the depth and width increasing,  the performance of deep NN also becomes better~\cite{srivastava2015training,zagoruyko2016wide},  which is far beyond the traditional learning theory regime. 

In the traditional neural networks and kernel methods literature, 
it is well known   the connection between the infinite width neural networks and Gaussian process~\cite{hornik1989multilayer},  and the universal approximation power of NN~\cite{leshno1993multilayer}.  However, these theories cannot  explain why the success of deep neural networks.   A recent work, Neural Tangent Kernel~\cite{jacot2018neural} (NTK),  shows the connection between training an infinite-width NN  and performing functional gradient descent in a Reproducing Kernel Hilbert Space(RKHS) associated with the NTK. Because of the convexity of the functional optimization problem,   Jacot et al. show the global convergence for infinite-width NN under the NTK regime.  Along this direction,  Hanin et al.~\cite{hanin2019finite} analyze the NTK with finite width and depth.    Shankar et al.~\cite{shankar2020neural} empirically investigate the performance of some simple compositional kernels, NTKs,  and deep neural networks.    Nitanda et al.~\cite{nitanda2020optimal} further show the minimax optimal convergence rate of  average stochastic gradient descent in a two-layer NTK regime.

Despite  the success of NTK~\cite{jacot2018neural} on showing the global convergence of NN, its expressive power is  limited. Zhu et al.~\cite{allen2019can} provide an example that shallow kernel methods (including NTK) need a much larger number of training samples to achieve the same small population risk compared with a three-layer ResNet.  They further point out the importance of hierarchical learning in deep neural networks~\cite{allen2020backward}.  In~\cite{allen2020backward}, they give the theoretical analysis of learning a target network family with square activation function under deep NN regime.  Besides,  there are quite a few works focus on the analysis of two-layer networks~\cite{bakshi2019learning, boob2017theoretical,kawaguchi2016deep,li2018learning,du2018gradient,yehudai2019power}  and shallow kernel methods without hierarchical learning~\cite{arora2019exact,ghorbani2021linearized,zou2019improved,daniely2016toward,lee2019wide}.  

Although some particular examples show deep models have more powerful  expressive power than  shallow ones~\cite{allen2019can,eldan2016power,allen2020backward}, how and why deep neural networks benefit from the depth remain unclear.  Zhu et al.~\cite{allen2020backward}  highlight the importance of a backward feature correction.  To better under deep neural networks,   we investigate the deep NN from a different kernel method perspective. 

   

Our contributions are summarized as follows:
\begin{itemize}
    \item  We propose a novel Neural Optimization Kernel (NOK) family that broadens the connection between kernel methods and deep neural networks.

    \item Theoretically, we show that the architecture of NOK performs  optimization of implicit regularization problems.  We prove the monotonic descent property for a wide range of both convex and non-convex regularized problems. Moreover, we prove a $O(1/T)$ convergence rate for convex regularized problems. Namely, our NOK family performs an  optimization through model architecture. A $T$-layer model performs $T$-step monotonic descent updates. 
    \item We propose a novel data-dependent structured approximation method, which establishes the connection between training deep neural networks and kernel methods associated with NOKs. The resultant computation graph is a ResNet-type finite width NN.   The activation function of NN  specifies the regularization problem explicitly or implicitly.   Our structured approximation preserved the monotonic descent property and  $O(1/T)$ convergence rate. Furthermore,    we propose both supervised and unsupervised learning schemes.  Moreover,  we prove a new Rademacher complexity bound and generalization bound of our structured approximated NOK architecture.
    \item Empirically,  we show that our unsupervised data-dependent structured approximation block can serve as a simple plug-in of popular backbones for a good generalization against input noise. Extensive experiments on  CIFAR10 and CIFAR100 with ResNet and DenseNet backbones show the good generalization of our structured approximated NOK against the Gaussian noise, Laplace noise, and FGSM adversarial  attack~\cite{goodfellow2014explaining}. 
\end{itemize}


\section{Neural Optimization Kernel}

Denote $\mathcal{L}_2$ as the Gaussian square-integrable function space, i.e., $\mathcal{L}_2:= \{f \big| \mathbb{E}_{\boldsymbol{w} \sim \mathcal{N}(\boldsymbol{0},\boldsymbol{I}_d)}[f(\boldsymbol{w})^2]< \infty\} $, and denote $\overline{\mathcal{L}}_2 $ as the spherically square-integrable function space, i.e.,  $\overline{\mathcal{L}}_2 := \{f \big| \mathbb{E}_{\boldsymbol{w} \sim Uni[\sqrt{d}\mathbb{S}^{d-1}]}[f(\boldsymbol{w})^2]< \infty\} $.  


Denote $\mathcal{F}=\mathcal{L}_2$ or $\mathcal{F}=\overline{\mathcal{L}}_2$,
$f(\cdot,\boldsymbol{x}) \in \mathcal{F}$ is a function indexed by $\boldsymbol{x}$. We simplify the notation $f(\boldsymbol{w},\boldsymbol{x})$ as $f(\boldsymbol{w})$  when the dependence of $\boldsymbol{x}$ is clear from the context.

For  $ \forall f(\cdot,\boldsymbol{x}),f(\cdot,\boldsymbol{y}) \in \mathcal{F}$, where $\mathcal{F}=\mathcal{L}_2$ or $\mathcal{F}=\overline{\mathcal{L}}_2$, define function $k(\cdot,\cdot): \mathcal{X}\times \mathcal{X} \to \mathcal{R}$ as 
\begin{align}
    k(\boldsymbol{x},\boldsymbol{y})= \mathbb{E}_{\boldsymbol{w}}{[f(\boldsymbol{w},\boldsymbol{x}) f(\boldsymbol{w},\boldsymbol{y}) ]}.   
\end{align}
Then, we know $k(\cdot,\cdot)$ is a bounded  kernel, which is shown in Proposition~\ref{Fkernel}. All detailed proofs are given in Appendix.
\begin{property}
\label{Fkernel}
For $ \forall f(\cdot,\boldsymbol{x}),f(\cdot,\boldsymbol{y}) \in \mathcal{F}$  ( $\mathcal{F}=\mathcal{L}_2$ or $ \mathcal{F} = \overline{\mathcal{L}}_2$), define function $k(\boldsymbol{x},\boldsymbol{y})= \mathbb{E}_{\boldsymbol{w}}{[f(\boldsymbol{w},\boldsymbol{x}) f(\boldsymbol{w},\boldsymbol{y}) ]}   :  \mathcal{X}\times \mathcal{X} \to \mathcal{R}  $,  then $k(\boldsymbol{x},\boldsymbol{y})$ is a bounded kernel, i.e.,  $k(\boldsymbol{x},\boldsymbol{y}) = k(\boldsymbol{y}, \boldsymbol{x}) < \infty$ and $k(\boldsymbol{x},\boldsymbol{y})$ is positive definite.  
\end{property}

For  $ \forall f \in \mathcal{F}$, where $\mathcal{F}=\mathcal{L}_2$ or $\mathcal{F}=\overline{\mathcal{L}}_2$, 
define operator $\mathcal{A}(\cdot): \mathcal{F} \to \mathcal{R}^d$ as $\mathcal{A}(f):=\mathbb{E}_{\boldsymbol{w}}{[\boldsymbol{w}f(\boldsymbol{w})]}$.  Define operator $\mathcal{A}^*: \mathcal{R}^d \to \mathcal{F}$ as $\mathcal{A}^*(\boldsymbol{x}) = \boldsymbol{w}^\top \boldsymbol{x}$, $\boldsymbol{w} \sim \mathcal{N}(\boldsymbol{0},\boldsymbol{I}_d)$ or $\boldsymbol{w} \sim Uni[\sqrt{d}\mathbb{S}^{d-1}]$. We know $\mathcal{A} \circ \mathcal{A}^*  (\cdot) = \mathbb{E}_{\boldsymbol{w} } {[\boldsymbol{w} \boldsymbol{w}^\top] } = \boldsymbol{I}_d : \mathcal{R}^d \to \mathcal{R}^d$. Details are provided in Appendix. Define operator $\Phi_\lambda(
\cdot): \mathcal{F} \to \mathcal{R}$ as $\Phi_\lambda(f):=\mathbb{E}_{\boldsymbol{w}} [ \phi_\lambda (f(\boldsymbol{w}))] $, where $\phi_\lambda(\cdot)$ is a function  with   parameter $\lambda$ and bounded from below, and $\mathbb{E}_{\boldsymbol{w}} [ \phi_\lambda (f(\boldsymbol{w}))]$ exists for some $f \in \mathcal{F}$.  Several examples of $\phi_\lambda$ and the corresponding proximal operators are shown in Table~\ref{Operator}.  It is worth noting that $\phi_\lambda(\cdot)$ can be either convex or non-convex.

 \begin{table*}[t]
  \caption{Regularizers and Proximal Operators}
   \resizebox{1\textwidth}{!}{
  \centering
\begin{tabular}{c|c|c|c}
\hline
 & $l_0$-norm~\cite{donoho1994ideal}    &  $l_1$-norm~\cite{donoho1992maximum}  &  MCP~\cite{zhang2010nearly}   \\ \hline
$ \phi _\lambda(z)$  &  $  \lambda\| z \|_0$       & $  \lambda\| z \|_1$  &  $\lambda \int_{0}^{|z|} max(0,1- x/(\gamma \lambda) ) \,dx $     \\ \hline
$h(z)$  & $ h(z) \! =\! \left\{
                \begin{array}{ll}
                  z,  \;\;\;   |z| \ge \sqrt{2\lambda} \\
                  0,    \;\;\;   |z| < \sqrt{2\lambda}
                \end{array}
              \right. .$   &   $  h(z) \! =\! \text{sign}(z) \max (0, |z| \!-\! \lambda ) $  &  $ h(z) \! =\! \left\{
                \begin{array}{ll}
                  z,   \;\;\;\;\;\;\;\;\;\;\;\;\;\;\;\;\;\; \;\;  |z| > {\gamma \lambda} \\ 
                  \frac{sign(z)(|z|-\lambda)}{1-1/\gamma},  \;\;   \lambda < |z| \le {\gamma \lambda}  \\
                  0,    \;\;\;\;\;\;\;\;\;\;\;\;\;\;\;\;\;\; \;\; |z| \le \lambda
                \end{array}
              \right. .$     \\ \hline
    & Capped $l_1$-norm~\cite{zhang2010analysis}  & SCAD~\cite{fan2001variable}  &   MCP0~\cite{olsson2017non}     \\ \hline     
$ \phi _\lambda(z)$ &  $\lambda \min(|z|,\gamma)$  &  $\lambda \int_{0}^{|z|} min(1, \frac{max(0,\gamma\lambda-z)}{(\gamma-1)\lambda}  ) \,dx  \; ,(\gamma>2)$  &  $  \phi _\lambda(z) \! =\! \frac{1}{2} (\lambda \!-\! max(\sqrt{\lambda} \!-\! |z| , 0)^2 )  $   \\ \hline 

$h(z)$  &   $h(z) \! =\! \left\{
                \begin{array}{ll}
                  x_1,   \;\;  q(x_1) \le  q(x_2) \\ 
                  x_2,    \;\;  q(x_1) >  q(x_2)
                \end{array} 
              \right.  , \; where    \begin{array}{ll}
                    x_1 = sign(z)\max(|z|,\gamma)  \\
                     x_2 = sign(z) \min(\gamma,\max(0,|z|-\lambda)) \\
                     q(x)=0.5(x-z)^2+\lambda \min(|x|,\gamma)
              \end{array}  $     &  $h(z) \! =\! \left\{
                \begin{array}{ll}
                  z,   \;\;\;\;\;\;\;\;\;\;\;\;\;\;\;\;\;\; \;\;  |z| > {\gamma \lambda} \\ 
                 \frac{(\gamma-1)z-sign(z)\gamma\lambda}{\gamma-2},  \;\;   2\lambda < |z| \le {\gamma \lambda}  \\
                  sign(z)\max(|z|-\lambda,0),   \;\;\; \;\; |z| \le 2\lambda
                \end{array}
              \right. $  &  $ h(z) \! =\! \left\{
                \begin{array}{ll}
                  z,   \;\;\;\;\;\;\;\;\;\;\;\;\;\;\;\;\;\; \;\;  |z| > {\sqrt{\lambda}} \\ 
                 \beta\sqrt{\lambda} \;\;\; (|\beta| \le 1) ,  \;\;  |z| = {\sqrt {\lambda}}  \\
                  0,    \;\;\;\;\;\;\;\;\;\;\;\;\;\;\;\;\;\; \;\; |z| < {\sqrt {\lambda}} 
                \end{array}
              \right. .$    \\ \hline 

\end{tabular}
}
\label{Operator}
\vspace{-3mm}
\end{table*}

Our Neural Optimization Kernel (NOK) is defined upon the solution of optimization problems. 
Before giving our Neural Optimization Kernel (NOK) definition, we first introduce a family of functional optimization problems.  The $\Phi_\lambda$-regularized optimization problem is defined as 
\begin{align}
\label{FOP1}
\min _ {f \in \mathcal{F}}  \frac{1}{2} & \|\boldsymbol{x} - \mathcal{A}(f) \|_{2}^2 +  \Phi_\lambda(f) \nonumber \\ & =   \frac{1}{2}  \|\boldsymbol{x} -  \mathbb{E}_{\boldsymbol{w}}[\boldsymbol{w}f(\boldsymbol{w})] \|_{2}^2 +  \mathbb{E}_{\boldsymbol{w}} [ \phi_\lambda (f(\boldsymbol{w}))] ,
\end{align}
where $\mathcal{F}=\mathcal{L}_2$ or $\mathcal{F}=\overline{\mathcal{L}}_2$.  $f(\cdot):=f(\cdot,\boldsymbol{x})$ is a function indexed by $\boldsymbol{x}$. We simplify the notation $f(\boldsymbol{w},\boldsymbol{x})$ as $f(\boldsymbol{w})$ as the dependence of $\boldsymbol{x}$ is clear from the context.

\textbf{Intuition:} The reconstruction problem in Eq.(\ref{FOP1}) can be viewed as an autoencoder.   We find a function embedding $f(\cdot,{\boldsymbol{x}})$ to represent the input data $\boldsymbol{x}$.  In contrast, the standard autoencoders usually extract finite-dimensional vector features to represent the input data  for downstream tasks~\cite{goodfellow2016deep,he2021masked}. Function representation may encode richer information than finite-dimensional vector features.

For $\phi_\lambda(\cdot)$ with efficient proximal operators $h(\cdot)$ defined as $h(z) = \argmin _{x} {\frac{1}{2}(x-z)^2 +  \phi_\lambda (x)} $, we can optimize the problem~(\ref{FOP1}) by iterative updating with Eq.(\ref{FUpdateRule}):
\begin{align}
\label{FUpdateRule}
    f_{t+1}(\cdot) = h\big( \mathcal{A}^* ( \boldsymbol{x} ) + f_t(\cdot) - \mathcal{A}^* \circ \mathcal{A} (f_t(\cdot)) \big).
\end{align}
 The  initialization is $f_0(\cdot)=0$. 
 
 \textbf{Remark:} In the update rule~(\ref{FUpdateRule}), the term $-\mathcal{A}^* \circ \mathcal{A} (f_t(\cdot))$ can be viewed as a two-layer transformed residual modular of $f_t(\cdot)$. Then adding a skip connection $f_t(\cdot)$ and a biased term $\mathcal{A}^* ( \boldsymbol{x} )$.  As shown in~\cite{allen2019can,allen2020backward}, a ResNet-type architecture (residual modular with skip connections) is crucial for  obtaining a small error with  sample and time efficiency.

For both convex and non-convex function $\phi_\lambda$, our update rule in Eq.(\ref{FUpdateRule}) leads to a monotonic descent. 
\begin{theorem}(Monotonic Descent)
\label{NonConvexF}
For a function $ \phi_\lambda (\cdot)$, 
 denote $h(\cdot)$ as the proximal operator of $ \phi_\lambda (\cdot)$.   Suppose $ |h(x)| \le c|x| $ (or $ |h(x)| \le c  $), $0<c<\infty$ (e.g.,  hard thresholding function).  Given a bouned $\boldsymbol{x} \in \mathcal{R}^d$,  set function $f_{t+1}(\cdot) = h\big( \mathcal{A}^* ( \boldsymbol{x} ) + f_t(\cdot) - \mathcal{A}^* \circ \mathcal{A} (f_t(\cdot)) \big) $ and $f_0  \in \mathcal{F}$ (e.g., $f_0=0$). 
Denote $ Q(f)= \frac{1}{2}  \|\boldsymbol{x} - \mathcal{A}(f) \|_{2}^2 +  \Phi_\lambda(f) $. For $ \forall t\ge 0$,    we have
\begin{align}
   &  Q(f_{t\!+\!1})  \! \le\! Q(f_t) \! \nonumber \\ &  \!-\! \frac{1}{2}\mathbb{E}_{\boldsymbol{w}}[ \left.( f_{t\!+\!1}(\boldsymbol{w})\! -\!f_t(\boldsymbol{w}) \!-\! \boldsymbol{w}^\top \mathbb{E}_{\boldsymbol{w}}[\boldsymbol{w} (f_{t\!+\!1}(\boldsymbol{w})\! -\!f_t(\boldsymbol{w})) ] \right.)^2 ] \nonumber \\ & \le  Q(f_{t})
\end{align}
\end{theorem}
\textbf{Remark:} Assumption$ |h(x)| \le c|x| $ (or $ |h(x)| \le c  $) is  used to ensure that each $f_t \in \mathcal{F}$. Neural networks with a activation function $h(\cdot)$, e.g.,  sigmoid, tanh,  and ReLU,  as long as $h(\cdot)$ satisfies the above assumption,  it corresponds to a  (\textbf{implicit})  $\phi(\cdot)$-regularized problem. Theorem~\ref{NonConvexF} shows that a $T$-layer network performs  $T$-steps monotonic descent updates of the $\phi(\cdot)$-regularized objective $Q(\cdot)$.


For a convex $\phi_\lambda$, we can achieve a $O(\frac{1}{T})$ convergence rate, which is formally shown in Theorem~\ref{ConvexF}.
\begin{theorem}
\label{ConvexF}
For a convex function $ \phi_\lambda (\cdot)$, 
 denote $h(\cdot)$ as the proximal operator of $ \phi_\lambda (\cdot)$.   Suppose $ |h(x)| \le c|x| $ (or $ |h(x)| \le c  $), $0< c <\infty$.  Given a bouned $\boldsymbol{x} \in \mathcal{R}^d$,  set function $f_{t+1}(\cdot) = h\big( \mathcal{A}^* ( \boldsymbol{x} ) + f_t(\cdot) - \mathcal{A}^* \circ \mathcal{A} (f_t(\cdot)) \big) $ and $f_0  \in \mathcal{F}$ (e.g., $f_0=0$). 
Denote $ Q(f)= \frac{1}{2}  \|\boldsymbol{x} - \mathcal{A}(f) \|_{2}^2 +  \Phi_\lambda(f) $ and $f_* \in \mathcal{F}$ as an optimal of $Q(\cdot)$. For $\forall T\ge 1$,    we have
\begin{align}
 &  T\big( Q(f_T  ) -  Q(f_*) \big) \nonumber \\ &  \le \frac{1}{2}\mathbb{E}_{\boldsymbol{w}} {[ (f_{0}(\boldsymbol{w})  \! - \!f_*(\boldsymbol{w}))^2  ]} \!-\! \frac{1}{2}\mathbb{E}_{\boldsymbol{w}} {[ (f_{T}(\boldsymbol{w})  \! - \!f_*(\boldsymbol{w}))^2  ]}   \nonumber \\ & -\frac{1}{2}\sum_{t=0}^{T-1}{ \|\mathbb{E}_{\boldsymbol{w}} { [ \boldsymbol{w} \big(f_{t}(\boldsymbol{w}) \!-\!f_*(\boldsymbol{w})\big) ]  }\|_2^2  } \nonumber \\  & -\frac{1}{2} \sum_{t=0}^{T-1}{  { (t+1)\mathbb{E}_{\boldsymbol{w}} {[ (f_{t\!+\!1}(\boldsymbol{w})  \! - \!f_t(\boldsymbol{w}))^2  ]}   }    } .
\end{align}
\end{theorem}
\textbf{Remark:} A ReLU $h(z)=\text{max}(z,0)$  corresponds a $\phi_\lambda(z)=\! =\! \left\{
                \begin{array}{ll}
                  0,  \;\;\;\;\;   z \ge 0 \\
                  +\infty,    \;   z < 0
                \end{array}
              \right. $ (a lower semi-continuous convex function) , which results in a convex regularization problem. A $T$-layer NN obtains $O(1/T)$ convergence rate , which is faster than non-convex cases. This explains the success of ReLU on training deep NN from a NN architecture optimization perspective. When ReLU and $f_0=0$ is used, the resultant first-layer kernel  is the arc-cosine kernel in \cite{cho2012kernel}. More interestingly,  when ReLU is used (related to indicator function $\phi_\lambda(\cdot)$),  the learned function representation $p(\boldsymbol{w})f_t( \boldsymbol{w} )$ is an  unnormalized  non-negative measure,  where $p(\boldsymbol{w})$ denotes the density of Gaussian or Uniform sphere surface distribution.  We can achieve a probability measure representation by normalizing $p(\boldsymbol{w})f_t(\boldsymbol{w}) $ with $Z= \int  {p(\boldsymbol{w})f_t(\boldsymbol{w}) } \,d\boldsymbol{w} $. 

Our \textbf{Neural Optimization Kernel (NOK)} is defined upon the optimized function $f_T$ ($T$-layer) as 
\begin{align}
    k_{T,\infty}(\boldsymbol{x},\boldsymbol{y}):= \mathbb{E}_{\boldsymbol{w}}{[f_T(\boldsymbol{w},\boldsymbol{x}) f_T(\boldsymbol{w},\boldsymbol{y}) ]}   .
\end{align}
With $\forall f_0(\cdot,\boldsymbol{x}), f_0(\cdot,\boldsymbol{y})  \in \mathcal{F}$, we know $ f_T(\cdot,\boldsymbol{x}), f_T(\cdot,\boldsymbol{y}) \in \mathcal{F}$. From the Proposition~\ref{Fkernel}, we know $k_{T,\infty}(\boldsymbol{x},\boldsymbol{y})$ is a bounded kernel.  

\section{Structured Approximation}

The orthogonal sampling~\cite{yu2016orthogonal}  and spherically structured sampling~\cite{lyu2017spherical,lyu2020subgroup} have been successfully used for Gaussian and spherical integral approximation.   In the QMC area, randomization of structured points set is standard and widely used   to achieve an unbiased estimator (the same marginal distribution $p(\boldsymbol{w})$).   In the hypercube domain $[0,1]^d$, a uniformly distributed vector shift  is employed. In the hypersphere domain $\mathbb{S}^{d-1}$, a uniformly random rotation  is used. For the purpose of acceleration,   \cite{lyu2017spherical}  employs   a diagonal random rotation matrix  to approximate the full matrix rotation, which results in a $O(d)$ rotation time complexity instead of $O(d^3)$ complexity in  computing SVD of random Gaussian matrix (full rotation).  When the goal is to reduce approximation error,  we can use the standard full matrix  random orthogonal rotation of the structured points~\cite{lyu2017spherical} as an unbiased estimator of integral on $Uni[\mathbb{S}^{d-1}]$. Moreover, we propose a new diagonal rotation method that maintains the $O(n\text{log}n)$ time complexity and $O(d)$ space complexity by FFT as  \cite{lyu2017spherical}, which may of independent interest for integral approximation. 


For all-layer trainable networks, we propose a data-dependent structured approximation as 
\begin{align}
    \boldsymbol{W} = \sqrt{d}\boldsymbol{R}^\top \boldsymbol{B}    \in \mathcal{R}^{d \times N},
\end{align}
where $\boldsymbol{R}^\top\boldsymbol{R} = \boldsymbol{R}\boldsymbol{R}^\top = \boldsymbol{I}_d $ is a trainable orthogonal matrix  parameter, $N$ denotes the number of samples, and $\boldsymbol{W}\boldsymbol{W}^\top /N= \boldsymbol{I}_d$.  The structured matrix $\boldsymbol{B}$ can either be a concatenate of random orthogonal matrices~\cite{yu2016orthogonal}, or be the structured matrix in~\cite{lyu2017spherical,lyu2020subgroup} to satisfy $\boldsymbol{W}\boldsymbol{W}^\top /N= \boldsymbol{I}_d$.  

Define operator $\widehat{\mathcal{A}}:=\frac{1}{N}\boldsymbol{W} \; : \mathcal{R}^N \to \mathcal{R}^d $ and $\widehat{\mathcal{A}}^*:=\boldsymbol{W}^\top \; : \mathcal{R}^d \to \mathcal{R}^N $.  Operator $\widehat{\mathcal{A}} $ is an approximation of $\mathcal{A} $ by taking expectation over finite samples.
Remarkably, by using our structured approximation, 
we have $\widehat{\mathcal{A}} \circ \widehat{\mathcal{A}}^*= \frac{1}{N}\boldsymbol{W}\boldsymbol{W}^\top =  \boldsymbol{I}_d$.

\textbf{Remark:}
The orthogonal property of the operator $\mathcal{A} \circ \mathcal{A}^* = \boldsymbol{I}_d$ is vitally important to achieve $O(\frac{1}{T})$ convergence rate with our update rule. It leads to a ResNet-type network architecture, which enables a stable gradient flow for training.   
When approximation with finite samples, standard Monte Carlo sampling does not maintain the orthogonal property, which degenerates the convergence. In contrast, our structured approximation  preserves the second order moment $\mathbb{E}{[\boldsymbol{w}\boldsymbol{w}^\top]}=\boldsymbol{I}_d$. Namely, our approximation   maintains the orthogonal property, i.e., $\widehat{\mathcal{A}} \circ \widehat{\mathcal{A}}^*=    \boldsymbol{I}_d$ .     With the orthogonal property,  we can obtain the same convergence rate (w.r.t. the approximation objective) with  our update rule. 
 Moreover, for a k-sparse constrained problem, we prove the strictly monotonic descent property of our structured approximation  when using   $\boldsymbol{B}$ in~\cite{lyu2017spherical,lyu2020subgroup}.

\subsection{Convergence Rate for Finite Dimensional Approximation Problem }
The finite approximation of problem~(\ref{FOP1}) is given as 
\begin{equation}
  \resizebox{1.0\hsize}{!}{  $\widehat{Q}(\boldsymbol{y})\!:=\!  \frac{1}{2} \! \|\boldsymbol{x} - \widehat{\mathcal{A}} ({\boldsymbol{y} }) \|_2^2 \!+\! \frac{1}{N} \phi_\lambda(\boldsymbol{y}) =  \frac{1}{2} \|\boldsymbol{x} -  \frac{1}{N}\boldsymbol{W} \boldsymbol{y} \|_2^2 \!+\! \frac{1}{N}\! \phi_\lambda(\boldsymbol{y}),$}
\end{equation}
where $\boldsymbol{y} \in \mathcal{R}^N$  and  $\phi_\lambda (\boldsymbol{y}) :=\sum_{i=1}^N \phi_\lambda(y_i) $.

The finite dimension update rule is given as :
\begin{align}
\label{AFUpdateRule}
    {\boldsymbol{y}_{t+1}} = h \big( \boldsymbol{W}^\top \boldsymbol{x} + (\boldsymbol{I} - \frac{1}{N} \boldsymbol{W}^\top \boldsymbol{W}) \boldsymbol{y}_t   \big).
\end{align}
Thanks to the structured $\boldsymbol{W}=\sqrt{d}\boldsymbol{R}^\top\boldsymbol{B}$,  we show the monotonic descent property, convergence rate for convex $\phi_\lambda$,  and a strictly monotonic descent for a k-sparse constrained problem. 

For both convex and non-convex $\phi_\lambda$, our update rule in Eq.(\ref{AFUpdateRule}) leads to a monotonic descent. 
\begin{theorem}(Monotonic Descent)
\label{ANonConvexF}
For a function $ \phi_\lambda (\cdot)$, 
 denote $h(\cdot)$ as the proximal operator of $ \phi_\lambda (\cdot)$.    Given a bouned $\boldsymbol{x} \in \mathcal{R}^d$,  set $  {\boldsymbol{y}_{t+1}} = h \big( \boldsymbol{W}^\top \boldsymbol{x} + (\boldsymbol{I} - \frac{1}{N} \boldsymbol{W}^\top \boldsymbol{W}) \boldsymbol{y}_t   \big)$ with $\frac{1}{N}\boldsymbol{W}\boldsymbol{W}^\top=\boldsymbol{I}_d$.
Denote $ \widehat{Q}(\boldsymbol{y}):=  \frac{1}{2}  \|\boldsymbol{x} - \widehat{\mathcal{A}} ({\boldsymbol{y} }) \|_2^2 + \frac{1}{N} \phi_\lambda(\boldsymbol{y})  $. For $t\ge 0$,    we have
\begin{align}
\label{FiniteMonotonicDescent}
       \widehat{Q}(\boldsymbol{y}_{t\!+\!1})  \! & \le\!  \widehat{Q}(\boldsymbol{y}_t) \!-\! \frac{1}{2N}\|\boldsymbol{y}_{t\!+\!1} \!-\! \boldsymbol{y}_t \|_2^2  \!+\! \frac{1}{2}\|\frac{1}{N}\boldsymbol{W}(\boldsymbol{y}_{t\!+\!1} \!-\! \boldsymbol{y}_t )\|_2^2 \nonumber \\ & = \widehat{Q}(\boldsymbol{y}_t) -\frac{1}{2N}\|(\boldsymbol{I}_d- \frac{1}{N}\boldsymbol{W}^\top\boldsymbol{W}) (\boldsymbol{y}_{t+1} -\boldsymbol{y}_t ) \|_2^2 \nonumber \\ & \le \widehat{Q}(\boldsymbol{y}_t) .
\end{align}
\end{theorem}
\textbf{Remark:} For the finite dimensional case, the monotonic descent property is preserved. For popular activation function, e.g.,  sigmoid, tanh and ReLU,  it corresponds to a finite dimensional (implicit) $\phi(\cdot)$-regularized problem. 
A $T$-layer NN performs $T$-steps monotonic descent of the $\phi(\cdot)$-regularized problem $\widehat{Q}(\cdot)$ desipte of the non-convexity of the activation function $h(\cdot)$. Interestingly,  by choosing different activation functions, we implicitly choose the regularizations of the optimization problem.  Our network structure can perform monotonic descent for a wide range of implicit optimization problems.

For convex $\phi_\lambda$, we can achieve a $O(\frac{1}{T})$ convergence rate, which is formally shown in Theorem~\ref{AConvexF}.
\begin{theorem}
\label{AConvexF}
For a convex function $ \phi_\lambda (\cdot)$, 
 denote $h(\cdot)$ as the proximal operator of $ \phi_\lambda (\cdot)$.     Given a bouned $\boldsymbol{x} \in \mathcal{R}^d$,  set  $  {\boldsymbol{y}_{t+1}} = h \big( \boldsymbol{W}^\top \boldsymbol{x} + (\boldsymbol{I} - \frac{1}{N} \boldsymbol{W}^\top \boldsymbol{W}) \boldsymbol{y}_t   \big)$ with $\frac{1}{N}\boldsymbol{W}\boldsymbol{W}^\top=\boldsymbol{I}_d$. 
Denote $ \widehat{Q}(\boldsymbol{y}):=  \frac{1}{2}  \|\boldsymbol{x} - \widehat{\mathcal{A}} ({\boldsymbol{y} }) \|_2^2 + \frac{1}{N} \phi_\lambda(\boldsymbol{y})  $ and $\boldsymbol{y}^* $ as an optimal of $\widehat{Q}(\cdot)$. For $T\ge 1$,    we have
\begin{align}
\label{FiniteConvergenceRate}
 &  T\big(\widehat{Q} (\boldsymbol{y}_T  ) - \widehat{Q}(\boldsymbol{y}^*)  \big) \nonumber \\ & \le \!\frac{1}{2N}\!\| \boldsymbol{y}_0 \!-\! \boldsymbol{y}^* \|_2^2  \!-\! \frac{1}{2N}\!\|\boldsymbol{y}_T \!-\! \boldsymbol{y}^* \|_2^2  \! -\!\frac{1}{2}\!\sum_{t=0}^{T-1}{\!  \|\frac{1}{N}\boldsymbol{W}(\boldsymbol{y}_t \!-\! \boldsymbol{y}^*) \|_2^2} \nonumber \\ & -\frac{1}{2} \sum_{t=0}^{T-1}{  { \frac{t+1}{N} \| \boldsymbol{y}_{t\!+\!1} -\boldsymbol{y}_t  \|_2^2 } } .
\end{align}
\end{theorem}
\textbf{Remark:} Term  $ -\frac{1}{2} \sum_{t=0}^{T-1}{  { \frac{t+1}{N} \| \boldsymbol{y}_{t\!+\!1} -\boldsymbol{y}_t  \|_2^2 } } $,   $   \! -\!\frac{1}{2}\!\sum_{t=0}^{T-1}{\!  \|\frac{1}{N}\boldsymbol{W}(\boldsymbol{y}_t \!-\! \boldsymbol{y}^*) \|_2^2} $,      and  term  $\!-\! \frac{1}{2N}\!\|\boldsymbol{y}_T \!-\! \boldsymbol{y}^* \|_2^2 $,   are always non-positive. Thus, we know $\widehat{Q} (\boldsymbol{y}_T  ) - \widehat{Q}(\boldsymbol{y}^*)  \le O(\frac{1}{T})$.


\begin{theorem}(Strictly Monotonic Descent of $k$-sparse  problem)
\label{StrictMDKsparse}
Let $
    L(\boldsymbol{y}) = \frac{1}{2} \| {\boldsymbol{x}} - \boldsymbol{D}\boldsymbol{y}  \|_2^2, \; s.t. \; \|\boldsymbol{y} \|_0 \le k$ with $\boldsymbol{D} = \frac{\sqrt{d}}{\sqrt{N}} \boldsymbol{R}^\top\boldsymbol{B}$, where $\boldsymbol{B}$ is constructed  as in~\cite{lyu2020subgroup} with $N=2n, d=2m$. 
    Set $\boldsymbol{y}_{t+1} = h(    \boldsymbol{a}_{t+1} ) $ with sparity $k$ and $\boldsymbol{a}_{t+1} = \boldsymbol{D}^\top\boldsymbol{x} +  (\boldsymbol{I}-\boldsymbol{D}^\top\boldsymbol{D})\boldsymbol{y}_t $. For $\forall t \ge 1$, we have
    \begin{align}
    L(\boldsymbol{y}_{t+1}) & \le    L(\boldsymbol{y}_t) + \frac{1}{2}\|\boldsymbol{y}_{t+1} - \boldsymbol{a}_{t+1}  \|_2^2 - \frac{1}{2}\|\boldsymbol{y}_t - \boldsymbol{a}_{t+1}  \|_2^2  \nonumber \\ & - \frac{n- (2k-1)\sqrt{n} -m }{2n}\| \boldsymbol{y}_{t+1} -\boldsymbol{y}_t \|_2^2 \le  L(\boldsymbol{y}_t) ,
   \end{align}
    where $h(\cdot)$ is defined as 
    \begin{align}
         h({z}_j) = \left\{
                \begin{array}{ll}
                  \!\!\!{z}_j  \;\;\;  \text{if} \;\;\;  | {z}_j | \; \text{is one of the k-highest  values of } |\boldsymbol{z} | \\
                  \!\!0    \;\;\;\;  \text{otherwise}
                \end{array}
              \right.
    \end{align}
\end{theorem}
\textbf{Remark:}  When sparsity $k< \frac{n-m+\sqrt{n}}{2\sqrt{n}}$, we have $L(\boldsymbol{y}_{t+1})  < L(\boldsymbol{y}_t) $ unless $\boldsymbol{y}_{t+1} = \boldsymbol{y}_{t} $.  Our update with the structured  $\boldsymbol{D}$ makes a strictly monotonic descent progress each step.

\subsection{Learning Parameter $\boldsymbol{R}$} 

\textbf{Supervised Learning:}  For the each $t^{th}$ layer, we can maintain an orthogonal matrix $\boldsymbol{R}_t$.  The orthogonal matrix $\boldsymbol{R}_t$ can be parameterized  by exponential mapping  or  Cayley mapping~\cite{helfrich2018orthogonal} of a skew-symmetric matrix. We can employ the Cayley mapping  to enable  gradient update w.r.t a   loss function $\ell(\cdot)$ in an end-to-end training.  Specifically, the orthogonal matrix $\boldsymbol{R}_t$ can be obtained by the  Cayley mapping of a skew-symmetric matrix as 
\begin{align}
    \boldsymbol{R}_t = (\boldsymbol{I} + \boldsymbol{M}_t ) (\boldsymbol{I} - \boldsymbol{M}_t)^{-1},
\end{align}
where $\boldsymbol{M}_t$ is a skew-symmetric matrix, i.e., $\boldsymbol{M}_t = -\boldsymbol{M}_t^\top \in \mathbb{R}^{d \times d}$.  For a skew-symmetric matrix $\boldsymbol{M}_t$, only the upper triangular matrix (without main diagonal) are free parameters. Thus, total the number of free parameters of $T$-Layer  is $Td(d-1)/2$.  Particularly, \textbf{when sharing the orthogonal matrix parameter}, i.e.,  $\boldsymbol{R}_1=\cdots=\boldsymbol{R}_T=\boldsymbol{R}$, the monotonic descent property and the convergence rate of the regularized optimization  problems are well maintained. In this case,  it is a \textbf{recurrent}  neural network architecture. 



\textbf{Unsupervised Learning:} The parameter $\boldsymbol{R}$ can also be learned in an unsupervised manner. Specifically,  for a finite dataset $\boldsymbol{X}$,  the finite dimensional approximation problem with the structured  $\boldsymbol{W}= \sqrt{d}\boldsymbol{R}^\top\boldsymbol{B}$ is given as 
\begin{align}
\label{FiniteRegularization}
    & \min _ {\boldsymbol{Y},\boldsymbol{R}} \frac{1}{2} \|\boldsymbol{X} - \frac{\sqrt{d}}{N}\boldsymbol{R}^\top\boldsymbol{B}\boldsymbol{Y} \|_{F}^2 +  \frac{1}{N}\phi _\lambda(\boldsymbol{Y}) \\  \nonumber 
& \text{subject to} \;\;   \boldsymbol{R}^\top\boldsymbol{R} = \boldsymbol{R} \boldsymbol{R}^\top = \boldsymbol{I}_d ,
\end{align}
where $\phi_\lambda(\cdot)$ is a separable non-convex or convex regularization function with parameter $\lambda$, i.e., $\phi_\lambda(\boldsymbol{Y}) = \sum_{i}{\phi_\lambda(\boldsymbol{y}^{(i)})}$. 

The problem~(\ref{FiniteRegularization}) can be solved by the alternative descent method. For a  fixed $\boldsymbol{R}$, we perform a iterative update of $\boldsymbol{Y}$ a few steps to decrease the objective. For the fixed   $\boldsymbol{Y}$,  parameter $\boldsymbol{R}$  has a closed-form solution.

\textit{Fix $\boldsymbol{R}$, Optimize $\boldsymbol{Y}$:} The problem~(\ref{FiniteRegularization}) can be rewritten as :
\begin{align}
    \frac{1}{2} \|\boldsymbol{X} - \frac{\sqrt{d}}{N}\boldsymbol{R}^\top\boldsymbol{B}\boldsymbol{Y} \|_{F}^2 +  \frac{1}{N}\phi _\lambda(\boldsymbol{Y}) = \sum_{i}{ \widehat{Q}(\boldsymbol{y}^{(i)}) }.
\end{align}
Thus, with fixed $\boldsymbol{R}$,  we can  update each $\boldsymbol{y}^{(i)}$ by Eq.(\ref{AFUpdateRule}) in parallel. We can perform $T_1$ steps update with initialization as the output of previous alternative phase, i.e., $\boldsymbol{Y}^j_{0}=\boldsymbol{Y}^{j-1}_{T_1}$   (and initialization $\boldsymbol{Y}^0_0=0$ and $\boldsymbol{R}_0=\boldsymbol{I}_d$). 

\textit{Fix $\boldsymbol{Y}$, Optimize  $\boldsymbol{R}$:} This is the nearest orthogonal matrix problem, which has a closed-form solution as shown in~\cite{schonemann1966generalized}. Let $\frac{\sqrt{d}}{N}\boldsymbol{B}\boldsymbol{Y}\boldsymbol{X}^\top =\boldsymbol{U}\Gamma\boldsymbol{V}^\top$ obtained  by singular value decomposition (SVD),  where $\boldsymbol{U},\boldsymbol{V}$ are orthgonal matrix. Then, Eq.(\ref{FiniteRegularization}) is minimized by $ \boldsymbol{R} = \boldsymbol{U}\boldsymbol{V}^\top$.

\textbf{Remark:} A $T_2$-step  alternative descent computation graph of $\boldsymbol{R}$ and $\boldsymbol{Y}$ can be viewed as a $T_1T_2$-layer NN block, which can be used as a plug-in of popular backbones  for a good generalization against  input noise.

\subsection{Kernel Approximation}

Define $k_{T,N}(\boldsymbol{x},\boldsymbol{x}')=\frac{1}{N}\left.<{\boldsymbol{y}}_T(\boldsymbol{W},\boldsymbol{x}),{\boldsymbol{y}}_T(\boldsymbol{W}, \boldsymbol{x}') \right.> $, where ${\boldsymbol{y}}_T(\boldsymbol{W},\boldsymbol{x}) : \mathcal{R}^d \to \mathcal{R}^N $ is a finite approximation of $f_T(\cdot,\boldsymbol{x}) \in \mathcal{H}_{k_T}$. We know $k_{T,N}(\boldsymbol{x},\boldsymbol{x}')$ is bounded kernel, and it is  an approximation of kernel $k_{T,\infty}=\mathbb{E}_{w}[f_T(\boldsymbol{w},\boldsymbol{x})f_T(\boldsymbol{w},\boldsymbol{x}')]$. 

\textbf{Remark:} Let $\boldsymbol{B}$ be a points set that  marginally uniformly distributed  on the surface of sphere $\mathbb{S}^{d-1}$ (e.g, block-wise random orthogonal rotation of  structured samples~\cite{lyu2017spherical}).   Employing our structured approximation $\boldsymbol{W}\!=\!\boldsymbol{R}^\top\boldsymbol{B}$,  we know   $\forall \boldsymbol{R} \in SO(d) $ and $\forall f \in \overline{\mathcal{L}}_2$,   $\lim_{N \to \infty}{\frac{\sum_{i=1}^N{f(\boldsymbol{w}_i)  }}{N} = \mathbb{E}_{\boldsymbol{w}\sim Uni[\mathbb{S}^{d-1}]}[f(\boldsymbol{w})]  } $.  It means that although the orthogonal rotation parameter $\boldsymbol{R}$ is learned, we still maintain an unbiased estimator of $\mathbb{E}_ {\boldsymbol{w}} {[f(\boldsymbol{w})] } $.

\textbf{First-Layer Kernel:} Set $\boldsymbol{y}=\boldsymbol{0}$ and $f_0=0$, we know $y_i(\boldsymbol{x})=h(\boldsymbol{w}_i^\top\boldsymbol{x})$ and $f_1(\boldsymbol{w},\boldsymbol{x})= h(\boldsymbol{w}^\top\boldsymbol{x})$. Suppose $ |h(x)| \le c|x| $ (or $ |h(x)| \le c  $), $0< c <\infty$, it follows that
\begin{align}
\label{FirstLayerApp}
    \lim_{N \to \infty}   {k_{1,N}(\boldsymbol{x},\boldsymbol{x}')}  & = \lim_{N \to \infty} {  \frac{1}{N}{\sum_{i=1}^N{ h( \boldsymbol{w}_i^\top \boldsymbol{x})  h( \boldsymbol{w}_i^\top \boldsymbol{x}'   )  }} } 
  \nonumber \\ &  = \mathbb{E}_{\boldsymbol{w}}{[ h( \boldsymbol{w}^\top \boldsymbol{R} \boldsymbol{x})  h( \boldsymbol{w}^\top \boldsymbol{R}\boldsymbol{x}'   ) ]} \nonumber \\ & =  \mathbb{E}_{\boldsymbol{w}}{[ h( \boldsymbol{w}^\top \boldsymbol{x})  h( \boldsymbol{w}^\top \boldsymbol{x}'   ) ]} = k_{1,\infty}(\boldsymbol{x},\boldsymbol{x}')
\end{align}
 In Eq.(\ref{FirstLayerApp}), we use the fact that a rotation does not change the uniform surface measure on $\mathbb{S}^{d-1}$.  The first layer  kernel   $k_{1,N}$ uniformly converge to $k_{1,\infty}$     over a bounded domain $\mathcal{X}\times\mathcal{X}$.

\textbf{Higher-Layer Kernel:} 
For both the shared $\boldsymbol{R}$ case and the unsupervised updating $\boldsymbol{R}$ case,  the monotonic descent property and convergence rate is well preserved for any bounded  $\boldsymbol{x}  \in \mathcal{X}$. 
With the same assumption of $h(\cdot)$ and $\boldsymbol{y}=\boldsymbol{0}$,   as $N \to \infty$,  we know $\boldsymbol{y}_t \to \widehat{f}_t \in \overline{\mathcal{L}}_2$ , where $\widehat{f}_t $ is a countable-infinite dimensional  function. And    inequality~(\ref{FiniteMonotonicDescent}) and inequality~(\ref{FiniteConvergenceRate}) uniformly  converges   to inequality~(\ref{InfhatMon})  and inequality~(\ref{InfhatConv})    over a bounded domain $\mathcal{X}$, respectively.  
\begin{equation}
\label{InfhatMon}
\begin{array}{l}
    \resizebox{1.0\hsize}{!}{  $  Q(\widehat{f}_{t\!+\!1})      \!  \le\!  Q(\widehat{f}_t)  \!-\! \frac{1}{2}\!\mathbb{E}_{\boldsymbol{w}} {[ \big(\widehat{f}_{t\!+\!1}(\boldsymbol{w})\! -\!\widehat{f}_t(\boldsymbol{w}) \big)^2 ]}  \!+\! \frac{1}{2}\!\|\mathbb{E}_{\boldsymbol{w}} { [ \boldsymbol{w} \big(\widehat{f}_{t\!+\!1}(\boldsymbol{w})\!-\!\widehat{f}_t(\boldsymbol{w})\big) ]  }\|_2^2 $ }  \\
    \resizebox{1\hsize}{!}{$\;\;\;\;\;\;\;\;\;\;\;\; = Q(\widehat{f}_t) - \frac{1}{2}\mathbb{E}_{\boldsymbol{w}}[ \left.( \widehat{f}_{t\!+\!1}(\boldsymbol{w})\! -\!\widehat{f}_t(\boldsymbol{w}) - \boldsymbol{w}^\top \mathbb{E}_{\boldsymbol{w}}[\boldsymbol{w} (\widehat{f}_{t\!+\!1}(\boldsymbol{w})\! -\!\widehat{f}_t(\boldsymbol{w})) ] \right.)^2 ]$}
    \\   \;\;\;\;\;\;\;\;\;   \le Q(\widehat{f}_t) . 
\end{array}
\end{equation}
\begin{equation}
\label{InfhatConv}
\begin{array}{l}
    \resizebox{1.0\hsize}{!}{  $  T\big( \! Q(\widehat{f}_T \! ) \!-\!  Q(f_*) \! \big)  \!\le \frac{1}{2}\mathbb{E}_{\boldsymbol{w}} {[ (f_{0}(\boldsymbol{w})  \! - \!f_*(\boldsymbol{w}))^2  ]} \!-\! \frac{1}{2}\mathbb{E}_{\boldsymbol{w}} {[ (\widehat{f}_{T}(\boldsymbol{w})  \! - \!f_*(\boldsymbol{w}))^2  ]}  $ } 
    \\ \resizebox{1.0\hsize}{!}{  $  \! -\!\frac{1}{2}\!\sum_{t=0}^{T-1}{\! \|\mathbb{E}_{\boldsymbol{w}} { [ \boldsymbol{w} \big(\widehat{f}_{t}(\boldsymbol{w}) \!-\!f_*(\boldsymbol{w})\big) ]  }\|_2^2  }  \!-\!\frac{1}{2} \!\sum_{t=0}^{T-1}{  { \! (t\!+\!1)\mathbb{E}_{\boldsymbol{w}} {[ (\widehat{f}_{t\!+\!1}(\boldsymbol{w})  \! - \!\widehat{f}_t(\boldsymbol{w}))^2  ]}   }    } $}
\end{array}
\end{equation}
It is worth noting that  $\lim_{N \to \infty} { {k}_{T,N}} $ converge   to a $\widehat{k}_{T,\infty}$ that is determined by the initialization $\boldsymbol{R}_0$ and dataset $\boldsymbol{X}$. Specifically,  for both the unsupervised learning case and the shared parameter case, the approximated kernel converge to a fixed kernel  as the width tends to infinity. As $N \to \infty$, training a finite structured NN with GD tends to perform a functional gradient descent with a fixed kernel. For a strongly convex regularized regression problem, functional gradient descent  leads to global convergence. 

For the case of  updating $T$-layer parameter $\boldsymbol{R}_t, t \in \{1,\cdots,T\}$ in a supervised manner, the sequence $\{\boldsymbol{R}_t\}$ determines the kernel.   When the data distribution is isotropic, e.g., $Uni[\mathbb{S}^{d-1}]$, the monotonic descent property   is preserved for the expectation $\mathbb{E}_{X}{[Q(\widehat{f}_t,X)]}$ (at least one step descent). Actually,  when parameters of each layer  are learned in a supervised manner,  the model is adjusted to fit the supervised signal.  When the prior regularization $\mathbb{E}_{\boldsymbol{w}} [ \phi_\lambda (\widehat{f}(\boldsymbol{w},X))] $ is consistent with learning the supervised signal, the monotonic descent property is well preserved. When the prior regularization contradicts  the supervised signal, the monotonic descent property for prior is weakened. 

\vspace{-3mm}
\section{Functional Optimization}
\vspace{-2mm}

We  can minimize a regularized  expected risk given as 
\begin{align}
     J(f) & := \underbrace{ \mathbb{E}_{ X,Y }{[ \ell( g(X) ), Y ] } \!+\! \frac{\lambda}{2} \|g\|_{\mathcal{H}_k}^2}_{{J_1}} \nonumber \\ & +  \beta \underbrace{\mathbb{E}_{X} {\big[ \frac{1}{2}  \|X \! -\!  \mathbb{E}_{\boldsymbol{w}}[\boldsymbol{w}f(\boldsymbol{w},X)] \|_{2}^2 +  \mathbb{E}_{\boldsymbol{w}} [ \phi_\lambda (f(\boldsymbol{w},X))]   \big] }}_{J_2} ,
\end{align}
where the function space $\mathcal{H}_{k} \ni g $ is is determined by the kernel  $ k(\boldsymbol{x},\boldsymbol{y}) \!=\! \mathbb{E}_{\boldsymbol{w}}{[f(\boldsymbol{w},\boldsymbol{x}) f(\boldsymbol{w},\boldsymbol{y}) ]}$.   $J_2$ can be viewed as an implicit  regularization to determine the candidate function family for supervised learning.  
Our NOK enables us to implicitly optimize the objective $J_2$ through neural network architecture.   Namely, the function space $ \mathcal{H}_{k_T} \ni g$ is determined by the kernel associated with the $T$-step update $f_T$.
With our NOK,  $J_2$ with convex regularization $\phi_\lambda(\cdot)$ can be optimized with a convergence rate $O(\frac{1}{T})$ by the $T$-layer network architecture. When $\phi_\lambda(\cdot)$ is an indicator function, the optimal $J_2$ actually is the $l_2$-norm optimal transport between $p(X)$ and a probability measure  induced by the transform of random variable $X$ (i.e., $ \mathbb{E}_{\boldsymbol{w}}[\boldsymbol{w}f^*_X(\boldsymbol{w},X)]  $).  By employing different activation function $h(\cdot)$, we implicitly choose the regularization term $J_2$ to be optimized.    

For a convex function $\ell(\cdot)$, $J_1(g)$ is strongly convex w.r.t the function $g\in \mathcal{H}_k$.  Functional gradient descent can converge to a minimizer of $J_1$. 
For regression problems, $\ell(z,y) =\frac{1}{2}(z-y)^2 $,  the functional gradient of $J_1$ is  
\begin{align}
      \partial J_1(g) & = \mathbb{E}_{X,Y}{ [ \partial_{z=g(X)} \ell(g(X),Y) k( \cdot,X) ]  } + \lambda g  \nonumber \\ & =  (\Sigma + \lambda I)g - \mathbb{E}_{X,Y}{ [ Y k(\cdot,X) ] }, 
    \label{FG}
\end{align}
where  $\Sigma:=\mathbb{E}_{X \sim p_X}{[k(\cdot,X) \otimes_{\mathcal{H}} k(X,\cdot )} ]  $ denotes the covariance operator. 

We can perform the average stochastic gradient descent using a stochastic unbiased estimator of Eq.(\ref{FG}). Since it is a strongly convex  problem, we can achieve $O(\frac{1}{T})$ convergence rate (Theorem A in \cite{nitanda2020optimal}). It means that training deep NN (with our structured approximated NOK architecture)  at the infinity width regime converges to a global minimum.

\vspace{3mm}
\section{Rademacher Complexity and Generalization Bound}

We show the  Rademacher complexity bound  and the generalization bound of our structured approximated NOK (SNOK).  

\textbf{Neural Network Structure:}
For structured approximated NOK networks (SNOK), the $1$-$T$ layers are given as 
\begin{align}
    \boldsymbol{y}_{t+1} = h( \boldsymbol{D}^\top\boldsymbol{R}_t\boldsymbol{x} +  (\boldsymbol{I}-\boldsymbol{D}^\top\boldsymbol{D})\boldsymbol{y}_t   ) ,
\end{align}
where $\boldsymbol{R}_t$ are free parameters such that $\boldsymbol{R}_t^\top\boldsymbol{R}_t=\boldsymbol{R}_t^\top \boldsymbol{R}_t = \boldsymbol{I}_d$. And $\boldsymbol{D}$ is a  scaled structured spherical samples such that $\boldsymbol{D}\boldsymbol{D}^\top = \boldsymbol{I}_d$~\cite{lyu2017spherical}, and $\boldsymbol{y}_0 =\boldsymbol{0}$. 

The  last layer ( $(T\!\!+\!\!1)^{th}$ layer) is given by ${z} = \boldsymbol{w}^\top\boldsymbol{y}_{T\!+\!1}$.  Consider a $L$-Lipschitz continuous loss function $\ell( {z}, y) :  {\mathcal{Z}}  \times \mathcal{Y} \to [0,1]  $ with Lipschitz constant $L$ w.r.t the input ${z}$.

\textbf{Rademacher Complexity~\cite{bartlett2002rademacher}:} Rademacher complexity of a function class $\mathcal{G}$ is defined as 
\begin{align}
    \mathfrak{R}_N(\mathcal{G}):= \frac{1}{N} \mathbb{E} \left[ \sup _{g \in \mathcal{G}} \sum _{i=1} ^{N} \epsilon_i g(\boldsymbol{x}_i)   \right], 
\end{align}
where $\epsilon_i, i \in \{1,\cdots,N\}$ are i.i.d. samples drawn uniformly from $\{+1,-1\}$ with probality $\text{P}[\epsilon_i=+1]=\text{P}[\epsilon_i=-1]=1/2$. And $\boldsymbol{x}_i,  i \in \{1,\cdots,N\}$ are i.i.d.  samples from $\mathcal{X}$.

\begin{theorem}
(Rademacher Complexity Bound) \label{RCB}
Consider  a Lipschitz continuous loss function $\ell( {z}, y) :  {\mathcal{Z}}  \times \mathcal{Y} \to [0,1]  $ with Lipschitz constant $L$ w.r.t the input ${z}$.    Let $\widetilde{\ell}({z},y):= \ell( z, y)  -  \ell( 0, y) $.  Let $\widehat{G}$ be the function class of our $(T\!\!+\!\!1)$-layer SNOK mapping from $\mathcal{X}$ to $\mathcal{Z}$. Suppose the activation function $|h(\boldsymbol{y})|\le |\boldsymbol{y}|$ (element-wise),  and the  $l_2$-norm of last layer weight is bounded, i.e.,   $\|\boldsymbol{w} \|_2 \le \mathcal{B}_w$.  Let $(\boldsymbol{x}_i, y_i)_{i=1}^N$ be i.i.d. samples drawn from $\mathcal{X} \times \mathcal{Y}$.   Let $\boldsymbol{Y}_{T\!+\!1}=[\boldsymbol{y}_{T\!+\!1}^{(1)}, \cdots, \boldsymbol{y}_{T\!+\!1}^{(N)} ]$ be the $T^{th}$ layer  output with input $\boldsymbol{X}$.  Denote the mutual coherence of $\boldsymbol{Y}_{T\!+\!1}$ as $\mu^*$, i.e.,  $ \mu^* =  \mu(\boldsymbol{Y}_{T\!+\!1}) \le 1$.   Then, we have
\begin{align}
    \mathfrak{R}_N(\widetilde{\ell} \circ \widehat{G} ) & = \frac{1}{N} \mathbb{E} \left[ \sup _{g \in \widehat{\mathcal{G}} } \sum _{i=1} ^{N} \epsilon_i  \widetilde{\ell}( g(\boldsymbol{x}_i) , y_i )    \right] \nonumber \\ &  \le  \frac{L\mathcal{B}_w \sqrt{ \big((N-1) \mu^* +1 \big) T} }{N}  \| \boldsymbol{X} \|_F ,
\end{align}
where    $\boldsymbol{X}\!=\![\boldsymbol{x}_1,\!\cdots\!,\boldsymbol{x}_N]$,  and $\|\cdot \|_F$ and $\|\cdot\|_2$ denote the matrix Frobenius norm and matrix spectral norm,respectively.
\end{theorem}
\textbf{Remark:} A small mutual coherence  $\mu(\boldsymbol{Y}_{T\!+\!1})$ leads to a small Rademacher complexity bound.  Moreover,  the Rademacher complexity bound  has a complexity  $O(\sqrt{T})$ w.r.t. the depth of NN  (SNOK).

\begin{theorem} (Generalization Bound) \label{Generalization}
Consider  a Lipschitz continuous loss function $\ell( {z}, y) :  {\mathcal{Z}}  \times \mathcal{Y} \to [0,1]  $ with Lipschitz constant $L$ w.r.t the input ${z}$.    Let $\widetilde{\ell}({z},y):= \ell( z, y)  -  \ell( 0, y) $.  Let $\widehat{G}$ be the function class of our $(T\!\!+\!\!1)$-layer SNOK mapping from $\mathcal{X}$ to $\mathcal{Z}$. Suppose the activation function $|h(\boldsymbol{y})|\le |\boldsymbol{y}|$ (element-wise),  and the  $l_2$-norm of last layer weight is bounded, i.e.,   $\|\boldsymbol{w} \|_2 \le \mathcal{B}_w$.  Let $(\boldsymbol{x}_i, y_i)_{i=1}^N$ be i.i.d. samples drawn from $\mathcal{X} \times \mathcal{Y}$.   Let $\boldsymbol{Y}_{T\!+\!1}=[\boldsymbol{y}_{T\!+\!1}^{(1)}, \cdots, \boldsymbol{y}_{T\!+\!1}^{(N)} ]$ be the $T^{th}$ layer  output with input $\boldsymbol{X}$. Denote the mutual coherence of $\boldsymbol{Y}_{T\!+\!1}$ as $\mu^*$, i.e.,  $ \mu^* =  \mu(\boldsymbol{Y}_{T\!+\!1}) \le 1$.   Then, for $\forall N$ and $\forall \delta, 0<\delta<1$, with a probability at least $1-\delta$, $\forall g \in \widehat{G}$,  we have
\begin{align}
 \mathbb{E} \big[ \ell(g(X),Y) \big]    & \le \!  \!\frac{1}{N}\!\sum_{i=1}^N \! \ell(g(\boldsymbol{x}_i),y_i) \nonumber \\ & \!+\!  \frac{L\mathcal{B}_w \sqrt{ \big((N\!-\!1) \mu^* \!+\!1 \big) T } }{N} \!   \| \boldsymbol{X} \|_{\!F}  \!+\!  \sqrt{\!\frac{8 \ln (2/\delta)}{N}}  
\end{align}
where    $\boldsymbol{X}\!=\![\boldsymbol{x}_1,\!\cdots\!,\boldsymbol{x}_N]$,   and  $\|\cdot \|_F$ and $\|\cdot\|_2$ denote the matrix Frobenius norm and matrix spectral norm,respectively.
\end{theorem}
\textbf{Remark:}  The mutual coherence  $\mu(\boldsymbol{Y}_{\!T\!+\!1})$ (or $\|\boldsymbol{Y}_{\!T\!+\!1}^\top\! \boldsymbol{Y}_{\!T\!+\!1} \!\!-\!\! \boldsymbol{I}\|_F^2$, etc.) of the last layer representation can serve as a good regularization to reduce Rademacher complexity and generalization bound.  
A small mutual coherence means that the direction feature vectors ($\frac{\boldsymbol{y}^i_{T+1}}{\|\boldsymbol{y}^i_{T+1}\|}$) are well spaced on the hypersphere.  Namely,  encouraging the last-layer embedding direction feature vectors  ($\frac{\boldsymbol{y}^i_{T+1}}{\|\boldsymbol{y}^i_{T+1}\|}$)  well spaced on the hypersphere leads to small Rademacher complexity and generalization bounds.
When the width of SNOK ($N_{\!D}$) is large enough, specifically, when $N_{\!D}\!>\! N$, it is possible to obtain $\mu(\boldsymbol{Y}_{\!T\!+\!1\!})\!=\!0$ (orthogonal representation), which significantly reduces the generalization bound. Namely, overparameterized deep NNs can increase the expressive power to reduce empirical risk \cite{allen2019can,eldan2016power,allen2020backward} and reduce the generalization bound at the same time.

\section{Experiments}


\begin{figure*}[h]
\centering
\subfigure[\scriptsize{DenseNet-CIFAR10}]{
\label{Densenet-CIFAR10-adv}
\includegraphics[width=0.224\linewidth]{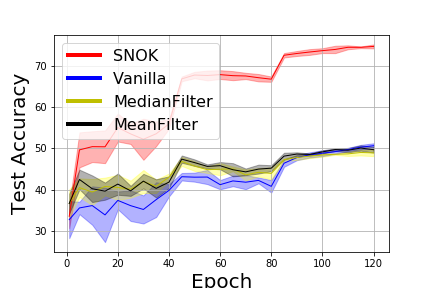}}
\subfigure[\scriptsize{DenseNet-CIFAR100}]{
\label{Densenet-CIFAR100-adv}
\includegraphics[width=0.224\linewidth]{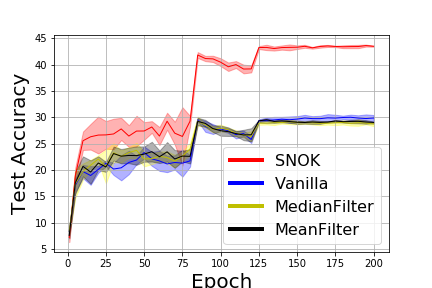}}
\subfigure[\scriptsize{ResNet34-CIFAR10}]{
\label{Resnet34-CIFAR10-adv}
\includegraphics[width=0.224\linewidth]{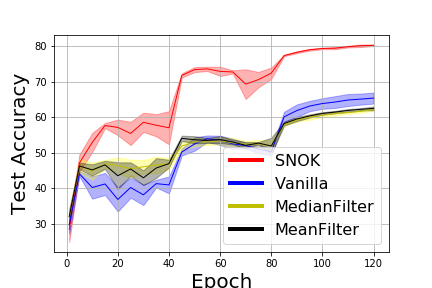}}
\subfigure[\scriptsize{ResNet34-CIFAR100}]{
\label{Resnet34-CIFAR100-adv}
\includegraphics[width=0.224\linewidth]{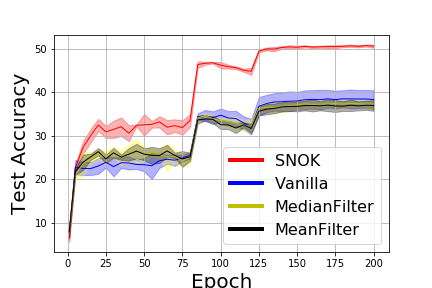}}
\caption{  Mean test accuracy $\pm$ std over 5 independent  runs on CIFAR10/CIFAR100 dataset under FGSM adversarial attack for DenseNet and ResNet  backbone. }
\label{ADVtest}
\end{figure*}

\begin{figure*}[h]
\centering
\subfigure[\scriptsize{CIFAR10-Clean}]{
\label{DensenetClean}
\includegraphics[width=0.224\linewidth]{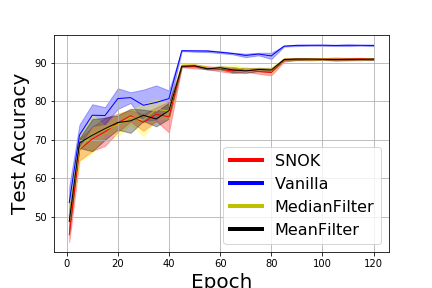}}
\subfigure[\scriptsize{CIFAR10-Gaussian-0.1}]{
\label{Densenet01}
\includegraphics[width=0.224\linewidth]{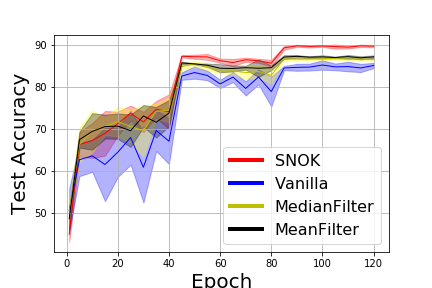}}
\subfigure[\scriptsize{CIFAR10-Gaussian-0.2}]{
\label{Densenet02}
\includegraphics[width=0.224\linewidth]{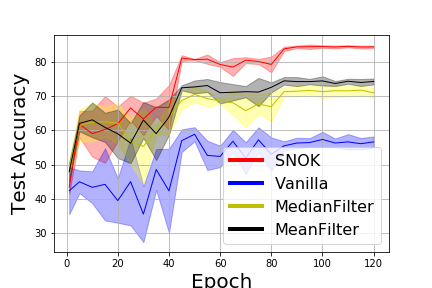}}
\subfigure[\scriptsize{CIFAR10-Gaussian-0.3}]{
\label{Densenet03}
\includegraphics[width=0.224\linewidth]{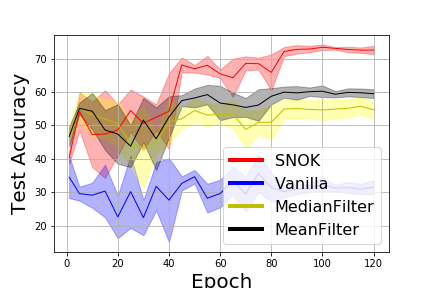}}
\subfigure[\scriptsize{CIFAR100-Clean}]{
\label{ResnetClean}
\includegraphics[width=0.224\linewidth]{./results/Densenet_clean_cifar10.png}}
\subfigure[\scriptsize{CIFAR100-Gaussian-0.1}]{
\label{Resnet01}
\includegraphics[width=0.224\linewidth]{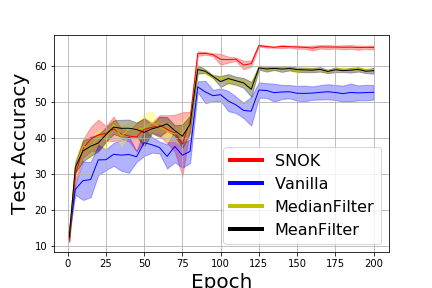}}
\subfigure[\scriptsize{CIFAR100-Gaussian-0.2}]{
\label{Resnet02}
\includegraphics[width=0.224\linewidth]{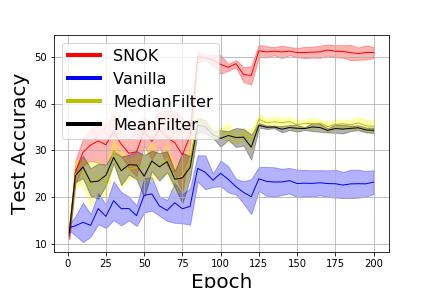}}
\subfigure[\scriptsize{CIFAR100-Gaussian-0.3}]{
\label{Resnet03}
\includegraphics[width=0.224\linewidth]{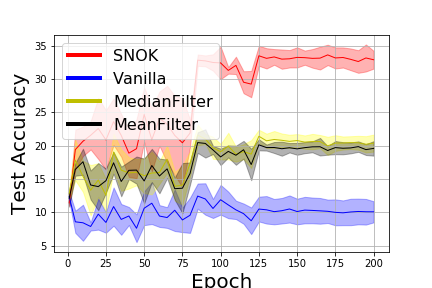}}
\caption{  Mean test accuracy $\pm$ std over 5 independent  runs on DenseNet  with Gaussian noise.   }
\label{DenseNetGaussian}
\end{figure*}
\vspace{5mm}

We evaluate the performance of our unsupervised SNOK blocks on classification tasks with input noise (Gaussian noise or Laplace noise), and under FGSM adversarial attack~\cite{goodfellow2014explaining}. In all the experiments, the input noise is added after  input normalization.  The standard deviation  of input  noise is set to  $\{0, 0.1, 0.2, 0.3\}$, respectively.      We employ both DenseNet-100~\cite{huang2017densely} and ResNet-34~\cite{he2016deep} as backbone.   We test the performance of  four methods in comparison:  (1) \textit{Vanilla Backbone}, (2) \textit{Backbone + Mean Filter}, (3) \textit{Backbone + Median Filter}, (4) \textit{Backbone + SNOK}.   For both \textit{ Mean Filter} and \textit{Median Filter} cases, we set  the filter neighborhood size as $3\times 3$ same as in~\cite{xie2019feature}.  For our SNOK case, we plug two SNOK blocks before and after the first  learnable \textit{Conv2D}  layer. In all the experiments,   CIFAR10 and CIFAR100 datasets~\cite{krizhevsky2009learning} are employed for evaluation. 
All the methods are evaluated over five independent runs with seeds $\{1,2,3,4,5\}$. During training, we stored the model every five epochs, and reported all evaluation results over all the stored models.  It covers the whole training trajectory, which is more informative.

The experimental results of different models under the FGSM attack are shown in Fig.~\ref{ADVtest}. Our SNOK plug-in achieves a significantly higher test accuracy than baselines.  The results of classification with   Gaussian input noise on DenseNet backbone are shown in Fig.~\ref{DenseNetGaussian}. Our SNOK obtains  competitive performance on the clean case and increasingly better performance as the std increases.  More detailed experimental results are presented in Appendix~\ref{AppendixExp}.  

\section{Conclusion and Future Work}
\label{Conclusion}


We proposed a novel kernel family NOK that broadens the connection between deep neural networks and kernel methods.  The architecture of our structured approximated NOK performs monotonic descent updates of implicit regularization problems.  We can implicitly choose the regularization problems by employing different activation functions, e.g., ReLU, max pooling, and soft-thresholding.  Moreover,  by taking advantage of the connection to kernel methods,   we show that training regularized NOK  at an infinite width regime with functional gradient descent converges to a global minimizer.  
Furthermore, we establish generalization bounds of our SNOK. We show that increasing 
the width of SNOK can increase the expressive power to reduce the empirical risk and potentially reduce the generalization bound simultaneously through last-layer feature mutual coherence regularization (i.e., $\mu(\boldsymbol{Y}_{T+1})$).  In particular,  when the width of SNOK is larger than the number of training data, last-layer orthogonal representation can significantly reduce the  generalization bound.    Our  unsupervised  structured approximated NOK block  can serve as a simple plug-in of popular backbones for a good generalization against input noise.  Extensive experiments on  CIFAR10 and CIFAR100 with ResNet and DenseNet backbones show the good generalization  of our structured approximated NOK against the Gaussian noise, Laplace noise, and FGSM adversarial attack. 

In the future, we will investigate the convergence behavior of training the supervised SNOK with SGD at a finite width regime.   More interestingly, we will investigate our SNOK with shared parameter $\boldsymbol{R}$ as recurrent neural network architectures.

\bibliography{DeepNOK}
\bibliographystyle{plain}

\appendix

\newpage
\onecolumn

\section{Proof of Proposition~\ref{Fkernel}}

\textbf{Kernel Property:}
\begin{proposition*}
For $\forall f(\cdot,\boldsymbol{x}),f(\cdot,\boldsymbol{y}) \in \mathcal{F}  $  ( $\mathcal{F}=\mathcal{L}_2$ or $ \mathcal{F} = \overline{\mathcal{L}}_2$), define function $k(\boldsymbol{x},\boldsymbol{y})= \mathbb{E}_{\boldsymbol{w}}{[f(\boldsymbol{w},\boldsymbol{x}) f(\boldsymbol{w},\boldsymbol{y}) ]}   :  \mathcal{X}\times \mathcal{X} \to \mathcal{R}  $,  then $k(\boldsymbol{x},\boldsymbol{y})$ is a bounded kernel, i.e.,  $k(\boldsymbol{x},\boldsymbol{y}) = k(\boldsymbol{y}, \boldsymbol{x}) < \infty$ and $k(\boldsymbol{x},\boldsymbol{y})$ is positive definite.  
\end{proposition*}
\begin{proof}
(i) Symmetric property is straightforward by definition.  

(ii) From Cauchy–Schwarz inequality, 
\begin{equation}
    k(\boldsymbol{x},\boldsymbol{y}) = \mathbb{E}_{\boldsymbol{w}}{[f(\boldsymbol{w},\boldsymbol{x}) f(\boldsymbol{w},\boldsymbol{y}) ]}  \le \sqrt{ \mathbb{E}_{\boldsymbol{w}}{[f(\boldsymbol{w},\boldsymbol{x})^2 ]} \mathbb{E}_{\boldsymbol{w}}{[f(\boldsymbol{w},\boldsymbol{y})^2 ]}   }  < \infty
\end{equation}

(iii) Positive definite property. For $\forall n \in \mathbb{N} $,   $ \forall \alpha_1 \cdots,  \alpha_n \in \mathcal{R}$ and $\forall \boldsymbol{x}_1, \cdots, \boldsymbol{x}_n \in \mathcal{X}$, we have
\begin{align}
    \sum_i {\sum_j {\alpha_i\alpha_j k(\boldsymbol{x}_i,\boldsymbol{x}_j) } } = \mathbb{E}_{\boldsymbol{w}} \big[  \big( \sum_i \alpha_i f(\boldsymbol{w},\boldsymbol{x}_i) \big)^2 \big]  \ge 0 \nonumber
\end{align}

\end{proof}


\section{Proof of Theorem~\ref{ConvexF}}

\textbf{Convex $\phi$-regularization:}
\begin{align}
& \min _ {f \in \mathcal{F}}  \frac{1}{2}  \|\boldsymbol{x} -  \mathbb{E}_{\boldsymbol{w}}[\boldsymbol{w}f(\boldsymbol{w})] \|_{2}^2 +  \mathbb{E}_{\boldsymbol{w}} [ \phi_\lambda (f(\boldsymbol{w}))] 
\end{align}
where $\mathcal{F}=\mathcal{L}_2$ or $\mathcal{F}=\overline{\mathcal{L}}_2$. And
 $\mathcal{L}_2$ denotes the Gaussian square integrable functional space, i.e., $\mathcal{L}_2:= \{f \big| \mathbb{E}_{\boldsymbol{w} \sim \mathcal{N}(\boldsymbol{0},\boldsymbol{I}_d)}[f(\boldsymbol{w})^2]< \infty\} $, $\overline{\mathcal{L}}_2 $ denotes the sphere square integrable functional space, i.e.,  $\overline{\mathcal{L}}_2 := \{f \big| \mathbb{E}_{\boldsymbol{w} \sim Uni[\sqrt{d}\mathbb{S}^{d-1}]}[f(\boldsymbol{w})^2]< \infty\} $   and $\phi_\lambda(\cdot)$ denotes a  convex function bounded from below. 

\begin{Lemma}
$\mathbb{E}_{\boldsymbol{w} \sim Uni[\sqrt{d}\mathbb{S}^{d-1}]}[\boldsymbol{w}\boldsymbol{w}^\top] = \boldsymbol{I}_d$.
\end{Lemma}
\begin{proof}
\begin{align}
    \boldsymbol{I}_d & = \mathbb{E}_{\boldsymbol{x} \sim \mathcal{N}(\boldsymbol{0},\boldsymbol{I}_d) }[\boldsymbol{x}\boldsymbol{x}^\top]  \nonumber \\
    & = \int \frac{1}{(2\pi)^\frac{d}{2}} e^{-\frac{\|\boldsymbol{x}\|_2^2}{2}} \boldsymbol{x}\boldsymbol{x}^\top \,d\boldsymbol{x}  \nonumber \\
    & = \int_{0}^{\infty}  { \int_{\mathbb{S}^{d-1}}   { \frac{2\pi^\frac{d}{2}}{\Gamma(\frac{d}{2})} r^{d-1} \cdot e^{-\frac{r^2}{2}}r^2 \cdot  \frac{1}{(2\pi)^\frac{d}{2}}  \boldsymbol{v}\boldsymbol{v}^\top \,d\sigma(\boldsymbol{v})  }  \,d r} \\
    & = \int_{0}^{\infty}  { \frac{2\pi^\frac{d}{2}}{\Gamma(\frac{d}{2})} r^{d-1}  e^{-\frac{r^2}{2}}r^2 \cdot  \frac{1}{(2\pi)^\frac{d}{2}}  \,d r }  \int_{\mathbb{S}^{d-1}}   {  \boldsymbol{v}\boldsymbol{v}^\top \,d\sigma(\boldsymbol{v}) } \\
    & = \int_{0}^{\infty}  { \frac{r^{d-1} e^{-\frac{r^2}{2}}}{2^{\frac{d}{2}-1}\Gamma(\frac{d}{2})} r^2    \,d r }  \int_{\mathbb{S}^{d-1}}   {  \boldsymbol{v}\boldsymbol{v}^\top \,d\sigma(\boldsymbol{v}) } \\ 
    & =\mathbb{E}_{r \sim \chi(d)}{[r^2]}  \int_{\mathbb{S}^{d-1}}   {  \boldsymbol{v}\boldsymbol{v}^\top \,d\sigma(\boldsymbol{v}) }  \\
    & = d  \int_{\mathbb{S}^{d-1}}   {  \boldsymbol{v}\boldsymbol{v}^\top \,d\sigma(\boldsymbol{v}) }  \\
    & = \mathbb{E}_{\boldsymbol{w} \sim Uni[\sqrt{d}\mathbb{S}^{d-1}]}[\boldsymbol{w}\boldsymbol{w}^\top]
\end{align}
where $\sigma(\cdot)$ denotes the normalized  surface measure, $\chi(d)$ denotes the Chi distribution with degree $d$, $\Gamma(\cdot)$ denotes the gamma function.
\end{proof}

\begin{Lemma}
\label{BoundedMean}
Let  $ f \in  \mathcal{F}  $ with $\mathcal{F}=\mathcal{L}_2$ or $\mathcal{F}=\overline{\mathcal{L}}_2$, then  we have
\begin{align}
    \mathbb{E}_{\boldsymbol{w}}[f(\boldsymbol{w})^2] -  \|\mathbb{E}_{\boldsymbol{w}}[\boldsymbol{w}f(\boldsymbol{w})] \|_2^2= \mathbb{E}_{\boldsymbol{w}}[ \big( f(\boldsymbol{w}) - \boldsymbol{w}^\top \mathbb{E}_{\boldsymbol{w}}[\boldsymbol{w}f(\boldsymbol{w})] \big)^2 ] \ge 0
\end{align}
\end{Lemma}
\begin{proof}
Let $w_i$ denote the $i^{th}$ component of $\boldsymbol{w}$, from Cauchy–Schwarz inequality, we know that
\begin{align}
    (\mathbb{E}_{\boldsymbol{w}}[{w}_if(\boldsymbol{w})] )^2 \le \mathbb{E}_{\boldsymbol{w}}[{w}_i^2] \mathbb{E}_{\boldsymbol{w}}[f(\boldsymbol{w})^2] =  \mathbb{E}_{\boldsymbol{w}}[f(\boldsymbol{w})^2]  < \infty
\end{align}
Thus the expectation $\mathbb{E}_{\boldsymbol{w}}[\boldsymbol{w}f(\boldsymbol{w})]$ exits. 

Since $\mathbb{E}_{\boldsymbol{w}}[\boldsymbol{w}\boldsymbol{w}^\top]= \boldsymbol{I}_d$, we have
\begin{align}
     \| \mathbb{E}_{\boldsymbol{w}}[\boldsymbol{w}f(\boldsymbol{w})] \|_{2}^2 & = ( \mathbb{E}_{\boldsymbol{w}}[\boldsymbol{w}f(\boldsymbol{w})])^\top( \mathbb{E}_{\boldsymbol{w}}[\boldsymbol{w}f(\boldsymbol{w})]) \\
     & = ( \mathbb{E}_{\boldsymbol{w}}[\boldsymbol{w}f(\boldsymbol{w})])^\top \mathbb{E}_{\boldsymbol{w}}{ [ \boldsymbol{w}\boldsymbol{w}^\top ]}( \mathbb{E}_{\boldsymbol{w}}[\boldsymbol{w}f(\boldsymbol{w})]) \\
     & = \mathbb{E}_{\boldsymbol{w}} { [ ( \boldsymbol{w}^\top \mathbb{E}_{\boldsymbol{w}}[\boldsymbol{w}f(\boldsymbol{w})]  )^2 ]}
\end{align}
It follows that
\begin{align}
 &  \mathbb{E}_{\boldsymbol{w}}[f(\boldsymbol{w})^2] -  \|\mathbb{E}_{\boldsymbol{w}}[\boldsymbol{w}f(\boldsymbol{w})] \|_2^2 \nonumber \\ & =  \mathbb{E}_{\boldsymbol{w}}[f(\boldsymbol{w})^2] -  2\|\mathbb{E}_{\boldsymbol{w}}[\boldsymbol{w}f(\boldsymbol{w})] \|_2^2 + \|\mathbb{E}_{\boldsymbol{w}}[\boldsymbol{w}f(\boldsymbol{w})] \|_2^2 \\
   & = \mathbb{E}_{\boldsymbol{w}}[f(\boldsymbol{w})^2] -2 ( \mathbb{E}_{\boldsymbol{w}}[\boldsymbol{w}f(\boldsymbol{w})])^\top( \mathbb{E}_{\boldsymbol{w}}[\boldsymbol{w}f(\boldsymbol{w})]) + \mathbb{E}_{\boldsymbol{w}} { [ ( \boldsymbol{w}^\top \mathbb{E}_{\boldsymbol{w}}[\boldsymbol{w}f(\boldsymbol{w})]  )^2 ]} \\
   & = \mathbb{E}_{\boldsymbol{w}}[f(\boldsymbol{w})^2] -2  \mathbb{E}_{\boldsymbol{w}}[f(\boldsymbol{w}) \boldsymbol{w}^\top  \mathbb{E}_{\boldsymbol{w}}[\boldsymbol{w}f(\boldsymbol{w})] ] + \mathbb{E}_{\boldsymbol{w}} { [ ( \boldsymbol{w}^\top \mathbb{E}_{\boldsymbol{w}}[\boldsymbol{w}f(\boldsymbol{w})]  )^2 ]} \\
   & =  \mathbb{E}_{\boldsymbol{w}}[ \big( f(\boldsymbol{w}) - \boldsymbol{w}^\top \mathbb{E}_{\boldsymbol{w}}[\boldsymbol{w}f(\boldsymbol{w})] \big)^2 ] \ge 0
\end{align}

\end{proof}


\begin{Lemma}
Let $f \in \mathcal{F} $ with $\mathcal{F}=\mathcal{L}_2$ or $\mathcal{F}=\overline{\mathcal{L}}_2$, we have
\begin{align}
     \|\boldsymbol{x} - \mathbb{E}_{\boldsymbol{w}}[\boldsymbol{w}f(\boldsymbol{w})] \|_{2}^2  =  \mathbb{E}_{\boldsymbol{w}} { [ ( \boldsymbol{w}^\top (\boldsymbol{x}- \mathbb{E}_{\boldsymbol{w}}[\boldsymbol{w}f(\boldsymbol{w})] ) )^2 ]}
\end{align}
\end{Lemma}
\begin{proof}
Since $\mathbb{E}_{\boldsymbol{w}}[\boldsymbol{w}\boldsymbol{w}^\top]= \boldsymbol{I}_d$, we have
\begin{align}
     \|\boldsymbol{x} - \mathbb{E}_{\boldsymbol{w}}[\boldsymbol{w}f(\boldsymbol{w})] \|_{2}^2 & = (\boldsymbol{x} - \mathbb{E}_{\boldsymbol{w}}[\boldsymbol{w}f(\boldsymbol{w})])^\top(\boldsymbol{x} - \mathbb{E}_{\boldsymbol{w}}[\boldsymbol{w}f(\boldsymbol{w})]) \\
     & = (\boldsymbol{x} - \mathbb{E}_{\boldsymbol{w}}[\boldsymbol{w}f(\boldsymbol{w})])^\top \mathbb{E}_{\boldsymbol{w}}{ [ \boldsymbol{w}\boldsymbol{w}^\top ]}(\boldsymbol{x} - \mathbb{E}_{\boldsymbol{w}}[\boldsymbol{w}f(\boldsymbol{w})]) \\
     & = \mathbb{E}_{\boldsymbol{w}} { [ ( \boldsymbol{w}^\top (\boldsymbol{x}- \mathbb{E}_{\boldsymbol{w}}[\boldsymbol{w}f(\boldsymbol{w})] ) )^2 ]}
\end{align}
\end{proof}

\begin{Lemma}
\label{FL}
Denote $L(f):= \frac{1}{2} \|\boldsymbol{x} -  \mathbb{E}_{\boldsymbol{w}}[\boldsymbol{w}f(\boldsymbol{w})] \|_{2}^2   $. 
For  $\forall {f},{g} \in \mathcal{F} $ with $\mathcal{F}=\mathcal{L}_2$ or $\mathcal{F}=\overline{\mathcal{L}}_2$, we have $ L(f) =  L(g) + \mathbb{E}_{\boldsymbol{w}} {[ \boldsymbol{w}^\top (\mathbb{E}_{\boldsymbol{w}}[\boldsymbol{w}g(\boldsymbol{w})]  -\boldsymbol{x}  ) \big(f(\boldsymbol{w}) -g(\boldsymbol{w}) \big) ]  } + \frac{1}{2}\|\mathbb{E}_{\boldsymbol{w}} { [ \boldsymbol{w} \big(f(\boldsymbol{w})-g(\boldsymbol{w})\big) ]  }\|_2^2 $.
\end{Lemma}
\begin{proof}
\begin{align}
    \frac{1}{2} \|\boldsymbol{x} - \mathbb{E}_{\boldsymbol{w}}[\boldsymbol{w}f(\boldsymbol{w})] \|_{2}^2   & = \frac{1}{2} \|\boldsymbol{x} -  \mathbb{E}_{\boldsymbol{w}}[\boldsymbol{w}g(\boldsymbol{w})]  + \mathbb{E}_{\boldsymbol{w}}[\boldsymbol{w}g(\boldsymbol{w})] - \mathbb{E}_{\boldsymbol{w}}[\boldsymbol{w}f(\boldsymbol{w})] \|_{2}^2   \\
    & = \frac{1}{2} \|\boldsymbol{x} -  \mathbb{E}_{\boldsymbol{w}}[\boldsymbol{w}g(\boldsymbol{w})] \|_2^2 + \frac{1}{2} \big\| \mathbb{E}_{\boldsymbol{w}}[\boldsymbol{w}g(\boldsymbol{w})] - \mathbb{E}_{\boldsymbol{w}}[\boldsymbol{w}f(\boldsymbol{w})]  \big\|_2^2 \nonumber \\
    & + \left< \boldsymbol{x} -  \mathbb{E}_{\boldsymbol{w}}[\boldsymbol{w}g(\boldsymbol{w})],  \mathbb{E}_{\boldsymbol{w}}[\boldsymbol{w}g(\boldsymbol{w})] - \mathbb{E}_{\boldsymbol{w}}[\boldsymbol{w}f(\boldsymbol{w})]   \right> 
\end{align}
The inner product term can be rewritten as 
\begin{align}
    \left< \boldsymbol{x} -  \mathbb{E}_{\boldsymbol{w}}[\boldsymbol{w}g(\boldsymbol{w})],  \mathbb{E}_{\boldsymbol{w}}[\boldsymbol{w}g(\boldsymbol{w})] - \mathbb{E}_{\boldsymbol{w}}[\boldsymbol{w}f(\boldsymbol{w})]   \right>  &  =  (\boldsymbol{x} -  \mathbb{E}_{\boldsymbol{w}}[\boldsymbol{w}g(\boldsymbol{w})])^\top  \mathbb{E}_{\boldsymbol{w}}[\boldsymbol{w} \big( g(\boldsymbol{w}) - f(\boldsymbol{w} )\big) ]  \\
    & = \mathbb{E}_{\boldsymbol{w}}[ (\boldsymbol{x} -  \mathbb{E}_{\boldsymbol{w}}[\boldsymbol{w}g(\boldsymbol{w})])^\top \boldsymbol{w} \big( g(\boldsymbol{w}) - f(\boldsymbol{w} )\big) ]
\end{align}
It follows that 
\begin{align}
     \frac{1}{2} \|\boldsymbol{x} - \mathbb{E}_{\boldsymbol{w}}[\boldsymbol{w}f(\boldsymbol{w})] \|_{2}^2 & = \frac{1}{2} \|\boldsymbol{x} -  \mathbb{E}_{\boldsymbol{w}}[\boldsymbol{w}g(\boldsymbol{w})] \|_2^2 + \frac{1}{2} \big\| \mathbb{E}_{\boldsymbol{w}}[\boldsymbol{w}\big( g(\boldsymbol{w}) - f(\boldsymbol{w} )\big) ]   \big\|_2^2 \nonumber \\
     & + \mathbb{E}_{\boldsymbol{w}} {[ \boldsymbol{w}^\top (\mathbb{E}_{\boldsymbol{w}}[\boldsymbol{w}g(\boldsymbol{w})]  -\boldsymbol{x}  ) \big(f(\boldsymbol{w}) -g(\boldsymbol{w}) \big) ]  } 
\end{align}

\end{proof}

\begin{Lemma}
\label{FLineq}
For $\forall f_t,f_{t+1},f^* \in \mathcal{F}$ with $\mathcal{F}=\mathcal{L}_2$ or $\mathcal{F}=\overline{\mathcal{L}}_2$, we have
\begin{align}
    L(f_{t+1}) \! & = \! L(f^*) \!\!+\!\! \mathbb{E}_{\boldsymbol{w}} {[ \boldsymbol{w}^\top (\mathbb{E}_{\boldsymbol{w}}[\boldsymbol{w}f_t(\boldsymbol{w})] \!\!-\!\!\boldsymbol{x}  ) \big(f_{t+1}(\boldsymbol{w}) \! - \!f^*(\boldsymbol{w}) \big) ]  }  \!\!+\!\!   \frac{1}{2}\|\mathbb{E}_{\boldsymbol{w}} { [ \boldsymbol{w} \big(f_{t+1}(\boldsymbol{w}) \!-\!f_t(\boldsymbol{w})\big) ]  }\|_2^2  \nonumber \\
    &  -\frac{1}{2}\|\mathbb{E}_{\boldsymbol{w}} { [ \boldsymbol{w} \big(f_{t}(\boldsymbol{w}) \!-\!f^*(\boldsymbol{w})\big) ]  }\|_2^2
\end{align}
\end{Lemma}
\begin{proof}
From Lemma~\ref{FL}, we know that
\begin{align}
\label{FL1}
 &   L(f_{t\!+\!1}) =  L(f_t) + \mathbb{E}_{\boldsymbol{w}} {[ \boldsymbol{w}^\top (\mathbb{E}_{\boldsymbol{w}}[\boldsymbol{w}f_t(\boldsymbol{w})]  -\boldsymbol{x}  ) \big(f_{t\!+\!1}(\boldsymbol{w}) -f_t(\boldsymbol{w}) \big) ]  } + \frac{1}{2}\|\mathbb{E}_{\boldsymbol{w}} { [ \boldsymbol{w} \big(f_{t\!+\!1}(\boldsymbol{w})-f_t(\boldsymbol{w})\big) ]  }\|_2^2  \\
 & L(f^*) =  L(f_t) + \mathbb{E}_{\boldsymbol{w}} {[ \boldsymbol{w}^\top (\mathbb{E}_{\boldsymbol{w}}[\boldsymbol{w}f_t(\boldsymbol{w})]  -\boldsymbol{x}  ) \big(f^*(\boldsymbol{w}) -f_t(\boldsymbol{w}) \big) ]  } + \frac{1}{2}\|\mathbb{E}_{\boldsymbol{w}} { [ \boldsymbol{w} \big(f^*(\boldsymbol{w})-f_t(\boldsymbol{w})\big) ]  }\|_2^2 
\end{align}
Plug $L(f_t)$ into Eq.(\ref{FL1}), we can obtain that
\begin{align}
     L(f_{t+1}) \! & = \! L(f^*) \!\!-\!\! \mathbb{E}_{\boldsymbol{w}} {[ \boldsymbol{w}^\top (\mathbb{E}_{\boldsymbol{w}}[\boldsymbol{w}f_t(\boldsymbol{w})] \!\!-\!\!\boldsymbol{x}  ) \big(f^*(\boldsymbol{w}) \! - \!f_t(\boldsymbol{w}) \big) ]  } 
    -\frac{1}{2}\|\mathbb{E}_{\boldsymbol{w}} { [ \boldsymbol{w} \big(f^*(\boldsymbol{w}) \!-\!f_t(\boldsymbol{w})\big) ]  }\|_2^2 \nonumber \\
    & +\!\! \mathbb{E}_{\boldsymbol{w}} {[ \boldsymbol{w}^\top (\mathbb{E}_{\boldsymbol{w}}[\boldsymbol{w}f_t(\boldsymbol{w})] \!\!-\!\!\boldsymbol{x}  ) \big(f_{t+1}(\boldsymbol{w}) \! - \!f_t(\boldsymbol{w}) \big) ]  }  \! + \!    \frac{1}{2}\|\mathbb{E}_{\boldsymbol{w}} { [ \boldsymbol{w} \big(f_{t+1}(\boldsymbol{w}) \!-\!f_t(\boldsymbol{w})\big) ]  }\|_2^2     \\
    & = \! L(f^*) \!\!+\!\! \mathbb{E}_{\boldsymbol{w}} {[ \boldsymbol{w}^\top (\mathbb{E}_{\boldsymbol{w}}[\boldsymbol{w}f_t(\boldsymbol{w})] \!\!-\!\!\boldsymbol{x}  ) \big(f_{t+1}(\boldsymbol{w}) \! - \!f^*(\boldsymbol{w}) \big) ]  }  \!\!+\!\!   \frac{1}{2}\|\mathbb{E}_{\boldsymbol{w}} { [ \boldsymbol{w} \big(f_{t+1}(\boldsymbol{w}) \!-\!f_t(\boldsymbol{w})\big) ]  }\|_2^2  \nonumber \\
    &  -\frac{1}{2}\|\mathbb{E}_{\boldsymbol{w}} { [ \boldsymbol{w} \big(f_{t}(\boldsymbol{w}) \!-\!f^*(\boldsymbol{w})\big) ]  }\|_2^2
\end{align}

\end{proof}

\begin{Lemma}
\label{FLineqPhi}
For a convex function $ \phi_\lambda (\cdot)$, 
 denote $h(\cdot)$ as the proximal operator of $ \phi_\lambda (\cdot)$, i.e., $h(z) = \argmin _{x} {\frac{1}{2}(x-z)^2 +  \phi_\lambda (x)} $, let $f_{t+1}  = h \circ g_{t+1} \in \mathcal{F} $  with $\mathcal{F}=\mathcal{L}_2$ or $\mathcal{F}=\overline{\mathcal{L}}_2$, then for $\forall f^* \in \mathcal{F}$,  we have 
\begin{align}
    \mathbb{E}_{\boldsymbol{w}} [ \phi_\lambda (f_{t+1}(\boldsymbol{w}))]  \le \mathbb{E}_{\boldsymbol{w}} [ \phi_\lambda (f^*(\boldsymbol{w}))] - \mathbb{E}_{\boldsymbol{w}} { [ \big( g_{t+1}(\boldsymbol{w}) -f_{t+1}(\boldsymbol{w})  \big) \big( f^*(\boldsymbol{w}) - f_{t+1} (\boldsymbol{w}) \big )  ]}
\end{align}
\end{Lemma}
\begin{proof}

Since $ \phi_\lambda (\cdot)$ is convex function and    $f_{t+1}(\boldsymbol{w})= \argmin _{{x} }  \phi_\lambda({x}) + \frac{1}{2} \| x - {g}_{t+1}(\boldsymbol{w})  \|_{2}^2  $,  we have
\begin{align}
 {0} \in \partial \phi_\lambda(f_{t+1}(\boldsymbol{w})) +  (f_{t+1}(\boldsymbol{w})  -g_{t+1}(\boldsymbol{w}) )   
 \implies	 (g_{t+1}(\boldsymbol{w})  -f_{t+1}(\boldsymbol{w}) )   \in \partial \phi_\lambda({f}_{t+1}(\boldsymbol{w})) 
\end{align}
From the definition of subgradient and  convex function $ \phi_\lambda (\cdot)$, we have
\begin{align}
    \phi_\lambda(f_{t+1}(\boldsymbol{w})) \le \phi_\lambda(f^*(\boldsymbol{w})) -   (g_{t+1}(\boldsymbol{w})  -f_{t+1}(\boldsymbol{w}) )  ( f^*(\boldsymbol{w}) - f_{t+1}(\boldsymbol{w})  )
\end{align}
It follows that
\begin{align}
    \mathbb{E}_{\boldsymbol{w}} [ \phi_\lambda (f_{t+1}(\boldsymbol{w}))]  \le \mathbb{E}_{\boldsymbol{w}} [ \phi_\lambda (f^*(\boldsymbol{w}))] - \mathbb{E}_{\boldsymbol{w}} { [ \big( g_{t+1}(\boldsymbol{w}) -f_{t+1}(\boldsymbol{w})  \big) \big( f^*(\boldsymbol{w}) - f_{t+1} (\boldsymbol{w}) \big )  ]} 
\end{align}
\end{proof}

\begin{Lemma}
\label{EnsurefinF}
 Denote $h(\cdot)$ as the proximal operator of $ \phi_\lambda (\cdot)$. Suppose $ |h(x)| \le c |x| $  (or $ |h(x)| \le c  $) , $0<c<\infty$ .  Given a bouned $\boldsymbol{x} \in \mathcal{R}^d$,  set function $g_{t+1}(\boldsymbol{w}) = \boldsymbol{w}^\top\boldsymbol{x}  + f_t(\boldsymbol{w}) - \boldsymbol{w}^\top \mathbb{E}_{\boldsymbol{w}}{[\boldsymbol{w}f_t(\boldsymbol{w})]} $ with $f_t \in \mathcal{L}_2$ and $\boldsymbol{w} \sim \mathcal{N}(\boldsymbol{0},\boldsymbol{I}_d)$ (or $f_t \in \overline{\mathcal{L}}_2$ and $\boldsymbol{w} \sim Uni[\sqrt{d}\mathbb{S}^{d-1}]$). Set $f_{t+1} = h \circ g_{t+1}$, then, we know $f_{t+1} \in \mathcal{F}$ with $\mathcal{F}=\mathcal{L}_2$ or $\mathcal{F}=\overline{\mathcal{L}}_2$,respectively. 
\end{Lemma}
\begin{proof}
\textbf{Case $|h(x)|\le c, \; 0<c<\infty$:}  It is straightforward to know $\mathbb{E}_{\boldsymbol{w}} { [  h (g_{t+1} (\boldsymbol{w}) )^2  ] } \le c^2 < \infty$, thus $f_{t+1} \in \mathcal{F}$.

\textbf{Case $|h(x)|\le c|x|, \; 0<c<\infty$:} 
Since $ |h(x)| \le  c |x| $, we know that
\begin{align}
    h (g_{t+1} (\boldsymbol{w}) )^2 \le  c^2 g_{t+1} (\boldsymbol{w})^2 & =  c^2 \big( \boldsymbol{w}^\top\boldsymbol{x}  + f_t(\boldsymbol{w}) - \boldsymbol{w}^\top \mathbb{E}_{\boldsymbol{w}}{[\boldsymbol{w}f_t(\boldsymbol{w})]} \big)^2  \\
    & \le   2 c^2  (\boldsymbol{w}^\top ( \boldsymbol{x}  - \mathbb{E}_{\boldsymbol{w}}{[\boldsymbol{w}f_t(\boldsymbol{w})]}   ) )^2 + 2 c^2  f_t(\boldsymbol{w})^2  
\end{align}
It follows that
\begin{align}
  \mathbb{E}_{\boldsymbol{w}} { [  h (g_{t+1} (\boldsymbol{w}) )^2  ] }  &  \le   c^2 \mathbb{E}_{\boldsymbol{w}} { [ \big( \boldsymbol{w}^\top\boldsymbol{x}  + f_t(\boldsymbol{w}) - \boldsymbol{w}^\top \mathbb{E}_{\boldsymbol{w}}{[\boldsymbol{w}f_t(\boldsymbol{w})]} \big)^2  ]}  \\
  & \le 2 c^2  \mathbb{E}_{\boldsymbol{w}} { [ (\boldsymbol{w}^\top ( \boldsymbol{x}  - \mathbb{E}_{\boldsymbol{w}}{[\boldsymbol{w}f_t(\boldsymbol{w})]}   ) )^2 ]} + 2  c^2 \mathbb{E}_{\boldsymbol{w}} { [  f_t(\boldsymbol{w})^2]} \\
  & = 2 c^2  \| \boldsymbol{x}  - \mathbb{E}_{\boldsymbol{w}}{[\boldsymbol{w}f_t(\boldsymbol{w})]}   \|_2^2 + 2  c^2  \mathbb{E}_{\boldsymbol{w}} { [  f_t(\boldsymbol{w})^2]}  \\
  & \le 4 c^2   \|\boldsymbol{x} \|_2^2 + 4  c^2  \|\mathbb{E}_{\boldsymbol{w}}[\boldsymbol{w}f_t(\boldsymbol{w})] \|_2^2 +  2  c^2  \mathbb{E}_{\boldsymbol{w}} { [  f_t(\boldsymbol{w})^2]} 
\end{align}
From Lemma~\ref{BoundedMean}, we know  $\|\mathbb{E}_{\boldsymbol{w}}[\boldsymbol{w}f_t(\boldsymbol{w})] \|_2^2 \le \mathbb{E}_{\boldsymbol{w}} { [  f_t(\boldsymbol{w})^2]} $ is bounded, together with $\| \boldsymbol{x}\|_2 < \infty$,it follows that $ \mathbb{E}_{\boldsymbol{w}} { [  f_{t+1} (\boldsymbol{w}) )^2  ] } =  \mathbb{E}_{\boldsymbol{w}} { [  h (g_{t+1} (\boldsymbol{w}) )^2  ] } < \infty$. Thus, $f_{t+1} \in \mathcal{F}$.

\end{proof}

\begin{Lemma}
\label{KeyFonestep}
For a convex function $ \phi_\lambda (\cdot)$, 
 denote $h(\cdot)$ as the proximal operator of $ \phi_\lambda (\cdot)$, i.e., $h(z) = \argmin _{x} {\frac{1}{2}(x-z)^2 +  \phi_\lambda (x)} $.   Suppose $ |h(x)| \le c |x| $ (or $ |h(x)| \le c  $), $0 < c < \infty$ (e.g., soft thresholding function).  Given a bouned $\boldsymbol{x} \in \mathcal{R}^d$,  set function $g_{t+1}(\boldsymbol{w}) = \boldsymbol{w}^\top\boldsymbol{x}  + f_t(\boldsymbol{w}) - \boldsymbol{w}^\top \mathbb{E}_{\boldsymbol{w}}{[\boldsymbol{w}f_t(\boldsymbol{w})]} $ with $f_t \in \mathcal{L}_2$ and $\boldsymbol{w} \sim \mathcal{N}(\boldsymbol{0},\boldsymbol{I}_d)$ (or $f_t \in \overline{\mathcal{L}}_2$ and $\boldsymbol{w} \sim Uni[\sqrt{d}\mathbb{S}^{d-1}]$). Set $f_{t+1} = h \circ g_{t+1}$.
Denote $ Q(f)=L(f)+ \mathbb{E}_{\boldsymbol{w}} [ \phi_\lambda (f(\boldsymbol{w}))]  $ with  $L(f):= \frac{1}{2} \|\boldsymbol{x} - \mathbb{E}_{\boldsymbol{w}}[\boldsymbol{w}f(\boldsymbol{w})] \|_{2}^2   $,  for $\forall f^* \in \mathcal{F}$ with $\mathcal{F}=\mathcal{L}_2$ or $\mathcal{F}=\overline{\mathcal{L}}_2$,  we have
\begin{align}
     Q(f_{t+1}) &  \le  Q(f^*)  \!+\! \frac{1}{2}\mathbb{E}_{\boldsymbol{w}} {[ (f_{t}(\boldsymbol{w})  \! - \!f^*(\boldsymbol{w}))^2  ]} \!-\! \frac{1}{2}\mathbb{E}_{\boldsymbol{w}} {[ (f_{t\!+\!1}(\boldsymbol{w})  \! - \!f^*(\boldsymbol{w}))^2  ]} \nonumber
   \\ & -  \frac{1}{2}\|\mathbb{E}_{\boldsymbol{w}} { [ \boldsymbol{w} \big(f_{t}(\boldsymbol{w}) \!-\!f^*(\boldsymbol{w})\big) ]  }\|_2^2 
\end{align}
\end{Lemma}
\begin{proof}
From Lemma~\ref{FLineq}, we know that 
\begin{align}
    L(f_{t+1}) \! & = \! L(f^*) \!\!+\!\! \mathbb{E}_{\boldsymbol{w}} {[ \boldsymbol{w}^\top (\mathbb{E}_{\boldsymbol{w}}[\boldsymbol{w}f_t(\boldsymbol{w})] \!\!-\!\!\boldsymbol{x}  ) \big(f_{t+1}(\boldsymbol{w}) \! - \!f^*(\boldsymbol{w}) \big) ]  }  \!\!+\!\!   \frac{1}{2}\|\mathbb{E}_{\boldsymbol{w}} { [ \boldsymbol{w} \big(f_{t+1}(\boldsymbol{w}) \!-\!f_t(\boldsymbol{w})\big) ]  }\|_2^2  \nonumber \\
    &  -\frac{1}{2}\|\mathbb{E}_{\boldsymbol{w}} { [ \boldsymbol{w} \big(f_{t}(\boldsymbol{w}) \!-\!f^*(\boldsymbol{w})\big) ]  }\|_2^2
\end{align}
Together with Lemma~\ref{FLineqPhi}, it follows that 
\begin{align}
  &  Q(f_{t+1}) \\ & \le \! Q(f^*) \!\! -\!\! \mathbb{E}_{\boldsymbol{w}} { [ \big( g_{t\!+\!1}(\boldsymbol{w}) \!-\! f_{t+1}(\boldsymbol{w})  \big) \big( f^*(\boldsymbol{w}) \!\!-\!\! f_{t\!+\!1} (\boldsymbol{w}) \big )  ]}  \!\!+\!\! \mathbb{E}_{\boldsymbol{w}} {[ \boldsymbol{w}^\top (\mathbb{E}_{\boldsymbol{w}}[\boldsymbol{w}f_t(\boldsymbol{w})] \!\!-\!\!\boldsymbol{x}  ) \big(f_{t+1}(\boldsymbol{w}) \! \! -\! \!f^*(\boldsymbol{w}) \big) ]  }    \nonumber \\
    & +  \frac{1}{2}\|\mathbb{E}_{\boldsymbol{w}} { [ \boldsymbol{w} \big(f_{t+1}(\boldsymbol{w}) \!-\!f_t(\boldsymbol{w})\big) ]  }\|_2^2  -\frac{1}{2}\|\mathbb{E}_{\boldsymbol{w}} { [ \boldsymbol{w} \big(f_{t}(\boldsymbol{w}) \!-\!f^*(\boldsymbol{w})\big) ]  }\|_2^2 \\
    & =  \! Q(f^*)   \!\!+\!\! \mathbb{E}_{\boldsymbol{w}} {[ \big(  g_{t\!+\!1}(\boldsymbol{w}) \!-\! f_{t+1}(\boldsymbol{w}) +   \boldsymbol{w}^\top (\mathbb{E}_{\boldsymbol{w}}[\boldsymbol{w}f_t(\boldsymbol{w})] \!\!-\!\!\boldsymbol{x}  ) \big) \big(f_{t+1}(\boldsymbol{w}) \! \! -\! \!f^*(\boldsymbol{w}) \big) ]  }    \nonumber \\
    & +  \frac{1}{2}\|\mathbb{E}_{\boldsymbol{w}} { [ \boldsymbol{w} \big(f_{t+1}(\boldsymbol{w}) \!-\!f_t(\boldsymbol{w})\big) ]  }\|_2^2  -\frac{1}{2}\|\mathbb{E}_{\boldsymbol{w}} { [ \boldsymbol{w} \big(f_{t}(\boldsymbol{w}) \!-\!f^*(\boldsymbol{w})\big) ]  }\|_2^2 \\ 
    & = \! Q(f^*)   \!\!+\!\! \mathbb{E}_{\boldsymbol{w}} {[ \big(  f_{t}(\boldsymbol{w}) \!-\! f_{t+1}(\boldsymbol{w}) \big) \big(f_{t+1}(\boldsymbol{w}) \! \! -\! \!f^*(\boldsymbol{w}) \big) ]  }    \nonumber \\
     & +  \frac{1}{2}\|\mathbb{E}_{\boldsymbol{w}} { [ \boldsymbol{w} \big(f_{t+1}(\boldsymbol{w}) \!-\!f_t(\boldsymbol{w})\big) ]  }\|_2^2  -\frac{1}{2}\|\mathbb{E}_{\boldsymbol{w}} { [ \boldsymbol{w} \big(f_{t}(\boldsymbol{w}) \!-\!f^*(\boldsymbol{w})\big) ]  }\|_2^2  \label{FQinner}
\end{align}
Note that
\begin{align}
  & \mathbb{E}_{\boldsymbol{w}} {[ \big(  f_{t}(\boldsymbol{w}) \!-\! f_{t+1}(\boldsymbol{w}) \big) \big(f_{t+1}(\boldsymbol{w}) \! \! -\! \!f^*(\boldsymbol{w}) \big) ]  } \nonumber \\ & =      \mathbb{E}_{\boldsymbol{w}} {[ \big(  f_{t}(\boldsymbol{w}) \!-\! f_{t+1}(\boldsymbol{w}) \big) \big(f_{t+1}(\boldsymbol{w})   \! \! -\! \! f_t(\boldsymbol{w})   \! \! +\! \! f_t(\boldsymbol{w})   \! \! -\! \!f^*(\boldsymbol{w}) \big) ]  }  \\
  & = \mathbb{E}_{\boldsymbol{w}} {[ \big(  f_{t}(\boldsymbol{w}) \!-\! f_{t+1}(\boldsymbol{w}) \big) \big(f_{t}(\boldsymbol{w}) \! \! -\! \!f^*(\boldsymbol{w}) \big) ]  } - \mathbb{E}_{\boldsymbol{w}} {[ \big(  f_{t}(\boldsymbol{w}) \!-\! f_{t+1}(\boldsymbol{w}) \big)^2 ]}
\end{align}
Also note that $ab=\frac{a^2+b^2-(a-b)^2}{2}$, it follows that
\begin{align}
    \big(  f_{t}(\boldsymbol{w}) \!-\! f_{t+1}(\boldsymbol{w}) \big) \big(f_{t}(\boldsymbol{w}) \!-\! f^*(\boldsymbol{w}) \big) = \frac{ ( f_{t}(\boldsymbol{w}) \!-\! f_{t+1}(\boldsymbol{w}))^2 + (f_{t}(\boldsymbol{w})  \! - \!f^*(\boldsymbol{w}))^2 \!-\! (f_{t\!+\!1}(\boldsymbol{w}) \!-\! f^*(\boldsymbol{w})  )^2   }{2}
\end{align}
It follows that
\begin{align}
     & \mathbb{E}_{\boldsymbol{w}} {[ \big(  f_{t}(\boldsymbol{w}) \!-\! f_{t+1}(\boldsymbol{w}) \big) \big(f_{t+1}(\boldsymbol{w}) \! \! -\! \!f^*(\boldsymbol{w}) \big) ]  } \nonumber \\ & =  
     \frac{1}{2}\mathbb{E}_{\boldsymbol{w}} {[ (f_{t}(\boldsymbol{w})  \! - \!f^*(\boldsymbol{w}))^2  ]} - \frac{1}{2}\mathbb{E}_{\boldsymbol{w}} {[ (f_{t\!+\!1}(\boldsymbol{w})  \! - \!f^*(\boldsymbol{w}))^2  ]} - \frac{1}{2}\mathbb{E}_{\boldsymbol{w}} {[ (f_{t\!+\!1}(\boldsymbol{w})  \! - \!f_t(\boldsymbol{w}))^2  ]}
     \label{Finner}
\end{align}
Plug Eq.(\ref{Finner}) into Eq.(\ref{FQinner}), we can obtain that
\begin{align}
    Q(f_{t+1}) & \le Q(f^*)  + \frac{1}{2}\mathbb{E}_{\boldsymbol{w}} {[ (f_{t}(\boldsymbol{w})  \! - \!f^*(\boldsymbol{w}))^2  ]} - \frac{1}{2}\mathbb{E}_{\boldsymbol{w}} {[ (f_{t\!+\!1}(\boldsymbol{w})  \! - \!f^*(\boldsymbol{w}))^2  ]} - \frac{1}{2}\mathbb{E}_{\boldsymbol{w}} {[ (f_{t\!+\!1}(\boldsymbol{w})  \! - \!f_t(\boldsymbol{w}))^2  ]} \nonumber \\
    & +  \frac{1}{2}\|\mathbb{E}_{\boldsymbol{w}} { [ \boldsymbol{w} \big(f_{t+1}(\boldsymbol{w}) \!-\!f_t(\boldsymbol{w})\big) ]  }\|_2^2  -\frac{1}{2}\|\mathbb{E}_{\boldsymbol{w}} { [ \boldsymbol{w} \big(f_{t}(\boldsymbol{w}) \!-\!f^*(\boldsymbol{w})\big) ]  }\|_2^2 
\end{align}
From Lemma~\ref{BoundedMean}, we know $\|\mathbb{E}_{\boldsymbol{w}} { [ \boldsymbol{w} \big(f_{t+1}(\boldsymbol{w}) \!-\!f_t(\boldsymbol{w})\big) ]  }\|_2^2  \le \mathbb{E}_{\boldsymbol{w}} {[ (f_{t\!+\!1}(\boldsymbol{w})  \! - \!f_t(\boldsymbol{w}))^2  ]}  $. It follows that
\begin{align}
     Q(f_{t+1}) & \le  Q(f^*)  \!+\! \frac{1}{2}\mathbb{E}_{\boldsymbol{w}} {[ (f_{t}(\boldsymbol{w})  \! - \!f^*(\boldsymbol{w}))^2  ]} \!-\! \frac{1}{2}\mathbb{E}_{\boldsymbol{w}} {[ (f_{t\!+\!1}(\boldsymbol{w})  \! - \!f^*(\boldsymbol{w}))^2  ]}
   \\ \nonumber & -  \frac{1}{2}\|\mathbb{E}_{\boldsymbol{w}} { [ \boldsymbol{w} \big(f_{t}(\boldsymbol{w}) \!-\!f^*(\boldsymbol{w})\big) ]  }\|_2^2 
\end{align}
\end{proof}

\begin{Lemma}\label{FMonotonic}
(Strictly Monotonic Descent (a.s.))
Following the same condition of Lemma~\ref{KeyFonestep},  we have
\begin{align}
    Q(f_{t+1})  \le  Q(f_t)  \!-\! \frac{1}{2}\mathbb{E}_{\boldsymbol{w}} {[ (f_{t\!+\!1}(\boldsymbol{w})  \! - \!f_t(\boldsymbol{w}))^2  ]}
\end{align}
\end{Lemma}
\begin{proof}
It follows directly from Lemma~\ref{KeyFonestep} by setting $f^*=f_t$. 
\end{proof}

\begin{theorem*}
For a convex function $ \phi_\lambda (\cdot)$, 
 denote $h(\cdot)$ as the proximal operator of $ \phi_\lambda (\cdot)$, i.e., $h(z) = \argmin _{x} {\frac{1}{2}(x-z)^2 +  \phi_\lambda (x)} $.   Suppose $ |h(x)| \le c |x| $ (or $ |h(x)| \le c  $), $0<c<\infty$.  Given a bouned $\boldsymbol{x} \in \mathcal{R}^d$,  set function $g_{t+1}(\boldsymbol{w}) = \boldsymbol{w}^\top\boldsymbol{x}  + f_t(\boldsymbol{w}) - \boldsymbol{w}^\top \mathbb{E}_{\boldsymbol{w}}{[\boldsymbol{w}f_t(\boldsymbol{w})]} $ with   $\boldsymbol{w} \sim \mathcal{N}(\boldsymbol{0},\boldsymbol{I}_d)$ (or  $\boldsymbol{w} \sim Uni[\sqrt{d}\mathbb{S}^{d-1}]$). Set $f_{t+1} = h \circ g_{t+1}$ and $f_0  \in \mathcal{F}$ with $\mathcal{F}=\mathcal{L}_2$ or $\mathcal{F}=\overline{\mathcal{L}}_2$ (e.g., $f_0=0$). 
Denote $ Q(f)=L(f)+ \mathbb{E}_{\boldsymbol{w}} [ \phi_\lambda (f(\boldsymbol{w}))]  $ with  $L(f):= \frac{1}{2} \|\boldsymbol{x} - \mathbb{E}_{\boldsymbol{w}}[\boldsymbol{w}f(\boldsymbol{w})] \|_{2}^2   $. Denote $f_* \in \mathcal{F}$ as an optimal of $Q(\cdot)$,    we have
\begin{align}
   T\big( Q(f_T  ) -  Q(f_*) \big) & \le \frac{1}{2}\mathbb{E}_{\boldsymbol{w}} {[ (f_{0}(\boldsymbol{w})  \! - \!f_*(\boldsymbol{w}))^2  ]} \!-\! \frac{1}{2}\mathbb{E}_{\boldsymbol{w}} {[ (f_{T}(\boldsymbol{w})  \! - \!f_*(\boldsymbol{w}))^2  ]}   \nonumber \\ & -\frac{1}{2}\sum_{t=0}^{T-1}{ \|\mathbb{E}_{\boldsymbol{w}} { [ \boldsymbol{w} \big(f_{t}(\boldsymbol{w}) \!-\!f_*(\boldsymbol{w})\big) ]  }\|_2^2  } -\frac{1}{2} \sum_{t=0}^{T-1}{  { (t+1)\mathbb{E}_{\boldsymbol{w}} {[ (f_{t\!+\!1}(\boldsymbol{w})  \! - \!f_t(\boldsymbol{w}))^2  ]}   }    }
\end{align}

\end{theorem*}
\begin{proof}
From Lemma~\ref{KeyFonestep}, by setting $f^*=f_*$, we can obtain that
\begin{align}
\label{FtinEQ}
     Q(f_{t+1}) & \le  Q(f_*)  \!+\! \frac{1}{2}\mathbb{E}_{\boldsymbol{w}} {[ (f_{t}(\boldsymbol{w})  \! - \!f_*(\boldsymbol{w}))^2  ]} \!-\! \frac{1}{2}\mathbb{E}_{\boldsymbol{w}} {[ (f_{t\!+\!1}(\boldsymbol{w})  \! - \!f_*(\boldsymbol{w}))^2  ]}
   \nonumber \\ & -  \frac{1}{2}\|\mathbb{E}_{\boldsymbol{w}} { [ \boldsymbol{w} \big(f_{t}(\boldsymbol{w}) \!-\!f_*(\boldsymbol{w})\big) ]  }\|_2^2 
\end{align}
Telescope the inequality~(\ref{FtinEQ}) from $t=0$ to $t=T-1$, we can obtain that
\begin{align}
\label{FTeqTmp1}
    \sum_{t=0}^{T-1}{ Q(f_{t+1}) }-TQ(f_*)  & \le  \frac{1}{2}\mathbb{E}_{\boldsymbol{w}} {[ (f_{0}(\boldsymbol{w})  \! - \!f_*(\boldsymbol{w}))^2  ]} \!-\! \frac{1}{2}\mathbb{E}_{\boldsymbol{w}} {[ (f_{T}(\boldsymbol{w})  \! - \!f_*(\boldsymbol{w}))^2  ]}   \nonumber \\ & -\frac{1}{2}\sum_{t=0}^{T-1}{ \|\mathbb{E}_{\boldsymbol{w}} { [ \boldsymbol{w} \big(f_{t}(\boldsymbol{w}) \!-\!f_*(\boldsymbol{w})\big) ]  }\|_2^2  } 
\end{align}

In addition, from Lemma~\ref{FMonotonic}, we can obtain that
\begin{align}
   Q(f_{T}) \le  Q(f_{t}) -\frac{1}{2}\sum_{i=t}^{T-1} {\mathbb{E}_{\boldsymbol{w}} {[ (f_{i\!+\!1}(\boldsymbol{w})  \! - \!f_i(\boldsymbol{w}))^2  ]} } 
\end{align}
It follows that 
\begin{align}
    TQ(f_T  ) - T Q(f_*) & \le \sum_{t=0}^{T-1}{ Q(f_{t+1}) }-TQ(f_*) - \frac{1}{2} \sum_{t=0}^{T-1}{ \sum_{i=t}^{T-1} { \mathbb{E}_{\boldsymbol{w}} {[ (f_{i\!+\!1}(\boldsymbol{w})  \! - \!f_i(\boldsymbol{w}))^2  ]}   }    }  \\
    & =  \sum_{t=0}^{T-1}{ Q(f_{t+1}) }-TQ(f_*) - \frac{1}{2} \sum_{t=0}^{T-1}{  { (t+1)\mathbb{E}_{\boldsymbol{w}} {[ (f_{t\!+\!1}(\boldsymbol{w})  \! - \!f_t(\boldsymbol{w}))^2  ]}   }    }  \label{TQtmp}
\end{align}
Plug inequality~(\ref{FTeqTmp1}) into inequality~(\ref{TQtmp}), we can obtain that
\begin{align}
     TQ(f_T  ) - T Q(f_*) & \le \frac{1}{2}\mathbb{E}_{\boldsymbol{w}} {[ (f_{0}(\boldsymbol{w})  \! - \!f_*(\boldsymbol{w}))^2  ]} \!-\! \frac{1}{2}\mathbb{E}_{\boldsymbol{w}} {[ (f_{T}(\boldsymbol{w})  \! - \!f_*(\boldsymbol{w}))^2  ]}   \nonumber \\ & -\frac{1}{2}\sum_{t=0}^{T-1}{ \|\mathbb{E}_{\boldsymbol{w}} { [ \boldsymbol{w} \big(f_{t}(\boldsymbol{w}) \!-\!f_*(\boldsymbol{w})\big) ]  }\|_2^2  } -\frac{1}{2} \sum_{t=0}^{T-1}{  { (t+1)\mathbb{E}_{\boldsymbol{w}} {[ (f_{t\!+\!1}(\boldsymbol{w})  \! - \!f_t(\boldsymbol{w}))^2  ]}   }    }
\end{align}
\end{proof}

\section{Proof of Theorem~\ref{NonConvexF}}

\textbf{Non-convex $\phi$-regularization:}
\begin{theorem*}
For a (non-convex) regularization function $ \phi_\lambda (\cdot)$, 
 denote $h(\cdot)$ as the proximal operator of $ \phi_\lambda (\cdot)$, i.e., $h(z) = \argmin _{x} {\frac{1}{2}(x-z)^2 +  \phi_\lambda (x)} $.   Suppose $ |h(x)| \le c|x| $ (or $ |h(x)| \le c  $), $0<c<\infty$ (e.g.,  hard thresholding function).  Given a bouned $\boldsymbol{x} \in \mathcal{R}^d$,  set function $g_{t+1}(\boldsymbol{w}) = \boldsymbol{w}^\top\boldsymbol{x}  + f_t(\boldsymbol{w}) - \boldsymbol{w}^\top \mathbb{E}_{\boldsymbol{w}}{[\boldsymbol{w}f_t(\boldsymbol{w})]} $ with $f_t \in \mathcal{L}_2$ and $\boldsymbol{w} \sim \mathcal{N}(\boldsymbol{0},\boldsymbol{I}_d)$ (or $f_t \in \overline{\mathcal{L}}_2$,  $\boldsymbol{w} \sim Uni[\sqrt{d}\mathbb{S}^{d-1}]$). Set $f_{t+1} = h \circ g_{t+1}$.
Denote $ Q(f)=L(f)+ \mathbb{E}_{\boldsymbol{w}} [ \phi_\lambda (f(\boldsymbol{w}))]  $ with  $L(f):= \frac{1}{2} \|\boldsymbol{x} - \mathbb{E}_{\boldsymbol{w}}[\boldsymbol{w}f(\boldsymbol{w})] \|_{2}^2   $,  we have
\begin{align}
   Q(f_{t\!+\!1})  \! & \le\!  Q(f_t) - \frac{1}{2}\mathbb{E}_{\boldsymbol{w}}[ \left.( f_{t\!+\!1}(\boldsymbol{w})\! -\!f_t(\boldsymbol{w}) - \boldsymbol{w}^\top \mathbb{E}_{\boldsymbol{w}}[\boldsymbol{w} (f_{t\!+\!1}(\boldsymbol{w})\! -\!f_t(\boldsymbol{w})) ] \right.)^2 ]  \le  Q(f_{t})
\end{align}
\end{theorem*}

\begin{proof}

From Lemma~\ref{FL},  we know that
\begin{align}
\label{FnonInFt}
    L(f_{t\!+\!1}) \!=\!  L(f_t) + \mathbb{E}_{\boldsymbol{w}} {[ \boldsymbol{w}^\top (\mathbb{E}_{\boldsymbol{w}}[\boldsymbol{w}f_t(\boldsymbol{w})]  \!-\!\boldsymbol{x}  ) \big(f_{t\!+\!1}(\boldsymbol{w})\! -\!f_t(\boldsymbol{w}) \big) ]  } + \frac{1}{2}\|\mathbb{E}_{\boldsymbol{w}} { [ \boldsymbol{w} \big(f_{t\!+\!1}(\boldsymbol{w})\!-\!f_t(\boldsymbol{w})\big) ]  }\|_2^2
\end{align}
Let $g_{t+1}(\boldsymbol{w}) = \boldsymbol{w}^\top\boldsymbol{x}  + f_t(\boldsymbol{w}) - \boldsymbol{w}^\top \mathbb{E}_{\boldsymbol{w}}{[\boldsymbol{w}f_t(\boldsymbol{w})]}  $, together with Eq.(\ref{FnonInFt}), we can obtain that
\begin{align}
\label{FnonInFt2}
    L(f_{t\!+\!1}) \!=\!  L(f_t) + \mathbb{E}_{\boldsymbol{w}} {[ \big( f_t(\boldsymbol{w}) -g_{t+1}(\boldsymbol{w})  \big) \big(f_{t\!+\!1}(\boldsymbol{w})\! -\!f_t(\boldsymbol{w}) \big) ]  } + \frac{1}{2}\|\mathbb{E}_{\boldsymbol{w}} { [ \boldsymbol{w} \big(f_{t\!+\!1}(\boldsymbol{w})\!-\!f_t(\boldsymbol{w})\big) ]  }\|_2^2
\end{align}
Note that $ab=\frac{(a+b)^2-a^2-b^2}{2}$, it follows that
\begin{align}
    \big( f_t(\boldsymbol{w}) \!-\!g_{t\!+\!1}(\boldsymbol{w})  \big) \big(f_{t\!+\!1}(\boldsymbol{w})\! -\!f_t(\boldsymbol{w}) \big) \!=\!  \frac{ \big( f_{t\!+\!1}(\boldsymbol{w}) \!-\!g_{t\!+\!1}(\boldsymbol{w}) \big)^2 \!\!-\!\! \big( f_t(\boldsymbol{w}) \!-\!g_{t\!+\!1}(\boldsymbol{w})  \big)^2 \!\!-\!\!  \big(f_{t\!+\!1}(\boldsymbol{w})\! -\!f_t(\boldsymbol{w}) \big)^2 }{2}
\end{align}
Since $f_{t+1}  = h \circ g_{t+1}$ is the solution of the proximal problem,  \\ i.e., $f_{t+1}(\boldsymbol{w})  = \argmin _{x} {  \frac{(x-g_{t+1}(\boldsymbol{w}))^2}{2} + \phi_\lambda(x)  }$, we know that 
\begin{align}
 \frac{\big( f_{t\!+\!1}(\boldsymbol{w}) \!-\!g_{t\!+\!1}(\boldsymbol{w}) \big)^2 }{2}     - \frac{\big( f_t(\boldsymbol{w}) \!-\!g_{t\!+\!1}(\boldsymbol{w})  \big)^2}{2}  \le \phi_\lambda(f_t(\boldsymbol{w})) -\phi_\lambda(f_{t\!+\!1}(\boldsymbol{w})) 
\end{align}
It follows that
\begin{align}
 &  \mathbb{E}_{\boldsymbol{w}} {[ \big( f_t(\boldsymbol{w}) -g_{t+1}(\boldsymbol{w})  \big) \big(f_{t\!+\!1}(\boldsymbol{w})\! -\!f_t(\boldsymbol{w}) \big) ]  }  \nonumber \\
 & = \mathbb{E}_{\boldsymbol{w}} {\left.[ \frac{ \big( f_{t\!+\!1}(\boldsymbol{w}) \!-\!g_{t\!+\!1}(\boldsymbol{w}) \big)^2 \!\!-\!\! \big( f_t(\boldsymbol{w}) \!-\!g_{t\!+\!1}(\boldsymbol{w})  \big)^2 \!\!-\!\!  \big(f_{t\!+\!1}(\boldsymbol{w})\! -\!f_t(\boldsymbol{w}) \big)^2 }{2} \right.]}  \\
 & \le  \mathbb{E}_{\boldsymbol{w}} {[\phi_\lambda(f_t(\boldsymbol{w}))  ]} -  \mathbb{E}_{\boldsymbol{w}} {[\phi_\lambda(f_{t+1}(\boldsymbol{w}))  ]} - \frac{1}{2}\mathbb{E}_{\boldsymbol{w}} {[ \big(f_{t\!+\!1}(\boldsymbol{w})\! -\!f_t(\boldsymbol{w}) \big)^2 ]} \label{NonInnerEqE}
\end{align}
Plug inequality~(\ref{NonInnerEqE}) into Eq.(\ref{FnonInFt2}), we can achieve that
\begin{align}
     L(f_{t\!+\!1}) + \mathbb{E}_{\boldsymbol{w}} {[\phi_\lambda(f_{t+1}(\boldsymbol{w}))  ]}  \! & \le\!  L(f_t) +\mathbb{E}_{\boldsymbol{w}} {[\phi_\lambda(f_t(\boldsymbol{w}))  ]} - \frac{1}{2}\mathbb{E}_{\boldsymbol{w}} {[ \big(f_{t\!+\!1}(\boldsymbol{w})\! -\!f_t(\boldsymbol{w}) \big)^2 ]} \nonumber \\ & + \frac{1}{2}\|\mathbb{E}_{\boldsymbol{w}} { [ \boldsymbol{w} \big(f_{t\!+\!1}(\boldsymbol{w})\!-\!f_t(\boldsymbol{w})\big) ]  }\|_2^2
\end{align}
From Lemma~\ref{BoundedMean}, we know that 
\begin{align}
  &  \mathbb{E}_{\boldsymbol{w}} {[ \big(f_{t\!+\!1}(\boldsymbol{w})\! -\!f_t(\boldsymbol{w}) \big)^2 ]}- \|\mathbb{E}_{\boldsymbol{w}} { [ \boldsymbol{w} \big(f_{t\!+\!1}(\boldsymbol{w})\!-\!f_t(\boldsymbol{w})\big) ]  }\|_2^2 \nonumber \\ &  = \mathbb{E}_{\boldsymbol{w}}[ \left( f_{t\!+\!1}(\boldsymbol{w})\! -\!f_t(\boldsymbol{w}) - \boldsymbol{w}^\top \mathbb{E}_{\boldsymbol{w}}[\boldsymbol{w} (f_{t\!+\!1}(\boldsymbol{w})\! -\!f_t(\boldsymbol{w})) ] \right)^2 ] 
\end{align}
It follows that
\begin{align}
  Q(f_{t\!+\!1})  \! & \le\!  Q(f_t) - \frac{1}{2}\mathbb{E}_{\boldsymbol{w}}[ \left.( f_{t\!+\!1}(\boldsymbol{w})\! -\!f_t(\boldsymbol{w}) - \boldsymbol{w}^\top \mathbb{E}_{\boldsymbol{w}}[\boldsymbol{w} (f_{t\!+\!1}(\boldsymbol{w})\! -\!f_t(\boldsymbol{w})) ] \right.)^2 ]  \le  Q(f_{t})
\end{align}

\end{proof}


\section{Proof of Theorem~\ref{ANonConvexF}}

To prove the Theorem~\ref{ANonConvexF}, we first show some useful Lemmas.
\begin{Lemma}
\label{FiniteSquareIneq}
Suppose $\frac{1}{N}\boldsymbol{W}\boldsymbol{W}^\top = \boldsymbol{I}_d$, for any bounded $ \boldsymbol{y} \in \mathcal{R}^N$,  we have $\frac{1}{N} \| \boldsymbol{y}\|_2^2 - \|\frac{1}{N}\boldsymbol{W}\boldsymbol{y} \|_2^2 = \frac{1}{N}\|\boldsymbol{y}-\frac{1}{N}\boldsymbol{W}^\top\boldsymbol{W}\boldsymbol{y} \|_2^2 \ge 0$.
\end{Lemma}
\begin{proof}
\begin{align}
   \frac{1}{N} \| \boldsymbol{y}\|_2^2 - \|\frac{1}{N}\boldsymbol{W}\boldsymbol{y} \|_2^2 & =   \| \boldsymbol{y}\|_2^2 - 2 \|\frac{1}{N}\boldsymbol{W}\boldsymbol{y} \|_2^2 +   \|\frac{1}{N}\boldsymbol{W}\boldsymbol{y} \|_2^2 \\
    & = \frac{1}{N}\|\boldsymbol{y}\|_2^2  - \frac{2}{N^2}\boldsymbol{y}^\top\boldsymbol{W}^\top\boldsymbol{W}\boldsymbol{y} + \frac{1}{N^2}\boldsymbol{y}^\top\boldsymbol{W}^\top\boldsymbol{W}\boldsymbol{y} \\
    & = \frac{1}{N}\|\boldsymbol{y}\|_2^2  - \frac{2}{N^2}\boldsymbol{y}^\top\boldsymbol{W}^\top\boldsymbol{W}\boldsymbol{y} + \frac{1}{N^2}\boldsymbol{y}^\top\boldsymbol{W}^\top \frac{1}{N} \boldsymbol{W}\boldsymbol{W}^\top \boldsymbol{W}\boldsymbol{y} \\
    & = \frac{1}{N}\|\boldsymbol{y}-\frac{1}{N}\boldsymbol{W}^\top\boldsymbol{W}\boldsymbol{y} \|_2^2 \ge 0
\end{align}
\end{proof}

\begin{Lemma}
\label{FiniteLeq}
Denote $L(\boldsymbol{y}):=\frac{1}{2}\|\boldsymbol{x}-\frac{1}{N}\boldsymbol{W}\boldsymbol{y} \|_2^2$. 
For  $\forall \boldsymbol{y},\boldsymbol{z} \in \mathcal{R}^N $, we have $ L(\boldsymbol{z}) =  L(\boldsymbol{y}) + \left.< \frac{1}{N^2}\boldsymbol{W}^\top\boldsymbol{W}\boldsymbol{y}-\frac{1}{N}\boldsymbol{W}^\top\boldsymbol{x}, \boldsymbol{z}-\boldsymbol{y}  \right.>    + \frac{1}{2}\| \frac{1}{N}\boldsymbol{W} (\boldsymbol{z}-\boldsymbol{y})  \|_2^2  $
\end{Lemma}

\begin{proof}

\begin{align}
    \frac{1}{2}\|\boldsymbol{x}-\frac{1}{N}\boldsymbol{W}\boldsymbol{z} \|_2^2 & = \frac{1}{2}\|\boldsymbol{x}-\frac{1}{N}\boldsymbol{W}\boldsymbol{y} + \frac{1}{N}\boldsymbol{W}\boldsymbol{y} -\frac{1}{N}\boldsymbol{W}\boldsymbol{z} \|_2^2 \\
    & = L(\boldsymbol{y}) + \left.< \frac{1}{N}\boldsymbol{W}\boldsymbol{y}-\boldsymbol{x}, \frac{1}{N}\boldsymbol{W}(\boldsymbol{z}-\boldsymbol{y})  \right.>    + \frac{1}{2}\| \frac{1}{N}\boldsymbol{W} (\boldsymbol{z}-\boldsymbol{y})  \|_2^2 \\
    & =L(\boldsymbol{y}) + \left.< \frac{1}{N^2}\boldsymbol{W}^\top\boldsymbol{W}\boldsymbol{y}-\frac{1}{N}\boldsymbol{W}^\top\boldsymbol{x}, \boldsymbol{z}-\boldsymbol{y}  \right.>    + \frac{1}{2}\| \frac{1}{N}\boldsymbol{W} (\boldsymbol{z}-\boldsymbol{y})  \|_2^2 
\end{align}

\end{proof}

\begin{theorem*}(Monotonic Descent)
For a function $ \phi_\lambda (\cdot)$, 
 denote $h(\cdot)$ as the proximal operator of $ \phi_\lambda (\cdot)$.    Given a bouned $\boldsymbol{x} \in \mathcal{R}^d$,  set $  {\boldsymbol{y}_{t+1}} = h \big( \boldsymbol{W}^\top \boldsymbol{x} + (\boldsymbol{I} - \frac{1}{N} \boldsymbol{W}^\top \boldsymbol{W}) \boldsymbol{y}_t   \big)$ with $\frac{1}{N}\boldsymbol{W}\boldsymbol{W}^\top=\boldsymbol{I}_d$.
Denote $ \widehat{Q}(\boldsymbol{y}):=  \frac{1}{2}  \|\boldsymbol{x} -  \frac{1}{N}\boldsymbol{W}{\boldsymbol{y} } \|_2^2 + \frac{1}{N} \phi_\lambda(\boldsymbol{y})  $. For $t\ge 0$,    we have
\begin{align}
     \widehat{Q}(\boldsymbol{y}_{t\!+\!1})  \! & \le\!  \widehat{Q}(\boldsymbol{y}_t) - \frac{1}{2N}\|(\boldsymbol{I}_d- \frac{1}{N}\boldsymbol{W}^\top\boldsymbol{W}) (\boldsymbol{y}_{t+1} -\boldsymbol{y}_t ) \|_2^2 \le \widehat{Q}(\boldsymbol{y}_t) 
\end{align}
\end{theorem*}

\begin{proof}
Denote $L(\boldsymbol{y}):=\frac{1}{2}\|\boldsymbol{x}-\frac{1}{N}\boldsymbol{W}\boldsymbol{y} \|_2^2$,  from Lemma~\ref{FiniteLeq}, we know that
\begin{align}
\label{LeqNonConvFinite}
    L(\boldsymbol{y}_{t+1}) = L(\boldsymbol{y}_t) + \left.<\! \frac{1}{N^2}\boldsymbol{W}^\top\boldsymbol{W}\boldsymbol{y}_{t}-\frac{1}{N}\boldsymbol{W}^\top\boldsymbol{x}, \boldsymbol{y}_{t\!+\!1}-\boldsymbol{y}_t  \!\right.>    + \frac{1}{2}\| \frac{1}{N}\boldsymbol{W} (\boldsymbol{y}_{t\!+\!1}-\boldsymbol{y}_t)  \|_2^2 
\end{align}
Let $\boldsymbol{a}_{t+1}=  \boldsymbol{W}^\top \boldsymbol{x} + (\boldsymbol{I} - \frac{1}{N} \boldsymbol{W}^\top \boldsymbol{W}) \boldsymbol{y}_t $. Together with Eq.(\ref{LeqNonConvFinite}), we can obtain that
\begin{align}
     L(\boldsymbol{y}_{t+1}) & = L(\boldsymbol{y}_t) + \left.<\! \frac{1}{N^2}\boldsymbol{W}^\top\boldsymbol{W}\boldsymbol{y}_{t}-\frac{1}{N}\boldsymbol{W}^\top\boldsymbol{x}, \boldsymbol{y}_{t\!+\!1}-\boldsymbol{y}_t  \!\right.>    + \frac{1}{2}\| \frac{1}{N}\boldsymbol{W} (\boldsymbol{y}_{t\!+\!1}-\boldsymbol{y}_t)  \|_2^2  \\
     & =  L(\boldsymbol{y}_t) + \frac{1}{N}\left.<\! \boldsymbol{y}_t -\boldsymbol{a}_{t+1} , \boldsymbol{y}_{t\!+\!1}-\boldsymbol{y}_t  \!\right.>    + \frac{1}{2}\| \frac{1}{N}\boldsymbol{W} (\boldsymbol{y}_{t\!+\!1}-\boldsymbol{y}_t)  \|_2^2 \label{NonConvexLeq2}
\end{align}
Note that $\boldsymbol{a}^\top\boldsymbol{b}= \frac{\|\boldsymbol{a} + \boldsymbol{b} \|_2^2 -  \|\boldsymbol{a} \|_2^2 - \|\boldsymbol{b} \|_2^2  }{2}$, it follows that
\begin{align}
\label{NonConvFiniteInnerEq}
    <\! \boldsymbol{y}_t -\boldsymbol{a}_{t+1} , \boldsymbol{y}_{t\!+\!1}-\boldsymbol{y}_t  \!> \; = \frac{\| \boldsymbol{y}_{t+1}-\boldsymbol{a}_{t+1}\|_2^2 - \|\boldsymbol{y}_t -\boldsymbol{a}_{t+1}   \|_2^2 - \| \boldsymbol{y}_{t\!+\!1}-\boldsymbol{y}_t  \|_2^2 }{2}
\end{align}
Since $\boldsymbol{y}_{t+1}=h(\boldsymbol{a}_{t+1})$ is the solution of the proximal problem, \\ i.e., $\boldsymbol{y}_{t+1}= \argmin_{\boldsymbol{y}} {\frac{1}{2}\|\boldsymbol{y} - \boldsymbol{a}_{t+1} \|_2^2 + \phi_\lambda(\boldsymbol{y}) }  $,  we can achieve that 
\begin{align}
    \frac{1}{2}\|\boldsymbol{y}_{t+1} - \boldsymbol{a}_{t+1} \|_2^2 + \phi_\lambda(\boldsymbol{y}_{t+1}) \le \frac{1}{2}\|\boldsymbol{y}_{t} - \boldsymbol{a}_{t+1} \|_2^2 + \phi_\lambda(\boldsymbol{y}_{t})
\end{align}
It can be rewritten as 
\begin{align}
\label{NonConvFiniteProxmialIneq}
      \frac{1}{2}\|\boldsymbol{y}_{t+1} - \boldsymbol{a}_{t+1} \|_2^2 -\frac{1}{2}\|\boldsymbol{y}_{t} - \boldsymbol{a}_{t+1} \|_2^2    \le   \phi_\lambda(\boldsymbol{y}_{t}) - \phi_\lambda(\boldsymbol{y}_{t+1})
\end{align}
Together with Eq.(\ref{NonConvexLeq2}),  Eq.(\ref{NonConvFiniteInnerEq}) and inequality~(\ref{NonConvFiniteProxmialIneq}), it follows that 
\begin{align}
     L(\boldsymbol{y}_{t+1}) + \frac{1}{N}\phi_\lambda(\boldsymbol{y}_{t+1}) \le  L(\boldsymbol{y}_{t}) + \frac{1}{N}\phi_\lambda(\boldsymbol{y}_{t}) -\frac{1}{2N}\| \boldsymbol{y}_{t\!+\!1}-\boldsymbol{y}_t  \|_2^2  + \frac{1}{2}\| \frac{1}{N}\boldsymbol{W} (\boldsymbol{y}_{t\!+\!1}-\boldsymbol{y}_t)  \|_2^2
\end{align}
Together with Lemma~\ref{FiniteSquareIneq}, we can achieve that
\begin{align}
       \widehat{Q}(\boldsymbol{y}_{t\!+\!1})  \! & \le\!  \widehat{Q}(\boldsymbol{y}_t) -\frac{1}{2N}\|(\boldsymbol{I}_d- \frac{1}{N}\boldsymbol{W}^\top\boldsymbol{W}) (\boldsymbol{y}_{t+1} -\boldsymbol{y}_t ) \|_2^2  \le \widehat{Q}(\boldsymbol{y}_t) 
\end{align}


\end{proof}

\section{Proof of Theorem~\ref{AConvexF}}

Before proving   Theorem~\ref{AConvexF}, we first show some useful Lemmas. 

\begin{Lemma}
\label{FiniteConvThreeL}
Denote $L(\boldsymbol{y}):=\frac{1}{2}\|\boldsymbol{x}-\frac{1}{N}\boldsymbol{W}\boldsymbol{y} \|_2^2$. 
For any bounded  $ \boldsymbol{y}_t,\boldsymbol{y}_{t+1}, \boldsymbol{z} \in \mathcal{R}^N $, we have
\begin{align}
   L(\boldsymbol{y}_{t+1})  & = L(\boldsymbol{z}) + \left<\frac{1}{N^2}\boldsymbol{W}^\top\boldsymbol{W}\boldsymbol{y}_t\!-\!\frac{1}{N}\boldsymbol{W}^\top\boldsymbol{x}, \boldsymbol{y}_{t+1}-\boldsymbol{z} \right> + \frac{1}{2}\| \frac{1}{N}\boldsymbol{W} (\boldsymbol{y}_{t+1}-\boldsymbol{y}_t)  \|_2^2  \nonumber \\ &  - \frac{1}{2}\| \frac{1}{N}\boldsymbol{W} (\boldsymbol{z}-\boldsymbol{y}_t)  \|_2^2 
\end{align}
\end{Lemma}
\begin{proof}
Denote $L(\boldsymbol{y}):=\frac{1}{2}\|\boldsymbol{x}-\frac{1}{N}\boldsymbol{W}\boldsymbol{y} \|_2^2$.  From Lemma~\ref{FiniteLeq}, we can achieve that
\begin{align}
  &  L(\boldsymbol{z}) =  L(\boldsymbol{y}_t) + \left.<\! \frac{1}{N^2}\boldsymbol{W}^\top\boldsymbol{W}\boldsymbol{y}_t-\frac{1}{N}\boldsymbol{W}^\top\boldsymbol{x}, \boldsymbol{z}-\boldsymbol{y}_t  \right.>    + \frac{1}{2}\| \frac{1}{N}\boldsymbol{W} (\boldsymbol{z}-\boldsymbol{y}_t)  \|_2^2  \\
 & L(\boldsymbol{y}_{t+1}) =  L(\boldsymbol{y}_t) + \left.<\! \frac{1}{N^2}\boldsymbol{W}^\top\boldsymbol{W}\boldsymbol{y}_t-\frac{1}{N}\boldsymbol{W}^\top\boldsymbol{x}, \boldsymbol{y}_{t+1}-\boldsymbol{y}_t  \! \right.>    + \frac{1}{2}\| \frac{1}{N}\boldsymbol{W} (\boldsymbol{y}_{t+1}-\boldsymbol{y}_t)  \|_2^2 
\end{align}
It follows that
\begin{align}
    L(\boldsymbol{y}_{t+1}) & \!= \! L(\boldsymbol{z}) \!- \!\left<\! \frac{1}{N^2}\boldsymbol{W}^\top\boldsymbol{W}\boldsymbol{y}_t\!-\!\frac{1}{N}\boldsymbol{W}^\top\boldsymbol{x}, \boldsymbol{z}\!-\!\boldsymbol{y}_t  \right> \!+ \!\left<\! \frac{1}{N^2}\boldsymbol{W}^\top\boldsymbol{W}\boldsymbol{y}_t\!-\!\frac{1}{N}\boldsymbol{W}^\top\boldsymbol{x}, \boldsymbol{y}_{t+1}\!-\!\boldsymbol{y}_t  \! \right>  \nonumber \\ & + \frac{1}{2}\| \frac{1}{N}\boldsymbol{W} (\boldsymbol{y}_{t+1}-\boldsymbol{y}_t)  \|_2^2 - \frac{1}{2}\| \frac{1}{N}\boldsymbol{W} (\boldsymbol{z}-\boldsymbol{y}_t)  \|_2^2 \\ 
    & = L(\boldsymbol{z}) + \left<\frac{1}{N^2}\boldsymbol{W}^\top\boldsymbol{W}\boldsymbol{y}_t\!-\!\frac{1}{N}\boldsymbol{W}^\top\boldsymbol{x}, \boldsymbol{y}_{t+1}-\boldsymbol{z} \right> + \frac{1}{2}\| \frac{1}{N}\boldsymbol{W} (\boldsymbol{y}_{t+1}-\boldsymbol{y}_t)  \|_2^2  \nonumber \\ &  - \frac{1}{2}\| \frac{1}{N}\boldsymbol{W} (\boldsymbol{z}-\boldsymbol{y}_t)  \|_2^2 
\end{align}
\end{proof}

\begin{Lemma}
\label{FiniteConvThreeQlemma}
For a convex function  $\phi_\lambda(\cdot)$, let $h(\cdot)$ be the proximal operator w.r.t $\phi_\lambda(\cdot)$. Denote  $ \widehat{Q}(\boldsymbol{y}):=  \frac{1}{2}  \|\boldsymbol{x} -  \frac{1}{N}\boldsymbol{W}{\boldsymbol{y} } \|_2^2 + \frac{1}{N} \phi_\lambda(\boldsymbol{y})  $,  for any bounded $\boldsymbol{y}_t,  \boldsymbol{z} \in \mathcal{R}^N$, set $\boldsymbol{a}_{t+1}=  \boldsymbol{W}^\top \boldsymbol{x} + (\boldsymbol{I} - \frac{1}{N} \boldsymbol{W}^\top \boldsymbol{W}) \boldsymbol{y}_t $ and $\boldsymbol{y}_{t+1}=h(\boldsymbol{a_{t+1}})$, then  we have
\begin{align}
      \widehat{Q}(\boldsymbol{y}_{t+1})  \le \widehat{Q}(\boldsymbol{z})+\frac{1}{2N}\big( \|\boldsymbol{y}_t -\boldsymbol{z}  \|_2^2 -\|\boldsymbol{y}_{t+1} -\boldsymbol{z}  \|_2^2  \big)   - \frac{1}{2}\| \frac{1}{N}\boldsymbol{W} (\boldsymbol{z}-\boldsymbol{y}_t)  \|_2^2 
\end{align}
\end{Lemma}
\begin{proof}

Since $\boldsymbol{y}_{t+1}= \argmin _{\boldsymbol{y} }  \phi_\lambda(\boldsymbol{y}) + \frac{1}{2} \| \boldsymbol{y} - \boldsymbol{a}_{t+1}  \|_{2}^2  $,  we have
\begin{align}
 \boldsymbol{0} \in \partial \phi(\boldsymbol{y}_{t+1}) +  (\boldsymbol{y}_{t+1}  -\boldsymbol{a}_{t+1} )   
 \implies	 (\boldsymbol{a}_{t+1} - \boldsymbol{y}_{t+1} ) \in \partial \phi(\boldsymbol{y}_{t+1}) 
\end{align}
For a convex function  $\phi_\lambda(\boldsymbol{y})$ and subgradient $\boldsymbol{g} \in \partial  \phi_\lambda(\boldsymbol{y})$, we know $\phi_\lambda(\boldsymbol{z}) \ge \phi_\lambda(\boldsymbol{y}) + \left<\boldsymbol{g}, \boldsymbol{z} -\boldsymbol{y} \right> $, it follows that
\begin{align}
    \phi_\lambda(\boldsymbol{z}) \ge   \phi_\lambda(\boldsymbol{y}_{t+1}) + \left<\boldsymbol{a}_{t+1} - \boldsymbol{y}_{t+1} , \boldsymbol{z}-\boldsymbol{y}_{t+1} \right>
\end{align}
Together with Lemma~\ref{FiniteConvThreeL}, we can obtain that
\begin{align}
    L(\boldsymbol{y}_{t+1})+\frac{1}{N}\phi_\lambda(\boldsymbol{y}_{t+1}) & \le L(\boldsymbol{z}) + \frac{1}{N}\phi_\lambda(\boldsymbol{z})- \frac{1}{N}\left<\boldsymbol{a}_{t+1} - \boldsymbol{y}_{t+1} , \boldsymbol{z}-\boldsymbol{y}_{t+1} \right> \nonumber \\
    & + \left<\frac{1}{N^2}\boldsymbol{W}^\top\boldsymbol{W}\boldsymbol{y}_t\!-\!\frac{1}{N}\boldsymbol{W}^\top\boldsymbol{x}, \boldsymbol{y}_{t+1}-\boldsymbol{z} \right> + \frac{1}{2}\| \frac{1}{N}\boldsymbol{W} (\boldsymbol{y}_{t+1}-\boldsymbol{y}_t)  \|_2^2  \nonumber \\ &  - \frac{1}{2}\| \frac{1}{N}\boldsymbol{W} (\boldsymbol{z}-\boldsymbol{y}_t)  \|_2^2 
\end{align}
It follows that
\begin{align}
\label{FiniteConvQinnerInEq}
    \widehat{Q}(\boldsymbol{y}_{t+1}) \le \widehat{Q}(\boldsymbol{z})+\frac{1}{N}\big<\boldsymbol{y}_t-\boldsymbol{y}_{t+1}, \boldsymbol{y}_{t+1} -\boldsymbol{z} \big> + \frac{1}{2}\| \frac{1}{N}\boldsymbol{W} (\boldsymbol{y}_{t+1}-\boldsymbol{y}_t)  \|_2^2   - \frac{1}{2}\| \frac{1}{N}\boldsymbol{W} (\boldsymbol{z}-\boldsymbol{y}_t)  \|_2^2 
\end{align}
Note that $\boldsymbol{a}^\top\boldsymbol{b}= \frac{\|\boldsymbol{a} + \boldsymbol{b} \|_2^2 -  \|\boldsymbol{a} \|_2^2 - \|\boldsymbol{b} \|_2^2  }{2}$, it follows that
\begin{align}
    \left< \boldsymbol{y}_t -\boldsymbol{y}_{t+1} , \boldsymbol{y}_{t+1} -\boldsymbol{z}   \right> = \frac{1}{2}  \|\boldsymbol{y}_t -\boldsymbol{z}  \|_2^2 -\frac{1}{2}\|\boldsymbol{y}_{t+1} -\boldsymbol{z}  \|_2^2 -\frac{1}{2} \|\boldsymbol{y}_t -\boldsymbol{y}_{t+1}  \|_2^2 
\end{align}
Together with inequality~(\ref{FiniteConvQinnerInEq}), we can achieve that 
\begin{align}
     \widehat{Q}(\boldsymbol{y}_{t+1}) & \le \widehat{Q}(\boldsymbol{z})+\frac{1}{2N}\big( \|\boldsymbol{y}_t -\boldsymbol{z}  \|_2^2 -\|\boldsymbol{y}_{t+1} -\boldsymbol{z}  \|_2^2 - \|\boldsymbol{y}_t -\boldsymbol{y}_{t+1}  \|_2^2 \big)+ \frac{1}{2}\| \frac{1}{N}\boldsymbol{W} (\boldsymbol{y}_{t+1}-\boldsymbol{y}_t)  \|_2^2 \nonumber \\ &  - \frac{1}{2}\| \frac{1}{N}\boldsymbol{W} (\boldsymbol{z}-\boldsymbol{y}_t)  \|_2^2 
\end{align}
From Lemma~\ref{FiniteSquareIneq}, we know $\frac{1}{N}\|\boldsymbol{y}_t -\boldsymbol{y}_{t+1}  \|_2^2 \ge \| \frac{1}{N}\boldsymbol{W} (\boldsymbol{y}_{t+1}-\boldsymbol{y}_t)  \|_2^2 $, it follows that 
\begin{align}
      \widehat{Q}(\boldsymbol{y}_{t+1})  \le \widehat{Q}(\boldsymbol{z})+\frac{1}{2N}\big( \|\boldsymbol{y}_t -\boldsymbol{z}  \|_2^2 -\|\boldsymbol{y}_{t+1} -\boldsymbol{z}  \|_2^2  \big)   - \frac{1}{2}\| \frac{1}{N}\boldsymbol{W} (\boldsymbol{z}-\boldsymbol{y}_t)  \|_2^2 
\end{align}
\end{proof}

\begin{Lemma}(Strictly Monotonic Descent)
\label{FiniteConvStrictDescent}
For a convex function  $\phi_\lambda(\cdot)$, let $h(\cdot)$ be the proximal operator w.r.t $\phi_\lambda(\cdot)$. Denote  $ \widehat{Q}(\boldsymbol{y}):=  \frac{1}{2}  \|\boldsymbol{x} -  \frac{1}{N}\boldsymbol{W}{\boldsymbol{y} } \|_2^2 + \frac{1}{N} \phi_\lambda(\boldsymbol{y})  $,  for any bounded $\boldsymbol{y}_t   \in \mathcal{R}^N$, set $\boldsymbol{a}_{t+1}=  \boldsymbol{W}^\top \boldsymbol{x} + (\boldsymbol{I} - \frac{1}{N} \boldsymbol{W}^\top \boldsymbol{W}) \boldsymbol{y}_t $ and $\boldsymbol{y}_{t+1}=h(\boldsymbol{a_{t+1}})$, then  we have
\begin{align}
      \widehat{Q}(\boldsymbol{y}_{t+1})  \le \widehat{Q}(\boldsymbol{y}_t) - \frac{1}{2N}\|\boldsymbol{y}_{t+1} -\boldsymbol{y}_t  \|_2^2  
\end{align}
\end{Lemma}
\begin{proof}
From Lemma~\ref{FiniteConvThreeQlemma}, setting $\boldsymbol{z}=\boldsymbol{y}_t $, we can directly get the result.
\end{proof}

\begin{theorem*}
For a convex function $ \phi_\lambda (\cdot)$, 
 denote $h(\cdot)$ as the proximal operator of $ \phi_\lambda (\cdot)$.     Given a bounded $\boldsymbol{x} \in \mathcal{R}^d$,   set $  {\boldsymbol{y}_{t+1}} = h \big( \boldsymbol{W}^\top \boldsymbol{x} + (\boldsymbol{I} - \frac{1}{N} \boldsymbol{W}^\top \boldsymbol{W}) \boldsymbol{y}_t   \big)$  with $\frac{1}{N}\boldsymbol{W}\boldsymbol{W}^\top=\boldsymbol{I}_d$. 
Denote $ \widehat{Q}(\boldsymbol{y}):=  \frac{1}{2}  \|\boldsymbol{x} - \widehat{\mathcal{A}} ({\boldsymbol{y} }) \|_2^2 + \frac{1}{N} \phi_\lambda(\boldsymbol{y})  $ and $\boldsymbol{y}^* $ as an optimal of $\widehat{Q}(\cdot)$, for $T\ge 1$,    we have
\begin{align}
   T\big(\widehat{Q} (\boldsymbol{y}_T  ) - \widehat{Q}(\boldsymbol{y}^*)  \big) & \le \frac{1}{2N}\| \boldsymbol{y}_0 - \boldsymbol{y}^* \|_2^2  - \frac{1}{2N}\|\boldsymbol{y}_T - \boldsymbol{y}^* \|_2^2  -\frac{1}{2}\sum_{t=0}^{T-1}{  \|\frac{1}{N}\boldsymbol{W}(\boldsymbol{y}_t - \boldsymbol{y}^*) \|_2^2} \nonumber \\ & -\frac{1}{2} \sum_{t=0}^{T-1}{  { \frac{t+1}{N} \| \boldsymbol{y}_{t\!+\!1} -\boldsymbol{y}_t  \|_2^2 } } 
\end{align}
\end{theorem*}
\begin{proof}
From Lemma~\ref{FiniteConvThreeQlemma}, setting $\boldsymbol{z}=\boldsymbol{y}^*$, we can achieve that 
\begin{align}
\label{FiniteQtT}
      \widehat{Q}(\boldsymbol{y}_{t+1})  \le \widehat{Q}(\boldsymbol{y}^*)+\frac{1}{2N}\big( \|\boldsymbol{y}_t -\boldsymbol{y}^*  \|_2^2 -\|\boldsymbol{y}_{t+1} -\boldsymbol{y}^*  \|_2^2  \big)   - \frac{1}{2}\| \frac{1}{N}\boldsymbol{W} (\boldsymbol{y}^*-\boldsymbol{y}_t)  \|_2^2 
\end{align}
Telescope the inequality~(\ref{FiniteQtT}) from $t=0$ to $t=T-1$, we can obtain that
\begin{align}
\label{FiniteConvTQtmp}
 \sum_{t=0}^{T-1} {   \widehat{Q}(\boldsymbol{y}_{t+1}) }  - T\widehat{Q}(\boldsymbol{y}^*) & \le \frac{1}{2N}\|\boldsymbol{y}_0 -\boldsymbol{y}^*  \|_2^2 - \frac{1}{2N}\|\boldsymbol{y}_{T} -\boldsymbol{y}^*  \|_2^2  -\frac{1}{2}\sum_{t=0}^{T-1}{  \|\frac{1}{N}\boldsymbol{W}(\boldsymbol{y}_t - \boldsymbol{y}^*) \|_2^2} 
\end{align}
From Lemma~\ref{FiniteConvStrictDescent}, we know that
\begin{align}
      \widehat{Q}(\boldsymbol{y}_{t+1})  \le \widehat{Q}(\boldsymbol{y}_t) - \frac{1}{2N}\|\boldsymbol{y}_{t+1} -\boldsymbol{y}_t  \|_2^2  
\end{align}
It follows that
\begin{align}
   \widehat{Q}(\boldsymbol{y}_{T}) \le  \widehat{Q}(\boldsymbol{y}_{t}) -\frac{1}{2N}\sum_{i=t}^{T-1} {\|\boldsymbol{y}_{i+1} -\boldsymbol{y}_i  \|_2^2 }  
\end{align}
Then, we can achieve that
\begin{align}
    T\widehat{Q}(\boldsymbol{y}_T)-T\widehat{Q}(\boldsymbol{y}^*) & \le \sum_{t=0}^{T-1}{\widehat{Q}(\boldsymbol{y}_{t+1})} - T\widehat{Q}(\boldsymbol{y}^*) - \frac{1}{2N} \sum_{t=0}^{T-1}{\sum_{i=t}^{T-1}{ \|\boldsymbol{y}_{i+1} -\boldsymbol{y}_i  \|_2^2 }} \\
    & = \sum_{t=0}^{T-1}{\widehat{Q}(\boldsymbol{y}_{t+1})} - T\widehat{Q}(\boldsymbol{y}^*)  - \frac{1}{2N}\sum_{t=0}^{T-1} {(t+1)\|\boldsymbol{y}_{t+1} -\boldsymbol{y}_t  \|_2^2} \label{FiniteConvTQtmp2}
\end{align}
Plug inequality~(\ref{FiniteConvTQtmp}) into inequality~(\ref{FiniteConvTQtmp2}), we obtain that
\begin{align}
     T\big(\widehat{Q} (\boldsymbol{y}_T  ) - \widehat{Q}(\boldsymbol{y}^*)  \big) & \le \frac{1}{2N}\| \boldsymbol{y}_0 - \boldsymbol{y}^* \|_2^2  - \frac{1}{2N}\|\boldsymbol{y}_T - \boldsymbol{y}^* \|_2^2  -\frac{1}{2}\sum_{t=0}^{T-1}{  \|\frac{1}{N}\boldsymbol{W}(\boldsymbol{y}_t - \boldsymbol{y}^*) \|_2^2} \nonumber \\ & -\frac{1}{2} \sum_{t=0}^{T-1}{  { \frac{t+1}{N} \| \boldsymbol{y}_{t\!+\!1} -\boldsymbol{y}_t  \|_2^2 } } 
\end{align}

\end{proof}

\section{Proof of Theorem~\ref{StrictMDKsparse} }

We first  show the structured samples $\boldsymbol{B}$ constructed in \cite{lyu2017spherical,lyu2020subgroup}. 

Without loss of generality, we assume that $d=2m, N=2n$. Let $ \boldsymbol{F} \in {\mathbb{C} ^{n \times n}} $ be an $n \times n$ discrete Fourier matrix. $ {\boldsymbol{F}_{k,j}} = {e^{\frac{{2\pi \boldsymbol{i}  kj}}{n}}} $ is the $(k, j) ^{th} $entry of  $\boldsymbol{F}$, where $\boldsymbol{i} = \sqrt { - 1} $. Let $\Lambda  = \{{k_1},{k_2},...,{k_m} \} \subset \{ 1,...,n - 1\} $ be a subset of indexes.

The structured  matrix $\boldsymbol{B}$  can be constructed as Eq.(\ref{eq22}).
\begin{equation}
\label{eq22}
\begin{array}{l}
\boldsymbol{B} = \frac{\sqrt{n}}{{\sqrt m }}\left[ {\begin{array}{*{20}{c}}
{{\mathop{\rm Re}\nolimits} {\boldsymbol{F}_\Lambda }}&{ - {\mathop{\rm Im}\nolimits} {\boldsymbol{F}_\Lambda }}\\
{{\mathop{\rm Im}\nolimits} {\boldsymbol{F}_\Lambda }}&{{\mathop{\rm Re}\nolimits} {\boldsymbol{F}_\Lambda }}
\end{array}} \right] \in {\mathbb{R}^{d \times N}}
\end{array}
\end{equation}
where  $\text{Re}$ and $\text{Im}$ denote the real and imaginary parts of a complex number,   and $ {\boldsymbol{F}_\Lambda } $ in Eq.~(\ref{eq23}) is the matrix constructed by $m$ rows of $\boldsymbol{F}$
\begin{equation}
\label{eq23}
\begin{array}{l}
{\boldsymbol{F}_\Lambda }{\rm{ = }} \frac{1}{\sqrt{n}} \left[ {\begin{array}{*{20}{c}}
{{e^{\frac{{2\pi \boldsymbol{i} {k_1}1}}{n}}}}& \cdots &{{e^{\frac{{2\pi  \boldsymbol{i} {k_1}n}}{n}}}}\\
 \vdots & \ddots & \vdots \\
{{e^{\frac{{2\pi \boldsymbol{i} {k_m}1}}{n}}}}& \cdots &{{e^{\frac{{2\pi \boldsymbol{i} {k_m}n}}{n}}}}
\end{array}} \right] \in {\mathbb{C}^{m \times n}}.
\end{array}
\end{equation}

The index set can be constructed by a closed-form solution~\cite{lyu2020subgroup} or by a coordinate descent method~\cite{lyu2017spherical}. 

Specifically, for a prime number $n$ such that $m$ divides $n\!-\!1$, i.e.,  $m|(n-1)$,  we can employ a closed-form construction as in \cite{lyu2020subgroup}.
Let $g$ denote a primitive root modulo $n$.  We can construct the index $\Lambda  = \{{k_1},{k_2},...,{k_m} \}$ as 
\begin{equation}
    \label{Sindex}
  \Lambda =  \{ g^0,g^{\frac{n-1}{m}}, g^{\frac{2(n-1)}{m}},\cdots, g^{\frac{(m-1)(n-1)}{m} } \} \; \text{mod} \; n. 
\end{equation}

The resulted structured matrix $\boldsymbol{B}$ has a bounded mutual coherence, which is shown in Theorem~\ref{Sbound1}.
\begin{theorem}
\label{Sbound1} \cite{lyu2020subgroup}
Suppose $d=2m, N=2n$, and $n$ is a prime such that $m|(n-1)$. Construct matrix $\boldsymbol{B}$ as in Eq.(\ref{eq22}) with index set $\Lambda$  as Eq.(\ref{Sindex}). Let mutual coherence $\mu(\boldsymbol{B}):= \max_{i \ne j} \frac{ |\boldsymbol{b}_i^\top \boldsymbol{b}_j|}{\|\boldsymbol{b}_i \|_2 \|\boldsymbol{b}_j \|_2} $. Then  $\mu(\boldsymbol{B}) \le \frac{\sqrt{n}}{m}$.
\end{theorem}
\textbf{Remark:} The bound of mutual coherence in Theorem~\ref{Sbound1} is non-trivial when $n < m^2$. For the case $n \ge m^2$, we can use the coordinate descent method in \cite{lyu2017spherical} to minimize the mutual coherence.



We now show the orthogonal property of our data-dependent structured samples $\boldsymbol{D}=\frac{\sqrt{d}}{\sqrt{N}} \boldsymbol{R}^\top\boldsymbol{B}$

\begin{property}
\label{TRo}
Suppose $d=2m, N=2n$. Let $\boldsymbol{D}=\frac{\sqrt{d}}{\sqrt{N}} \boldsymbol{R}^\top\boldsymbol{B}$ with $\boldsymbol{B}$ constructed as in Eq.(\ref{eq22}). Then $\boldsymbol{D}\boldsymbol{D}^\top = \boldsymbol{I}_d$ and column vector has constant norm, i.e.,  $\| \boldsymbol{d}_j \|_2 = \sqrt{\frac{m}{n}}$, $\forall j \in \{1,\cdots,N\}$.
\end{property}

\begin{proof}
Since $\boldsymbol{D}\boldsymbol{D}^\top = \frac{d}{N}\boldsymbol{B}\boldsymbol{B}^\top = \frac{m}{n}\boldsymbol{B}\boldsymbol{B}^\top  = \widetilde{\boldsymbol{B}}\widetilde{\boldsymbol{B}}^\top   $, where $\widetilde{\boldsymbol{B}}=\frac{\sqrt{m}}{\sqrt{n}}\boldsymbol{B} $. It follows that

\begin{equation}
\label{OrthgonalB}
\begin{array}{l}
\widetilde{\boldsymbol{B} }= \left[ {\begin{array}{*{20}{c}}
{{\mathop{\rm Re}\nolimits} {\boldsymbol{F}_\Lambda }}&{ - {\mathop{\rm Im}\nolimits} {\boldsymbol{F}_\Lambda }}\\
{{\mathop{\rm Im}\nolimits} {\boldsymbol{F}_\Lambda }}&{{\mathop{\rm Re}\nolimits} {\boldsymbol{F}_\Lambda }}
\end{array}} \right] \in {\mathbb{R}^{d \times N}}
\end{array}
\end{equation}

Let $\boldsymbol{c}_i \in \mathbb{C}^{1 \times n }$ be the $i^{th}$ row of matrix $\boldsymbol{F}_{\Lambda} \in \mathbb{C}^{m \times n}$ in Eq.(\ref{eq23}). Let $\boldsymbol{v}_i \in \mathbb{R}^{1 \times 2n}$ be the  $i^{th}$ row of matrix $\widetilde{\boldsymbol{B}} \in \mathbb{R}^{2m \times 2n}$ in Eq.(\ref{OrthgonalB}).
For $1 \le i,j \le m$, $i \ne j$, we know that 
\begin{align}
 & \boldsymbol{v}_i \boldsymbol{v}_{i+m}^\top = 0, \\
    &\boldsymbol{v}_{i+m} \boldsymbol{v}_{j+m}^\top = \boldsymbol{v}_i \boldsymbol{v}_j^\top = \text{Re} (\boldsymbol{c}_i\boldsymbol{c}_j^*), \\
    & \boldsymbol{v}_{i+m} \boldsymbol{v}_{j}^\top = -\boldsymbol{v}_i \boldsymbol{v}_{j+m}^\top =  \text{Im} (\boldsymbol{c}_i\boldsymbol{c}_j^*),
\end{align}
where $*$ denotes the complex conjugate, $\text{Re}(\cdot)$ and $\text{Im}(\cdot)$ denote the real and imaginary parts of the input complex number.

For a discrete Fourier matrix $\boldsymbol{F}$, we know that 
\begin{equation}
\label{cij}
     \boldsymbol{c}_i\boldsymbol{c}_j^* = \frac{1}{n} \sum_{k=0}^{n-1} {e^{\frac{2\pi (i-j)k \boldsymbol{i} }{n}}} = 
     \begin{cases}
      1, & \text{if}\ i=j \\
      0, & \text{otherwise}
    \end{cases}
\end{equation}

When $i \ne j$, from Eq.(\ref{cij}), we know $\boldsymbol{c}_i\boldsymbol{c}_j^* = 0$. Thus, we have
\begin{align}
    &\boldsymbol{v}_{i+m} \boldsymbol{v}_{j+m}^\top = \boldsymbol{v}_i \boldsymbol{v}_j^\top = \text{Re} (\boldsymbol{c}_i\boldsymbol{c}_j^*) = 0, \\
    & \boldsymbol{v}_{i+m} \boldsymbol{v}_{j}^\top = -\boldsymbol{v}_i \boldsymbol{v}_{j+m}^\top =  \text{Im} (\boldsymbol{c}_i\boldsymbol{c}_j^*) =0,
\end{align}

When $i=j$, we know that $\boldsymbol{v}_{i+m} \boldsymbol{v}_{i+m}^\top = \boldsymbol{v}_i \boldsymbol{v}_i^\top = \boldsymbol{c}_i\boldsymbol{c}_i^* =1 $.

Put two cases together, also note that $d=2m$,  we have  $\boldsymbol{D}\boldsymbol{D}^\top=\widetilde{\boldsymbol{B}}\widetilde{\boldsymbol{B}}^\top = \boldsymbol{I}_d$.

The $l_2$-norm of the column vector of $\widetilde{\boldsymbol{B}}$ is given as
\begin{equation}
    \| \widetilde{\boldsymbol{b}}_j  \|_2^2 = \frac{1}{n} \sum_{i=1}^{m}{ \big( \sin^2{\frac{2\pi k_i j }{n}} + \cos^2{\frac{2\pi k_i j}{n}} \big ) }  = \frac{m}{n}
\end{equation}
Thus, we have $ \| \boldsymbol{d}_j \|_2  =   \| \widetilde{\boldsymbol{b}}_j  \|_2^2  = \sqrt{\frac{m}{n}}$ for $j \in \{ 1,\cdots, M \}$

\end{proof}

\begin{Lemma}
\label{MuDbound}
Let $\boldsymbol{D} = \frac{\sqrt{d}}{\sqrt{N}} \boldsymbol{R}^\top\boldsymbol{B}$, where $\boldsymbol{B}$ is constructed  as as in Eq.(\ref{eq22}) with index set $\Lambda$  as Eq.(\ref{Sindex})~\cite{lyu2020subgroup} with $N=2n, d=2m$. 
$\forall \boldsymbol{y} \in \mathcal{R}^N, \|\boldsymbol{y} \|_0 \le 2k$, we have $ \|\boldsymbol{D}\boldsymbol{y} \|_2^2 - \| \boldsymbol{y}\|_2^2 \le - \frac{n- (2k-1)\sqrt{n} -m }{n}\|\boldsymbol{y}\|_2^2$
\end{Lemma}
\begin{proof}


Denote $\boldsymbol{M}= \boldsymbol{D}^\top \boldsymbol{D}$.  Since the column vector of $\boldsymbol{D}$ has constant norm, i.e.,  $\|\boldsymbol{d}_j\|_2^2 =\frac{m}{n}$,  it follows that
\begin{align}
    \|\boldsymbol{D}\boldsymbol{y} \|_2^2 & = \boldsymbol{y}^\top\boldsymbol{M}\boldsymbol{y} = \|\boldsymbol{d}_j \|_2^2 \big( \sum_{i=1}^N{y_i^2} +  \sum_{i=1}^N{\sum_{j=1, j \ne i}^N {y_i y_j M_{ij}} } \big) \\ & = \frac{m}{n} \|\boldsymbol{y} \|_2^2 + \frac{m}{n}\sum_{i=1}^N{\sum_{j=1, j \ne i}^N {y_i y_j M_{ij}} } 
    \\ & \le \frac{m}{n} \|\boldsymbol{y} \|_2^2 +  \frac{m}{n} \mu(\boldsymbol{D})  \big(\sum_{i=1}^N{\sum_{j=1, j \ne i}^N {|y_i| |y_j| } }  \big) \\
    & =\frac{m}{n}\|\boldsymbol{y} \|_2^2 + \frac{m}{n}\mu(\boldsymbol{D}) \left( \big(\sum_{i=1}^N{|y_i|} \big)^2 -  \sum_{i=1}^N{y_i^2}   \right)
\end{align}
Since $\|\boldsymbol{y} \|_0 \le 2k$, we know there is at most $2k$ non-zero elements among $\boldsymbol{y}$. Thus, we know that
\begin{align}
     \|\boldsymbol{D}\boldsymbol{y} \|_2^2 & \le \frac{m}{n}\|\boldsymbol{y} \|_2^2 + \frac{m}{n} \mu(\boldsymbol{D}) \left( \big(\sum_{i=1}^N{|y_i|} \big)^2 -  \sum_{i=1}^N{y_i^2}   \right) \\
     & \le \frac{m}{n} \|\boldsymbol{y} \|_2^2 + \frac{m}{n} \mu(\boldsymbol{D}) \big( 2k\sum_{i=1}^N{y_i^2}    -  \sum_{i=1}^N{y_i^2}   \big) \\
     &= \frac{m}{n} \|\boldsymbol{y} \|_2^2  + \frac{m}{n} \mu(\boldsymbol{D}) (2k-1) \| \boldsymbol{y}\|_2^2
\end{align}
Since $\mu(\boldsymbol{D})=\boldsymbol{B}$, from Theorem~\ref{Sbound1}, we know $\mu(\boldsymbol{D})\le \frac{\sqrt{n}}{m}$. It follows that
\begin{align}
     \|\boldsymbol{D}\boldsymbol{y} \|_2^2 & \le \frac{m}{n}\|\boldsymbol{y} \|_2^2  + \frac{m}{n} \mu(\boldsymbol{D}) (2k-1) \| \boldsymbol{y}\|_2^2  \\ 
     & \le \frac{m}{n}\|\boldsymbol{y} \|_2^2  + \frac{m}{n} \frac{(2k-1)\sqrt{n}}{m} \| \boldsymbol{y}\|_2^2  \\
     & = \frac{(2k-1)\sqrt{n}+m}{n}\| \boldsymbol{y}\|_2^2 
\end{align}
It follows that $ \|\boldsymbol{D}\boldsymbol{y} \|_2^2 - \|\boldsymbol{y}\|_2^2 \le \frac{(2k-1)\sqrt{n}+m - n }{n}\| \boldsymbol{y}\|_2^2 $.
\end{proof}

\begin{theorem*}(Strictly Monotonic Descent of $k$-sparse  problem )
Let $
    L(\boldsymbol{y}) = \frac{1}{2} \| {\boldsymbol{x}} - \boldsymbol{D}\boldsymbol{y}  \|_2^2, \; s.t. \; \|\boldsymbol{y} \|_0 \le k$ with $\boldsymbol{D} = \frac{\sqrt{d}}{\sqrt{N}} \boldsymbol{R}^\top\boldsymbol{B}$, where $\boldsymbol{B}$ is constructed  as as in Eq.(\ref{eq22}) with index set $\Lambda$  as Eq.(\ref{Sindex})~\cite{lyu2020subgroup} with $N=2n, d=2m$. 
    Set $\boldsymbol{y}_{t+1} = h(    \boldsymbol{a}_{t+1} ) $ with sparity $k$ and $\boldsymbol{a}_{t+1} = \boldsymbol{D}^\top\boldsymbol{x} +  (\boldsymbol{I}-\boldsymbol{D}^\top\boldsymbol{D})\boldsymbol{y}_t $, we have
    \begin{equation}
    L(\boldsymbol{y}_{t+1}) \le    L(\boldsymbol{y}_t) + \frac{1}{2}\|\boldsymbol{y}_{t+1} - \boldsymbol{a}_{t+1}  \|_2^2 - \frac{1}{2}\|\boldsymbol{y}_t - \boldsymbol{a}_{t+1}  \|_2^2 -\frac{n- (2k-1)\sqrt{n} -m }{2n}\| \boldsymbol{y}_{t+1} -\boldsymbol{y}_t \|_2^2 \le  L(\boldsymbol{y}_t) 
    \end{equation}
    where $h(\cdot)$ is defined as 
    \begin{align}
         h({z}_j) = \left\{
                \begin{array}{ll}
                  {z}_j  \;\;\;\;\;  \text{if} \;\;\;  | {z}_j | \; \text{is one of the k-highest  values of } \;|\boldsymbol{z} |\in \mathcal{R}^N \\
                  0    \;\;\;\;\;\;\;  \text{otherwise}
                \end{array}
              \right. .
    \end{align}
\end{theorem*}

\begin{proof}

Denote $L(\boldsymbol{y}):=\frac{1}{2}\|\boldsymbol{x}-\boldsymbol{D}\boldsymbol{y} \|_2^2 $. It follows that
\begin{align}
 L(\boldsymbol{y}_{t+1}) & =   \frac{1}{2}\|\boldsymbol{x}-\boldsymbol{D}\boldsymbol{y}_{t+1} \|_2^2= \frac{1}{2} \|\boldsymbol{x}-\boldsymbol{D}\boldsymbol{y}_{t} + \boldsymbol{D}\boldsymbol{y}_{t} - \boldsymbol{D}\boldsymbol{y}_{t+1} \|_2^2 \\ 
 & =  L(\boldsymbol{y}_{t}) + \left< \boldsymbol{x}-\boldsymbol{D}\boldsymbol{y}_{t} , \boldsymbol{D}( \boldsymbol{y}_{t} - \boldsymbol{y}_{t+1} ) \right> + \|\boldsymbol{D}( \boldsymbol{y}_{t} - \boldsymbol{y}_{t+1} ) \|_2^2 \\
 & = L(\boldsymbol{y}_{t}) + \left< \boldsymbol{D}^\top\boldsymbol{x}-\boldsymbol{D}^\top\boldsymbol{D}\boldsymbol{y}_{t} , \boldsymbol{y}_{t} - \boldsymbol{y}_{t+1}  \right> + \|\boldsymbol{D}( \boldsymbol{y}_{t} - \boldsymbol{y}_{t+1} ) \|_2^2 \\
 & = L(\boldsymbol{y}_{t}) + \left< \boldsymbol{D}^\top\boldsymbol{D}\boldsymbol{y}_{t}-\boldsymbol{D}^\top\boldsymbol{x} , \boldsymbol{y}_{t+1}  -\boldsymbol{y}_{t}   \right> + \|\boldsymbol{D}( \boldsymbol{y}_{t} - \boldsymbol{y}_{t+1} ) \|_2^2 \label{InkL}
\end{align}

Let $\boldsymbol{a}_{t+1}=  \boldsymbol{D}^\top\boldsymbol{x} + (\boldsymbol{I}-\boldsymbol{D}^\top\boldsymbol{D})\boldsymbol{y}_t $, together with Eq.(\ref{InkL}), we can obtain that
\begin{align}
     L(\boldsymbol{y}_{t+1}) & =  L(\boldsymbol{y}_t) + \left<\boldsymbol{y}_t -  \boldsymbol{a}_{t+1}, \boldsymbol{y}_{t+1}-\boldsymbol{y}_t \right> + \frac{1}{2}\| \boldsymbol{D}(\boldsymbol{y}_{t+1} -\boldsymbol{y}_t )\|_2^2   \\
     & = L(\boldsymbol{y}_t) + \frac{\|\boldsymbol{y}_{t+1} - \boldsymbol{a}_{t+1}  \|_2^2 - \|\boldsymbol{y}_t - \boldsymbol{a}_{t+1}  \|_2^2 - \|\boldsymbol{y}_{t+1}-\boldsymbol{y}_t \|_2^2 }{2} +  \frac{1}{2}\| \boldsymbol{D}(\boldsymbol{y}_{t+1} -\boldsymbol{y}_t )\|_2^2   
\end{align}
From Lemma~\ref{MuDbound}, we know that
\begin{align}
    \frac{1}{2}\| \boldsymbol{D}(\boldsymbol{y}_{t+1} -\boldsymbol{y}_t )\|_2^2    - \frac{1}{2}\|\boldsymbol{y}_{t+1}-\boldsymbol{y}_t \|_2^2 \le  -\frac{n- (2k-1)\sqrt{n} -m }{2n}\| \boldsymbol{y}_{t+1} -\boldsymbol{y}_t \|_2^2
\end{align}
It follows that
\begin{align}
      L(\boldsymbol{y}_{t+1}) \le L(\boldsymbol{y}_t) + \frac{1}{2} \|\boldsymbol{y}_{t+1} - \boldsymbol{a}_{t+1}  \|_2^2   -\frac{1}{2} \|\boldsymbol{y}_t - \boldsymbol{a}_{t+1}  \|_2^2  - \frac{n- (2k-1)\sqrt{n} -m }{2n}\| \boldsymbol{y}_{t+1} -\boldsymbol{y}_t \|_2^2
\end{align}

Note that $\boldsymbol{y}_{t+1} := \argmin _{\boldsymbol{y}, \|\boldsymbol{y} \|_0 \le k } {\|\boldsymbol{y} -\boldsymbol{a}_{t+1} \|_2^2}  $, we know $\|\boldsymbol{y}_{t+1} -\boldsymbol{a}_{t+1} \|_2^2 \le \|\boldsymbol{y}_t -\boldsymbol{a}_{t+1} \|_2^2$. It follows that 
$ L(\boldsymbol{y}_{t+1}) \le L(\boldsymbol{y}_t)$, in which the equality holds true when $\|\boldsymbol{y}_{t+1} -\boldsymbol{a}_{t+1} \|_2^2 = \|\boldsymbol{y}_t -\boldsymbol{a}_{t+1} \|_2^2$ and $\|\boldsymbol{y}_{t+1} -\boldsymbol{y}_t \|_2^2=0$

\end{proof}

\section{A Better Diagonal Random Rotation for SSF~\cite{lyu2017spherical}}

In~\cite{lyu2017spherical}, a diagonal rotation matrix $\boldsymbol{D}$ is constructed by sampling its diagonal elements uniformly from $\{-1,+1\}$.  In this section, we propose a better diagonal random rotation.  Without loss of generality, we assume that $d=2m, N=2n$.

We first generate a diagonal complex matrix $\boldsymbol{D} \in \mathbb{C}^{m \times m} $, in which the diagonal elements are constructed as
\begin{align}
    \boldsymbol{D}_{jj} = \text{cos} \theta_j + \boldsymbol{i} \; \text{sin} \theta_j \;  , \; \forall{j} \in \{1,\cdots,m\}
\end{align}
\sloppy
where $\theta_j, \forall{j} \in \{1,\cdots,m\}$ are i.i.d. samples from the uniform distribution $Uni[0,2\pi)$, and $\boldsymbol{i}=\sqrt{-1}$.

We then generate a uniformly random permutation $\Pi : \{1,\cdots,d\} \to \{1,\cdots,d\}$. The SSF samples can be constructed as  $\boldsymbol{H} = \Pi \circ \widetilde{\boldsymbol{B}}$ with $\widetilde{\boldsymbol{B}}$:
\begin{equation}
\begin{array}{l}
\widetilde{\boldsymbol{B}} = \frac{\sqrt{n}}{{\sqrt m }}\left[ {\begin{array}{*{20}{c}}
{{\mathop{\rm Re}\nolimits} {\widetilde{\boldsymbol{F}}_\Lambda }}&{ - {\mathop{\rm Im}\nolimits} {\widetilde{\boldsymbol{F}}_\Lambda }}\\
{{\mathop{\rm Im}\nolimits} {\widetilde{\boldsymbol{F}}_\Lambda }}&{{\mathop{\rm Re}\nolimits} {\widetilde{\boldsymbol{F}}_\Lambda }}
\end{array}} \right] \in {\mathbb{R}^{d \times N}}
\end{array}
\end{equation}
where $\widetilde{\boldsymbol{F}}_\Lambda  = \boldsymbol{D}\boldsymbol{F}_\Lambda $. 

It is worth noting that $\boldsymbol{H}^\top\boldsymbol{H} = \boldsymbol{B}^\top\boldsymbol{B}$, which means that the proposed the diagonal rotation scheme preserved the pairwise inner product of SSF~\cite{lyu2017spherical}.
Moreover, the  SSF with the proposed random rotation maintains $O(d)$ space complexity and $O(n\log n )$ (matrix-vector product)  time complexity by FFT.

\section{Rademacher Complexity}

\textbf{Neural Network Structure:}
For structured approximated NOK networks (SNOK), the $1$-$T$ layers are given as 
\begin{align}
    \boldsymbol{y}_{t+1} = h( \boldsymbol{D}^\top\boldsymbol{R}_t\boldsymbol{x} +  (\boldsymbol{I}-\boldsymbol{D}^\top\boldsymbol{D})\boldsymbol{y}_t   )
\end{align}
where $\boldsymbol{R}_t$ are free parameters such that $\boldsymbol{R}_t^\top\boldsymbol{R}_t=\boldsymbol{R}_t^\top \boldsymbol{R}_t = \boldsymbol{I}_d$. And $\boldsymbol{D}$ is the  scaled structured spherical samples such that $\boldsymbol{D}\boldsymbol{D}^\top = \boldsymbol{I}_d$, and $\boldsymbol{y}_0 =\boldsymbol{0}$. 

The  last layer ( $(T\!\!+\!\!1)^{th}$ layer) is given by ${z} = \boldsymbol{w}^\top\boldsymbol{y}_{T\!+\!1}$.  Consider a $L$-Lipschitz continuous loss function $\ell( {z}, y) :  {\mathcal{Z}}  \times \mathcal{Y} \to [0,1]  $ with Lipschitz constant $L$ w.r.t the input ${z}$.

\textbf{Rademacher Complexity:} Rademacher complexity of a function class $\mathcal{G}$ is defined as 
\begin{align}
    \mathfrak{R}_N(\mathcal{G}):= \frac{1}{N} \mathbb{E} \left[ \sup _{g \in \mathcal{G}} \sum _{i=1} ^{N} \epsilon_i g(\boldsymbol{x}_i)   \right]
\end{align}
where $\epsilon_i, i \in \{1,\cdots,N\}$ are i.i.d. samples drawn uniformly from $\{+1,-1\}$ with probality $\text{P}[\epsilon_i=+1]=\text{P}[\epsilon_i=-1]=1/2$. And $\boldsymbol{x}_i,  i \in \{1,\cdots,N\}$ are i.i.d.  samples from $\mathcal{X}$.

\begin{theorem*}
(Rademacher Complexity Bound)
Consider  a Lipschitz continuous loss function $\ell( {z}, y) :  {\mathcal{Z}}  \times \mathcal{Y} \to [0,1]  $ with Lipschitz constant $L$ w.r.t the input ${z}$.    Let $\widetilde{\ell}({z},y):= \ell( z, y)  -  \ell( 0, y) $.  Let $\widehat{G}$ be the function class of our $(T\!\!+\!\!1)$-layer SNOK mapping from $\mathcal{X}$ to $\mathcal{Z}$. Suppose the activation function $|h(\boldsymbol{y})|\le |\boldsymbol{y}|$ (element-wise),  and the  $l_2$-norm of last layer weight is bounded, i.e.,   $\|\boldsymbol{w} \|_2 \le \mathcal{B}_w$.  Let $(\boldsymbol{x}_i, y_i)_{i=1}^N$ be i.i.d. samples drawn from $\mathcal{X} \times \mathcal{Y}$.   Let $\boldsymbol{Y}_{T\!+\!1}=[\boldsymbol{y}_{T\!+\!1}^{(1)}, \cdots, \boldsymbol{y}_{T\!+\!1}^{(N)} ]$ be the $T^{th}$ layer  output with input $\boldsymbol{X}$. Denote the mutual coherence of $\boldsymbol{Y}_{T\!+\!1}$ as $\mu^*$, i.e.,  $ \mu^* =  \mu(\boldsymbol{Y}_{T\!+\!1}) =  \max _{i \ne j} \frac{\boldsymbol{y}_{T\!+\!1}^{(i)\top} \boldsymbol{y}_{T\!+\!1}^{(j)} }{ \|\boldsymbol{y}_{T\!+\!1}^{(i)}\|_2 \| \boldsymbol{y}_{T\!+\!1}^{(j)} \|_2}  \le 1$.   Then, we have
\begin{align}
    \mathfrak{R}_N(\widetilde{\ell} \circ \widehat{G} )  = \frac{1}{N} \mathbb{E} \left[ \sup _{g \in \widehat{\mathcal{G}} } \sum _{i=1} ^{N} \epsilon_i  \widetilde{\ell}( g(\boldsymbol{x}_i) , y_i )    \right] &  \le  \frac{L\mathcal{B}_w \sqrt{ \big((N-1) \mu^* +1 \big)T } }{N}   \| \boldsymbol{X} \|_F 
\end{align}
where    $\boldsymbol{X}=[\boldsymbol{x}_1,\cdots,\boldsymbol{x}_N]$.  $\| \cdot\|_2$ and   $\|\cdot \|_F$ denote the spectral norm and  the Frobenius norm of input matrix, respectively.
\end{theorem*}
\textbf{Remark:}   The Rademacher complexity bound  has a complexity  $O(\sqrt{T})$ w.r.t. the depth of NN  (SNOK).


\begin{proof}
Since $\widetilde{\ell}$ is $L$-Lipschitz continuous function, from the composition rule of Rademacher complexity, we know that
\begin{align}
    \mathfrak{R}_N(\widetilde{\ell} \circ \widehat{G} )  \le L \; \mathfrak{R}_N(\widehat{G} ) 
\end{align}
It follows that
\begin{align}
    \mathfrak{R}_N(\widehat{G} ) & = \frac{1}{N}\mathbb{E} \left[ \sup _{g \in \widehat{\mathcal{G}} } \sum _{i=1} ^{N} \epsilon_i   f(\boldsymbol{x}_i)    \right]  \\
    & =  \frac{1}{N}\mathbb{E} \left[ \sup _{\boldsymbol{w}, \{\boldsymbol{R}_t \in \text{SO}(d) \}_{t=1}^T} \sum _{i=1} ^{N}   \epsilon_i   \big<\boldsymbol{w}, \boldsymbol{y}_{T\!+\!1}^{(i)}  \big>   \right]  \\ 
    & = \frac{1}{N}\mathbb{E} \left[ \sup _{\boldsymbol{w}, \{\boldsymbol{R}_t \in \text{SO}(d) \}_{t=1}^T}   \big<\boldsymbol{w}, \sum _{i=1} ^{N}   \epsilon_i \boldsymbol{y}_{T\!+\!1}^{(i)}  \big>   \right]  \\
    & \le \frac{1}{N}\mathbb{E} \left[ \sup _{\boldsymbol{w}, \{\boldsymbol{R}_t \in \text{SO}(d) \}_{t=1}^T}   \| \boldsymbol{w} \|_2  \; \big\|\sum _{i=1} ^{N}   \epsilon_i \boldsymbol{y}_{T\!+\!1}^{(i)}  \big \|_2  \right]  \;\;  \text{(Cauchy-Schwarz  inequality)} \\
    & \le \frac{\mathcal{B}_w}{N}\mathbb{E} \left[ \sup _{ \{\boldsymbol{R}_t \in \text{SO}(d) \}_{t=1}^T} \big\|\sum _{i=1} ^{N}   \epsilon_i \boldsymbol{y}_{T\!+\!1}^{(i)}  \big \|_2  \right]  \\
    & = \frac{\mathcal{B}_w}{N}\mathbb{E} \left[ \sup _{ \{\boldsymbol{R}_t \in \text{SO}(d) \}_{t=1}^T}  \sqrt{ \sum _{i=1} ^{N}  \big \|   \epsilon_i \boldsymbol{y}_{T\!+\!1}^{(i)}  \big\|_2^2   +  \sum _{i=1} ^{N} {\sum _{j=1,j\ne i} ^{N} {  \epsilon_i  \epsilon_j  \boldsymbol{y}_{T\!+\!1}^{(i)\top} \boldsymbol{y}_{T\!+\!1}^{(j)}   } }  }    \right] \\
    & = \frac{\mathcal{B}_w}{N}\mathbb{E} \left[ \sup _{ \{\boldsymbol{R}_t \in \text{SO}(d) \}_{t=1}^T}  \sqrt{ \sum _{i=1} ^{N}  \big \|    \boldsymbol{y}_{T\!+\!1}^{(i)}  \big\|_2^2   +  \sum _{i=1} ^{N} {\sum _{j=1,j\ne i} ^{N} {  \epsilon_i  \epsilon_j  \boldsymbol{y}_{T\!+\!1}^{(i)\top} \boldsymbol{y}_{T\!+\!1}^{(j)}   } }  }    \right]  \\
    & = \frac{\mathcal{B}_w}{N}\mathbb{E} \left[   \sqrt{ \sup _{ \{\boldsymbol{R}_t \in \text{SO}(d) \}_{t=1}^T} \sum _{i=1} ^{N}  \big \|    \boldsymbol{y}_{T\!+\!1}^{(i)}  \big\|_2^2   +  \sum _{i=1} ^{N} {\sum _{j=1,j\ne i} ^{N} {  \epsilon_i  \epsilon_j  \boldsymbol{y}_{T\!+\!1}^{(i)\top} \boldsymbol{y}_{T\!+\!1}^{(j)}   } }  }    \right] \\
    & \le \frac{\mathcal{B}_w}{N}   \sqrt{ \mathbb{E} \left[ \sup _{ \{\boldsymbol{R}_t \in \text{SO}(d) \}_{t=1}^T} \sum _{i=1} ^{N}  \big \|    \boldsymbol{y}_{T\!+\!1}^{(i)}  \big\|_2^2   +  \sum _{i=1} ^{N} {\sum _{j=1,j\ne i} ^{N} {  \epsilon_i  \epsilon_j  \boldsymbol{y}_{T\!+\!1}^{(i)\top} \boldsymbol{y}_{T\!+\!1}^{(j)}   } } \right]  }    \label{Jensen}
\end{align}
Inequality~(\ref{Jensen}) is because of the Jensen inequality and concavity of the square root function. 

Note that $|\epsilon_i|=1, \forall i \in \{1,\cdots,N\}$,  and 
 the mutual coherence of $\boldsymbol{Y}_{T\!+\!1}$ is $\mu^*$, i.e., $ \mu^* =  \mu(\boldsymbol{Y}_{T\!+\!1}) \le 1$,   it follows that
\begin{align}
      \mathfrak{R}_N(\widehat{F} )  & \le \frac{\mathcal{B}_w}{N}   \sqrt{ \mathbb{E} \left[ \sup _{ \{\boldsymbol{R}_t \in \text{SO}(d) \}_{t=1}^T} \sum _{i=1} ^{N}  \big \|    \boldsymbol{y}_{T\!+\!1}^{(i)}  \big\|_2^2   +  \sum _{i=1} ^{N} {\sum _{j=1,j\ne i} ^{N} {  \epsilon_i  \epsilon_j  \boldsymbol{y}_{T\!+\!1}^{(i)\top} \boldsymbol{y}_{T\!+\!1}^{(j)}   } } \right]  }    \\
      & \le    \frac{\mathcal{B}_w}{N}   \sqrt{ \sup _{ \{\boldsymbol{R}_t \in \text{SO}(d) \}_{t=1}^T}   \sum _{i=1} ^{N}  \big \|    \boldsymbol{y}_{T\!+\!1}^{(i)}  \big\|_2^2      + \sum _{i=1} ^{N} {\sum _{j=1,j\ne i} ^{N} {  \big \|  \boldsymbol{y}_{T\!+\!1}^{(i)} \big \|_2 \big \| \boldsymbol{y}_{T\!+\!1}^{(j)} \big \|_2 \mu^*  } } }    \\
      & =   \frac{\mathcal{B}_w}{N}   \sqrt{ \sup _{ \{\boldsymbol{R}_t \in \text{SO}(d) \}_{t=1}^T} (1-\mu^*) \sum _{i=1} ^{N}  \big \|    \boldsymbol{y}_{T\!+\!1}^{(i)}  \big\|_2^2   +  \mu^* \big( \sum _{i=1} ^{N}  \big \|    \boldsymbol{y}_{T\!+\!1}^{(i)}  \big\|_2 \big)^2  }  \\
       & \le    \frac{\mathcal{B}_w}{N}   \sqrt{ \sup _{ \{\boldsymbol{R}_t \in \text{SO}(d) \}_{t=1}^T} (1-\mu^*) \|\boldsymbol{Y}_{T\!+\!1} \|_F^2   +  N \mu^* \|\boldsymbol{Y}_{T\!+\!1} \|_F^2  }   \; \; \text{Cauchy-Schwarz} \\
      &   \le    \frac{\mathcal{B}_w}{N}   \sqrt{ \sup _{ \{\boldsymbol{R}_t \in \text{SO}(d) \}_{t=1}^T}   \big((N-1) \mu^* +1 \big) \|\boldsymbol{Y}_{T\!+\!1} \|_F^2  }    \label{supIneq}
\end{align}
where $\boldsymbol{Y}_{T\!+\!1}=[\boldsymbol{y}_{T\!+\!1}^{(1)}, \cdots, \boldsymbol{y}_{T\!+\!1}^{(N)} ]$ and $\|\cdot \|_F$ denotes the Frobenius norm.

Since $|h(\boldsymbol{Y})| \le |\boldsymbol{Y}| $ (element-wise), (e.g., ReLU, max-pooling, soft-thresholding),   it follows that 
\begin{align}
   \|\boldsymbol{Y}_{T\!+\!1} \|_F^2 & =   \| h\big(  \boldsymbol{D}^\top\boldsymbol{R}_{T}\boldsymbol{X} +  (\boldsymbol{I}-\boldsymbol{D}^\top\boldsymbol{D})\boldsymbol{Y}_T \big) \|_F^2  \\
    & \le  \| \boldsymbol{D}^\top\boldsymbol{R}_{T}\boldsymbol{X} +  (\boldsymbol{I}-\boldsymbol{D}^\top\boldsymbol{D})\boldsymbol{Y}_{T}  \|_F^2 
\end{align}

In addition, we have
\begin{align}
   \| \boldsymbol{D}^\top\boldsymbol{R}_{T}\boldsymbol{X} +  (\boldsymbol{I}-\boldsymbol{D}^\top\boldsymbol{D})\boldsymbol{Y}_{T}  \|_F^2  & =   \| \boldsymbol{D}^\top\boldsymbol{R}_{T}\boldsymbol{X} \|_F^2 +   \| (\boldsymbol{I}-\boldsymbol{D}^\top\boldsymbol{D})\boldsymbol{Y}_{T}  \|_F^2 \nonumber \\ & + 2\big<\boldsymbol{D}^\top\boldsymbol{R}_{T}\boldsymbol{X}, (\boldsymbol{I}\!-\!\boldsymbol{D}^\top\boldsymbol{D})\boldsymbol{Y}_{T}  \big>
\end{align}
Note that $\boldsymbol{D}\boldsymbol{D}^\top = \boldsymbol{I}_d$ and $\boldsymbol{R}_T^\top\boldsymbol{R}_T= \boldsymbol{R}_T\boldsymbol{R}_T^\top = \boldsymbol{I}_d$, we have 
\begin{align}
    & \| \boldsymbol{D}^\top\boldsymbol{R}_{T}\boldsymbol{X} \|_F^2 = \| \boldsymbol{X} \|_F^2  \\
    & \big<\boldsymbol{D}^\top\boldsymbol{R}_{T}\boldsymbol{X}, (\boldsymbol{I}\!-\!\boldsymbol{D}^\top\boldsymbol{D})\boldsymbol{Y}_{T}  \big> = \text{tr}\left( \boldsymbol{X}^\top \boldsymbol{R}_T^\top \boldsymbol{D}(\boldsymbol{I}\!-\!\boldsymbol{D}^\top\boldsymbol{D})\boldsymbol{Y}_{T}   \right) =0
\end{align}
Denote $\beta =\| \boldsymbol{I}-\boldsymbol{D}^\top\boldsymbol{D} \|_2^2  $, it follows that
\begin{align}
   \| \boldsymbol{D}^\top\boldsymbol{R}_{T}\boldsymbol{X} +  (\boldsymbol{I}-\boldsymbol{D}^\top\boldsymbol{D})\boldsymbol{Y}_{T}  \|_F^2 &  =   \| \boldsymbol{D}^\top\boldsymbol{R}_{T}\boldsymbol{X} \|_F^2 +   \| (\boldsymbol{I}-\boldsymbol{D}^\top\boldsymbol{D})\boldsymbol{Y}_{T}  \|_F^2 \nonumber \\ & + 2\big<\boldsymbol{D}^\top\boldsymbol{R}_{T}\boldsymbol{X}, (\boldsymbol{I}\!-\!\boldsymbol{D}^\top\boldsymbol{D})\boldsymbol{Y}_{T}  \big> \\
   & =  \| \boldsymbol{X} \|_F^2 +  \| (\boldsymbol{I}-\boldsymbol{D}^\top\boldsymbol{D})\boldsymbol{Y}_{T}  \|_F^2  \\
   & \le  \| \boldsymbol{X} \|_F^2  + \|\boldsymbol{I}-\boldsymbol{D}^\top\boldsymbol{D} \|_2^2  \| \boldsymbol{Y}_{T}  \|_F^2   =  \| \boldsymbol{X} \|_F^2  + \beta \| \boldsymbol{y}_{T}  \|_F^2
\end{align}

Recursively apply the above procedure from $t=T $ to $t=1$, together with $\boldsymbol{Y}_0 = \boldsymbol{0}$,  we  can achieve that 
\begin{align}
    \|\boldsymbol{Y}_{T\!+\!1} \|_F^2 \le  \| \boldsymbol{X} \|_F^2 \big( \sum_{i=0}^{T-1} \beta^{i}  \big) 
\end{align} 
Together with inequality~(\ref{supIneq}), it follows that 
\begin{align}
  \mathfrak{R}_N(\widehat{G} ) & \le   \frac{\mathcal{B}_w}{N}   \sqrt{ \sup _{ \{\boldsymbol{R}_t \in \text{SO}(d) \}_{t=1}^T}   \big((N-1) \mu^* +1 \big) \|\boldsymbol{Y}_{T\!+\!1} \|_F^2  }    \\
   & \le  \frac{\mathcal{B}_w \sqrt{ \big((N-1) \mu^* +1 \big) } }{N}  \sqrt{ \sum_{i=0}^{T-1} \beta^{i} }  \| \boldsymbol{X} \|_F
\end{align}
Finally, we obtain that
\begin{align}
      \mathfrak{R}_N(\widetilde{\ell} \circ \widehat{G} )  =  \frac{1}{N} \mathbb{E} \left[ \sup _{g \in \widehat{\mathcal{G}} } \sum _{i=1} ^{N} \epsilon_i  \widetilde{\ell}( g(\boldsymbol{x}_i) , y_i )    \right] &  \le  \frac{L\mathcal{B}_w \sqrt{ \big((N-1) \mu^* +1 \big) } }{N}  \sqrt{ \sum_{i=0}^{T-1} \beta^{i}  }  \| \boldsymbol{X} \|_F 
\end{align}
Now, we show that $\beta = \| \boldsymbol{I}-\boldsymbol{D}^\top\boldsymbol{D} \|_2^2 \le 1 $.  From the definition of spectral norm, we  have that
\begin{align}
    \beta = \| \boldsymbol{I}-\boldsymbol{D}^\top\boldsymbol{D} \|_2^2 & = \sup _ {\| \boldsymbol{y} \|_2 =1} {\| (\boldsymbol{I}-\boldsymbol{D}^\top\boldsymbol{D}) \boldsymbol{y} \|_2^2} \\
    & = \sup _ {\| \boldsymbol{y} \|_2 =1} {\boldsymbol{y}^\top (\boldsymbol{I}-\boldsymbol{D}^\top\boldsymbol{D})^\top(\boldsymbol{I}-\boldsymbol{D}^\top\boldsymbol{D})\boldsymbol{y} } \\
    & = \sup _ {\| \boldsymbol{y} \|_2 =1} {\boldsymbol{y}^\top \big(\boldsymbol{I}-2\boldsymbol{D}^\top\boldsymbol{D} + \boldsymbol{D}^\top\boldsymbol{D}\boldsymbol{D}^\top\boldsymbol{D} \big)\boldsymbol{y} } \\
    & = \sup _ {\| \boldsymbol{y} \|_2 =1} {\boldsymbol{y}^\top  (\boldsymbol{I}-\boldsymbol{D}^\top\boldsymbol{D})\boldsymbol{y} } \\
    & = 1 - \min _{\| \boldsymbol{y} \|_2 =1} { \|\boldsymbol{D}\boldsymbol{y} \|_2^2 }  \le 1
\end{align}
Since matrix $\boldsymbol{D}$ is not full rank, we know $\beta=1$.

\end{proof}

\section{Generalization Bound}

\begin{theorem*}
Consider  a Lipschitz continuous loss function $\ell( {z}, y) :  {\mathcal{Z}}  \times \mathcal{Y} \to [0,1]  $ with Lipschitz constant $L$ w.r.t the input ${z}$.    Let $\widetilde{\ell}({z},y):= \ell( z, y)  -  \ell( 0, y) $.  Let $\widehat{G}$ be the function class of our $(T\!\!+\!\!1)$-layer SNOK mapping from $\mathcal{X}$ to $\mathcal{Z}$. Suppose the activation function $|h(\boldsymbol{y})|\le |\boldsymbol{y}|$ (element-wise),  and the  $l_2$-norm of last layer weight is bounded, i.e.,   $\|\boldsymbol{w} \|_2 \le \mathcal{B}_w$.  Let $(\boldsymbol{x}_i, y_i)_{i=1}^N$ be i.i.d. samples drawn from $\mathcal{X} \times \mathcal{Y}$.   Let $\boldsymbol{Y}_{T\!+\!1}$ be the $T^{th}$ layer  output with input $\boldsymbol{X}$. Denote the mutual coherence of $\boldsymbol{Y}_{T\!+\!1}$ as $\mu^*$, i.e.,  $ \mu^* =  \mu(\boldsymbol{Y}_{T\!+\!1}) \le 1$.    Then, for $\forall N$ and $\forall \delta, 0<\delta<1$, with a probability at least $1-\delta$, $\forall g \in \widehat{G}$,  we have
\begin{align}
 \mathbb{E} \big[ \ell(g(X),Y) \big]     \le  \frac{1}{N}\sum_{i=1}^N \ell(g(\boldsymbol{x}_i),y_i) +  \frac{L\mathcal{B}_w \sqrt{ \big((N-1) \mu^* +1 \big) T } }{N}    \| \boldsymbol{X} \|_F  +  \sqrt{\frac{8 \ln (2/\delta)}{N}} 
\end{align}
where    $\boldsymbol{X}=[\boldsymbol{x}_1,\cdots,\boldsymbol{x}_N]$, and  $\|\cdot \|_F$ denotes the Frobenius norm.
\end{theorem*}

\begin{proof}
Plug the Rademacher complexity bound of SNOK (our Theorem~\ref{RCB}) into the Theorem 8 in~\cite{bartlett2002rademacher}, we can obtain  the bound.
\end{proof}





\section{Rademacher Complexity and Generalization Bound for A More General Structured Neural Network Family}

\textbf{Neural Network Structure:}
For a more general  structured neural network family that includes SNOK, the $1$-$T$ layers are given as 
\begin{align}
    \boldsymbol{y}_{t+1} = h( \boldsymbol{D}^\top_t\boldsymbol{x} +  (\boldsymbol{I}-\boldsymbol{D}^\top_t\boldsymbol{D}_t)\boldsymbol{y}_t   )
\end{align}
where $\boldsymbol{D}_t \in \mathcal{R}^{d_D \times d}$ are free parameters such that  $\boldsymbol{D}_t\boldsymbol{D}^\top_t = \boldsymbol{I}_d$ and $d_D >  d$, and $\boldsymbol{y}_0 =\boldsymbol{0}$. 

The  last layer ( $(T\!\!+\!\!1)^{th}$ layer) is given by ${z} = \boldsymbol{w}^\top\boldsymbol{y}_{T\!+\!1}$.  Consider a $L$-Lipschitz continuous loss function $\ell( {z}, y) :  {\mathcal{Z}}  \times \mathcal{Y} \to [0,1]  $ with Lipschitz constant $L$ w.r.t the input ${z}$.

\begin{theorem}
\label{GRCB}
(Rademacher Complexity Bound)
Consider  a Lipschitz continuous loss function $\ell( {z}, y) :  {\mathcal{Z}}  \times \mathcal{Y} \to [0,1]  $ with Lipschitz constant $L$ w.r.t the input ${z}$.    Let $\widetilde{\ell}({z},y):= \ell( z, y)  -  \ell( 0, y) $.  Let $\widehat{G}$ be the function class of the above $(T\!\!+\!\!1)$-layer structured NN mapping from $\mathcal{X}$ to $\mathcal{Z}$. Suppose the activation function $|h(\boldsymbol{y})|\le |\boldsymbol{y}|$ (element-wise),  and the  $l_2$-norm of last layer weight is bounded, i.e.,   $\|\boldsymbol{w} \|_2 \le \mathcal{B}_w$.  Let $(\boldsymbol{x}_i, y_i)_{i=1}^N$ be i.i.d. samples drawn from $\mathcal{X} \times \mathcal{Y}$.   Let $\boldsymbol{Y}_{T\!+\!1}=[\boldsymbol{y}_{T\!+\!1}^{(1)}, \cdots, \boldsymbol{y}_{T\!+\!1}^{(N)} ]$ be the $T^{th}$ layer  output with input $\boldsymbol{X}$. Denote the mutual coherence of $\boldsymbol{Y}_{T\!+\!1}$ as $\mu^*$, i.e.,  $ \mu^* =  \mu(\boldsymbol{Y}_{T\!+\!1}) =  \max _{i \ne j} \frac{\boldsymbol{y}_{T\!+\!1}^{(i)\top} \boldsymbol{y}_{T\!+\!1}^{(j)} }{ \|\boldsymbol{y}_{T\!+\!1}^{(i)}\|_2 \| \boldsymbol{y}_{T\!+\!1}^{(j)} \|_2}  \le 1$.   Then, we have
\begin{align}
    \mathfrak{R}_N(\widetilde{\ell} \circ \widehat{G} )  = \frac{1}{N} \mathbb{E} \left[ \sup _{g \in \widehat{\mathcal{G}} } \sum _{i=1} ^{N} \epsilon_i  \widetilde{\ell}( g(\boldsymbol{x}_i) , y_i )    \right] &  \le  \frac{L\mathcal{B}_w \sqrt{ T\big((N-1) \mu^* +1 \big) } }{N}  \| \boldsymbol{X} \|_F 
\end{align}
where   $\boldsymbol{X}=[\boldsymbol{x}_1,\cdots,\boldsymbol{x}_N]$.  $\| \cdot\|_2$ and   $\|\cdot \|_F$ denote the spectral norm and  the Frobenius norm of input matrix, respectively.
\end{theorem}
\textbf{Remark:}   The Rademacher complexity bound  has a complexity  $O(\sqrt{T})$ w.r.t. the depth of NN.

\begin{proof}
Since $\widetilde{\ell}$ is $L$-Lipschitz continuous function, from the composition rule of Rademacher complexity, we know that
\begin{align}
    \mathfrak{R}_N(\widetilde{\ell} \circ \widehat{G} )  \le L \; \mathfrak{R}_N(\widehat{G} ) 
\end{align}
It follows that
\begin{align}
    \mathfrak{R}_N(\widehat{G} ) & = \frac{1}{N}\mathbb{E} \left[ \sup _{g \in \widehat{\mathcal{G}} } \sum _{i=1} ^{N} \epsilon_i   g(\boldsymbol{x}_i)    \right]  \\
    & =  \frac{1}{N}\mathbb{E} \left[ \sup _{\boldsymbol{w}, \{\boldsymbol{D}_t \in \mathcal{M} \}_{t=1}^T} \sum _{i=1} ^{N}   \epsilon_i   \big<\boldsymbol{w}, \boldsymbol{y}_{T\!+\!1}^{(i)}  \big>   \right]  \\ 
    & = \frac{1}{N}\mathbb{E} \left[ \sup _{\boldsymbol{w}, \{\boldsymbol{D}_t \in \mathcal{M} \}_{t=1}^T}   \big<\boldsymbol{w}, \sum _{i=1} ^{N}   \epsilon_i \boldsymbol{y}_{T\!+\!1}^{(i)}  \big>   \right]  \\
    & \le \frac{1}{N}\mathbb{E} \left[ \sup _{\boldsymbol{w}, \{ \boldsymbol{D}_t \in \mathcal{M} \}_{t=1}^T}   \| \boldsymbol{w} \|_2  \; \big\|\sum _{i=1} ^{N}   \epsilon_i \boldsymbol{y}_{T\!+\!1}^{(i)}  \big \|_2  \right]  \;\;  \text{(Cauchy-Schwarz  inequality)} \\
    & \le \frac{\mathcal{B}_w}{N}\mathbb{E} \left[ \sup _{ \{\boldsymbol{D}_t \in \mathcal{M} \}_{t=1}^T} \big\|\sum _{i=1} ^{N}   \epsilon_i \boldsymbol{y}_{T\!+\!1}^{(i)}  \big \|_2  \right]  \\
    & = \frac{\mathcal{B}_w}{N}\mathbb{E} \left[ \sup _{ \{\boldsymbol{D}_t \in \mathcal{M} \}_{t=1}^T}  \sqrt{ \sum _{i=1} ^{N}  \big \|   \epsilon_i \boldsymbol{y}_{T\!+\!1}^{(i)}  \big\|_2^2   +  \sum _{i=1} ^{N} {\sum _{j=1,j\ne i} ^{N} {  \epsilon_i  \epsilon_j  \boldsymbol{y}_{T\!+\!1}^{(i)\top} \boldsymbol{y}_{T\!+\!1}^{(j)}   } }  }    \right] \\
    & = \frac{\mathcal{B}_w}{N}\mathbb{E} \left[ \sup _{ \{\boldsymbol{D}_t \in \mathcal{M} \}_{t=1}^T}  \sqrt{ \sum _{i=1} ^{N}  \big \|    \boldsymbol{y}_{T\!+\!1}^{(i)}  \big\|_2^2   +  \sum _{i=1} ^{N} {\sum _{j=1,j\ne i} ^{N} {  \epsilon_i  \epsilon_j  \boldsymbol{y}_{T\!+\!1}^{(i)\top} \boldsymbol{y}_{T\!+\!1}^{(j)}   } }  }    \right]  \\
    & = \frac{\mathcal{B}_w}{N}\mathbb{E} \left[   \sqrt{ \sup _{ \{\boldsymbol{D}_t \in \mathcal{M} \}_{t=1}^T} \sum _{i=1} ^{N}  \big \|    \boldsymbol{y}_{T\!+\!1}^{(i)}  \big\|_2^2   +  \sum _{i=1} ^{N} {\sum _{j=1,j\ne i} ^{N} {  \epsilon_i  \epsilon_j  \boldsymbol{y}_{T\!+\!1}^{(i)\top} \boldsymbol{y}_{T\!+\!1}^{(j)}   } }  }    \right] \\
    & \le \frac{\mathcal{B}_w}{N}   \sqrt{ \mathbb{E} \left[ \sup _{ \{\boldsymbol{D}_t \in \mathcal{M} \}_{t=1}^T} \sum _{i=1} ^{N}  \big \|    \boldsymbol{y}_{T\!+\!1}^{(i)}  \big\|_2^2   +  \sum _{i=1} ^{N} {\sum _{j=1,j\ne i} ^{N} {  \epsilon_i  \epsilon_j  \boldsymbol{y}_{T\!+\!1}^{(i)\top} \boldsymbol{y}_{T\!+\!1}^{(j)}   } } \right]  }    \label{JensenS}
\end{align}
Inequality~(\ref{JensenS}) is because of the Jensen inequality and concavity of the square root function. 

Note that $|\epsilon_i|=1, \forall i \in \{1,\cdots,N\}$,  and 
 the mutual coherence of $\boldsymbol{Y}_{T\!+\!1}$ is $\mu^*$, i.e., $ \mu^* =  \mu(\boldsymbol{Y}_{T\!+\!1}) \le 1$,   it follows that
\begin{align}
      \mathfrak{R}_N(\widehat{G} )  & \le \frac{\mathcal{B}_w}{N}   \sqrt{ \mathbb{E} \left[ \sup _{ \{\boldsymbol{D}_t \in \mathcal{M} \}_{t=1}^T} \sum _{i=1} ^{N}  \big \|    \boldsymbol{y}_{T\!+\!1}^{(i)}  \big\|_2^2   +  \sum _{i=1} ^{N} {\sum _{j=1,j\ne i} ^{N} {  \epsilon_i  \epsilon_j  \boldsymbol{y}_{T\!+\!1}^{(i)\top} \boldsymbol{y}_{T\!+\!1}^{(j)}   } } \right]  }    \\
      & \le    \frac{\mathcal{B}_w}{N}   \sqrt{ \sup _{ \{\boldsymbol{D}_t \in \mathcal{M} \}_{t=1}^T}   \sum _{i=1} ^{N}  \big \|    \boldsymbol{y}_{T\!+\!1}^{(i)}  \big\|_2^2      + \sum _{i=1} ^{N} {\sum _{j=1,j\ne i} ^{N} {  \big \|  \boldsymbol{y}_{T\!+\!1}^{(i)} \big \|_2 \big \| \boldsymbol{y}_{T\!+\!1}^{(j)} \big \|_2 \mu^*  } } }    \\
      & =   \frac{\mathcal{B}_w}{N}   \sqrt{ \sup _{ \{\boldsymbol{D}_t \in \mathcal{M} \}_{t=1}^T} (1-\mu^*) \sum _{i=1} ^{N}  \big \|    \boldsymbol{y}_{T\!+\!1}^{(i)}  \big\|_2^2   +  \mu^* \big( \sum _{i=1} ^{N}  \big \|    \boldsymbol{y}_{T\!+\!1}^{(i)}  \big\|_2 \big)^2  }  \\
       & \le    \frac{\mathcal{B}_w}{N}   \sqrt{ \sup _{ \{\boldsymbol{D}_t \in \mathcal{M} \}_{t=1}^T} (1-\mu^*) \|\boldsymbol{Y}_{T\!+\!1} \|_F^2   +  N \mu^* \|\boldsymbol{Y}_{T\!+\!1} \|_F^2  }   \; \; \text{Cauchy-Schwarz} \\
      &   \le    \frac{\mathcal{B}_w}{N}   \sqrt{ \sup _{ \{\boldsymbol{D}_t \in \mathcal{M} \}_{t=1}^T}   \big((N-1) \mu^* +1 \big) \|\boldsymbol{Y}_{T\!+\!1} \|_F^2  }    \label{supIneqS}
\end{align}
where $\boldsymbol{Y}_{T\!+\!1}=[\boldsymbol{y}_{T\!+\!1}^{(1)}, \cdots, \boldsymbol{y}_{T\!+\!1}^{(N)} ]$ and $\|\cdot \|_F$ denotes the Frobenius norm.

Since $|h(\boldsymbol{Y})| \le |\boldsymbol{Y}| $ (element-wise), (e.g., ReLU, max-pooling, soft-thresholding),   it follows that 
\begin{align}
   \|\boldsymbol{Y}_{T\!+\!1} \|_F^2 & =   \| h\big(  \boldsymbol{D}^\top_T\boldsymbol{X} +  (\boldsymbol{I}-\boldsymbol{D}^\top_T\boldsymbol{D}_T)\boldsymbol{Y}_T \big) \|_F^2  \\
    & \le  \| \boldsymbol{D}^\top_{T}\boldsymbol{X} +  (\boldsymbol{I}-\boldsymbol{D}^\top_T\boldsymbol{D}_T)\boldsymbol{Y}_{T}  \|_F^2 
\end{align}

In addition, we have
\begin{align}
   \| \boldsymbol{D}^\top_T\boldsymbol{X} +  (\boldsymbol{I}-\boldsymbol{D}^\top_T\boldsymbol{D}_T)\boldsymbol{Y}_{T}  \|_F^2 =   \| \boldsymbol{D}^\top_{T}\boldsymbol{X} \|_F^2 +   \| (\boldsymbol{I}-\boldsymbol{D}^\top_T\boldsymbol{D}_T)\boldsymbol{Y}_{T}  \|_F^2 + 2\big<\boldsymbol{D}^\top_{T}\boldsymbol{X}, (\boldsymbol{I}\!-\!\boldsymbol{D}^\top_T\boldsymbol{D}_T)\boldsymbol{Y}_{T}  \big>
\end{align}
Note that $\boldsymbol{D}_T\boldsymbol{D}^\top_T = \boldsymbol{I}_d$,   we have 
\begin{align}
    & \| \boldsymbol{D}^\top_{T}\boldsymbol{X} \|_F^2 = \| \boldsymbol{X} \|_F^2  \\
    & \big<\boldsymbol{D}^\top_{T}\boldsymbol{X}, (\boldsymbol{I}\!-\!\boldsymbol{D}^\top_T\boldsymbol{D}_T)\boldsymbol{Y}_{T}  \big> = \text{tr}\left( \boldsymbol{X}^\top_T\boldsymbol{D}_T(\boldsymbol{I}\!-\!\boldsymbol{D}^\top_T\boldsymbol{D}_T)\boldsymbol{Y}_{T}   \right) =0
\end{align}
It follows that
\begin{align}
   \| \boldsymbol{D}^\top_{T}\boldsymbol{X} +  (\boldsymbol{I}-\boldsymbol{D}^\top_T\boldsymbol{D}_T)\boldsymbol{Y}_{T}  \|_F^2 &  =   \| \boldsymbol{D}^\top_{T}\boldsymbol{X} \|_F^2 +   \| (\boldsymbol{I}-\boldsymbol{D}^\top_T\boldsymbol{D}_T)\boldsymbol{Y}_{T}  \|_F^2 + 2\big<\boldsymbol{D}^\top_{T}\boldsymbol{X}, (\boldsymbol{I}\!-\!\boldsymbol{D}^\top_T\boldsymbol{D}_T)\boldsymbol{Y}_{T}  \big> \\
   & =  \| \boldsymbol{X} \|_F^2 +  \| (\boldsymbol{I}-\boldsymbol{D}^\top_T\boldsymbol{D}_T)\boldsymbol{Y}_{T}  \|_F^2  \\
   & \le  \| \boldsymbol{X} \|_F^2  + \|\boldsymbol{I}-\boldsymbol{D}^\top_T\boldsymbol{D}_T \|_2^2  \| \boldsymbol{Y}_{T}  \|_F^2   =  \| \boldsymbol{X} \|_F^2  + \| \boldsymbol{y}_{T}  \|_F^2
\end{align}

Recursively apply the above procedure from $t=T $ to $t=1$, together with $\boldsymbol{Y}_0 = \boldsymbol{0}$,  we  can achieve that 
\begin{align}
    \|\boldsymbol{Y}_{T\!+\!1} \|_F^2 \le  T \| \boldsymbol{X} \|_F^2 
\end{align} 
Together with inequality~(\ref{supIneqS}), it follows that 
\begin{align}
  \mathfrak{R}_N(\widehat{G} ) & \le   \frac{\mathcal{B}_w}{N}   \sqrt{ \sup _{ \{\boldsymbol{D}_t \in \mathcal{M} \}_{t=1}^T}   \big((N-1) \mu^* +1 \big) \|\boldsymbol{Y}_{T\!+\!1} \|_F^2  }    \\
   & \le  \frac{\mathcal{B}_w \sqrt{ T\big((N-1) \mu^* +1 \big) } }{N}  \| \boldsymbol{X} \|_F
\end{align}
Finally, we obtain that
\begin{align}
      \mathfrak{R}_N(\widetilde{\ell} \circ \widehat{G} )  =  \frac{1}{N} \mathbb{E} \left[ \sup _{g \in \widehat{\mathcal{G}} } \sum _{i=1} ^{N} \epsilon_i  \widetilde{\ell}( g(\boldsymbol{x}_i) , y_i )    \right] &  \le  \frac{L\mathcal{B}_w \sqrt{ T \big((N-1) \mu^* +1 \big) } }{N}    \| \boldsymbol{X} \|_F 
\end{align}

\end{proof}

\begin{theorem}
Consider  a Lipschitz continuous loss function $\ell( {z}, y) :  {\mathcal{Z}}  \times \mathcal{Y} \to [0,1]  $ with Lipschitz constant $L$ w.r.t the input ${z}$.    Let $\widetilde{\ell}({z},y):= \ell( z, y)  -  \ell( 0, y) $.  Let $\widehat{G}$ be the function class of our general $(T\!\!+\!\!1)$-layer structured NN mapping from $\mathcal{X}$ to $\mathcal{Z}$. Suppose the activation function $|h(\boldsymbol{y})|\le |\boldsymbol{y}|$ (element-wise),  and the  $l_2$-norm of last layer weight is bounded, i.e.,   $\|\boldsymbol{w} \|_2 \le \mathcal{B}_w$.  Let $(\boldsymbol{x}_i, y_i)_{i=1}^N$ be i.i.d. samples drawn from $\mathcal{X} \times \mathcal{Y}$.   Let $\boldsymbol{Y}_{T\!+\!1}$ be the $T^{th}$ layer  output with input $\boldsymbol{X}$. Denote the mutual coherence of $\boldsymbol{Y}_{T\!+\!1}$ as $\mu^*$, i.e.,  $ \mu^* =  \mu(\boldsymbol{Y}_{T\!+\!1}) \le 1$.    Then, for $\forall N$ and $\forall \delta, 0<\delta<1$, with a probability at least $1-\delta$, $\forall g \in \widehat{G}$,  we have
\begin{align}
 \mathbb{E} \big[ \ell(g(X),Y) \big]     \le  \frac{1}{N}\sum_{i=1}^N \ell(g(\boldsymbol{x}_i),y_i) +  \frac{L\mathcal{B}_w \sqrt{ T \big((N-1) \mu^* +1 \big) } }{N}   \| \boldsymbol{X} \|_F  +  \sqrt{\frac{8 \ln (2/\delta)}{N}} 
\end{align}
where    $\boldsymbol{X}=[\boldsymbol{x}_1,\cdots,\boldsymbol{x}_N]$, and  $\|\cdot \|_F$ denotes the Frobenius norm.
\end{theorem}

\begin{proof}
Plug the Rademacher complexity bound of general structured NN (our Theorem~\ref{GRCB}) into the Theorem 8 in~\cite{bartlett2002rademacher}, we can obtain  the bound.
\end{proof}

\section{Experimental Results on Classification with Gaussian Input Noise and  Laplace Input Noise} \label{AppendixExp}

\begin{figure*}[t]
\centering
\subfigure[\scriptsize{DenseNet-Clean}]{
\label{DensenetClean}
\includegraphics[width=0.224\linewidth]{./results/Densenet_clean_cifar10.png}}
\subfigure[\scriptsize{DenseNet-Gaussian-0.1}]{
\label{Densenet01}
\includegraphics[width=0.224\linewidth]{./results/Densenet_cifar10_Gaussian_01.png}}
\subfigure[\scriptsize{DenseNet-Gaussian-0.2}]{
\label{Densenet02}
\includegraphics[width=0.224\linewidth]{./results/Densenet_cifar10_Gaussian_02.png}}
\subfigure[\scriptsize{DenseNet-Gaussian-0.3}]{
\label{Densenet03}
\includegraphics[width=0.224\linewidth]{./results/Densenet_cifar10_Gaussian_03.png}}
\subfigure[\scriptsize{ResNet-Clean}]{
\label{ResnetClean}
\includegraphics[width=0.224\linewidth]{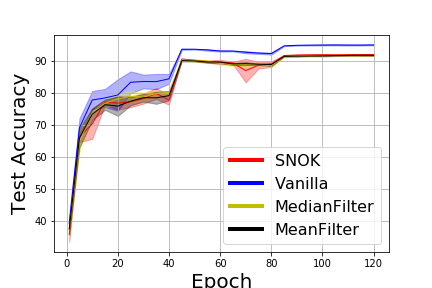}}
\subfigure[\scriptsize{ResNet-Gaussian-0.1}]{
\label{Resnet01}
\includegraphics[width=0.224\linewidth]{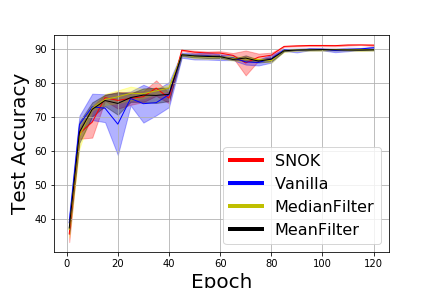}}
\subfigure[\scriptsize{ResNet-Gaussian-0.2}]{
\label{Resnet02}
\includegraphics[width=0.224\linewidth]{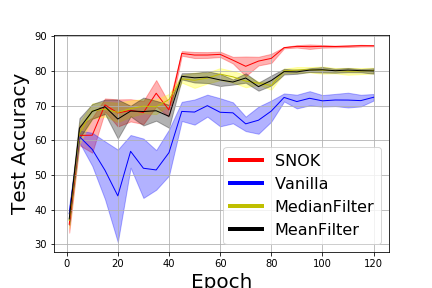}}
\subfigure[\scriptsize{ResNet-Gaussian-0.3}]{
\label{Resnet03}
\includegraphics[width=0.224\linewidth]{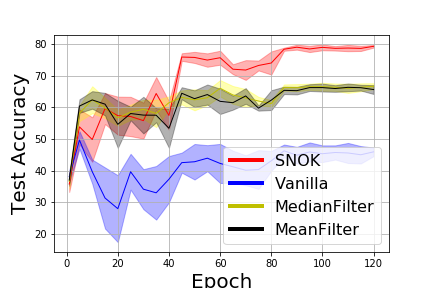}}
\caption{  Mean test accuracy $\pm$ std over 5 independent  runs on  CIFAR10 dataset with Gaussian noise  for DenseNet and ResNet backbone }
\label{CIFAR10Gaussian}
\end{figure*}

\begin{figure*}[t]
\centering
\subfigure[\scriptsize{DenseNet-Clean}]{
\label{DensenetCleanCIFAR100}
\includegraphics[width=0.224\linewidth]{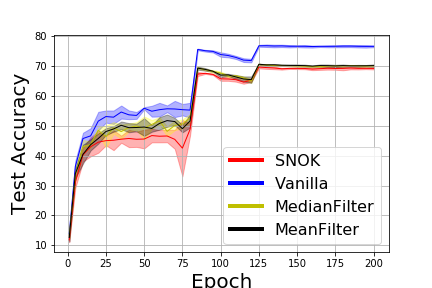}}
\subfigure[\scriptsize{DenseNet-Gaussian-0.1}]{
\label{Densenet01CIFAR100}
\includegraphics[width=0.224\linewidth]{./results/Densenet_cifar100_Gaussian_01.png}}
\subfigure[\scriptsize{DenseNet-Gaussian-0.2}]{
\label{Densenet02CIFAR100}
\includegraphics[width=0.224\linewidth]{./results/Densenet_cifar100_Gaussian_02.png}}
\subfigure[\scriptsize{DenseNet-Gaussian-0.3}]{
\label{Densenet03CIFAR100}
\includegraphics[width=0.224\linewidth]{./results/Densenet_cifar100_Gaussian_03.png}}
\subfigure[\scriptsize{ResNet-Clean}]{
\label{ResnetCleanCIFAR100}
\includegraphics[width=0.224\linewidth]{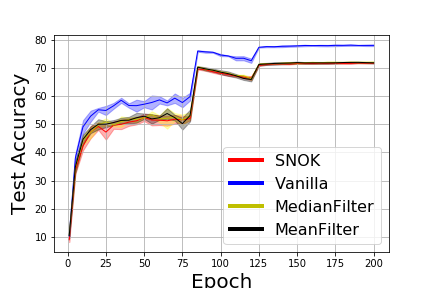}}
\subfigure[\scriptsize{ResNet-Gaussian-0.1}]{
\label{Resnet01CIFAR100}
\includegraphics[width=0.224\linewidth]{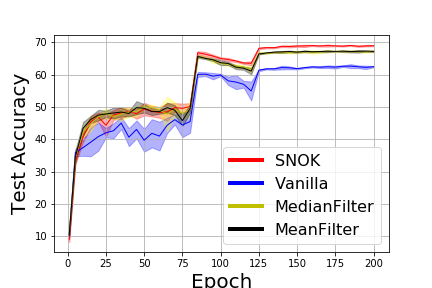}}
\subfigure[\scriptsize{ResNet-Gaussian-0.2}]{
\label{Resnet02CIFAR100}
\includegraphics[width=0.224\linewidth]{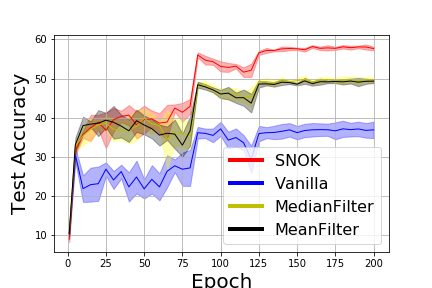}}
\subfigure[\scriptsize{ResNet-Gaussian-0.3}]{
\label{Resnet03CIFAR100}
\includegraphics[width=0.224\linewidth]{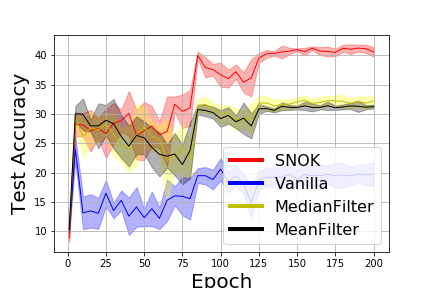}}
\caption{  Mean test accuracy $\pm$ std over 5 independent  runs on  CIFAR100 dataset with Gaussian noise for DenseNet and ResNet  backbone }
\label{CIFAR100Gaussian}
\end{figure*}

\newpage

\begin{figure*}[t]
\centering
\subfigure[\scriptsize{DenseNet-Clean}]{
\label{DensenetCleanL}
\includegraphics[width=0.224\linewidth]{./results/Densenet_clean_cifar10.png}}
\subfigure[\scriptsize{DenseNet-Laplace-0.1}]{
\label{Densenet01L}
\includegraphics[width=0.224\linewidth]{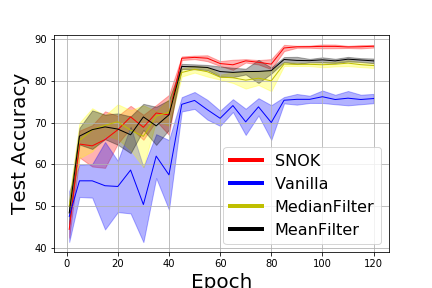}}
\subfigure[\scriptsize{DenseNet-Laplace-0.2}]{
\label{Densenet02L}
\includegraphics[width=0.224\linewidth]{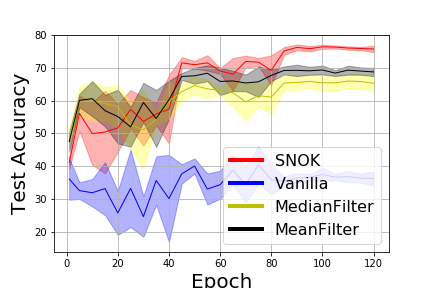}}
\subfigure[\scriptsize{DenseNet-Laplace-0.3}]{
\label{Densenet03L}
\includegraphics[width=0.224\linewidth]{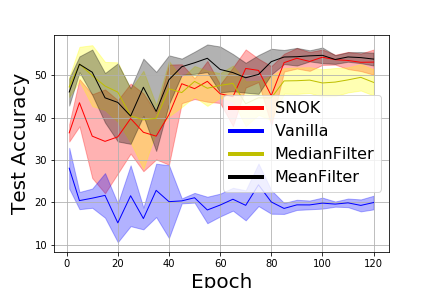}}
\subfigure[\scriptsize{ResNet-Clean}]{
\label{ResnetCleanL}
\includegraphics[width=0.224\linewidth]{./results/Resnet34_clean_cifar10.png}}
\subfigure[\scriptsize{ResNet-Laplace-0.1}]{
\label{Resnet01L}
\includegraphics[width=0.224\linewidth]{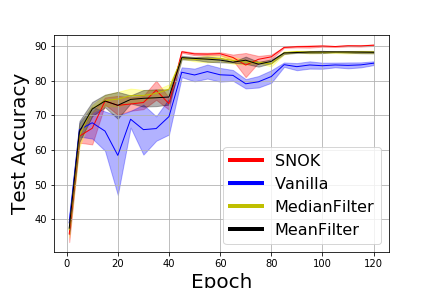}}
\subfigure[\scriptsize{ResNet-Laplace-0.2}]{
\label{Resnet02L}
\includegraphics[width=0.224\linewidth]{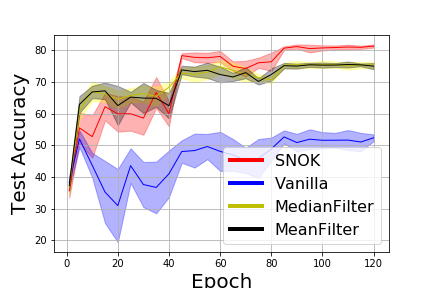}}
\subfigure[\scriptsize{ResNet-Laplace-0.3}]{
\label{Resnet03L}
\includegraphics[width=0.224\linewidth]{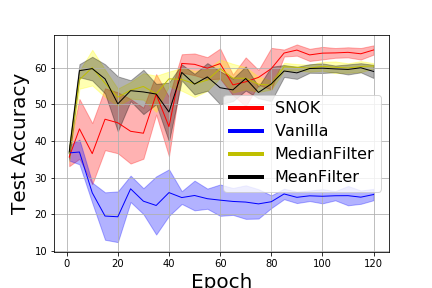}}
\caption{  Mean test accuracy $\pm$ std over 5 independent  runs on  CIFAR10 dataset with Laplace noise  for DenseNet and ResNet backbone }
\label{CIFAR10Laplace}
\end{figure*}

\begin{figure*}[t]
\centering
\subfigure[\scriptsize{DenseNet-Clean}]{
\label{DensenetCleanCIFAR100L}
\includegraphics[width=0.224\linewidth]{./results/Densenet_clean_cifar100.png}}
\subfigure[\scriptsize{DenseNet-Laplace-0.1}]{
\label{Densenet01CIFAR100L}
\includegraphics[width=0.224\linewidth]{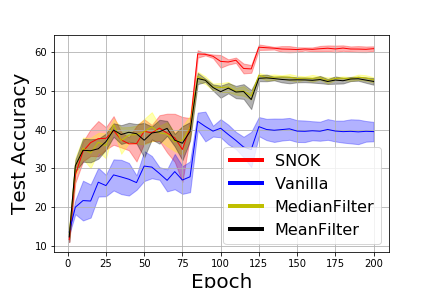}}
\subfigure[\scriptsize{DenseNet-Laplace-0.2}]{
\label{Densenet02CIFAR100L}
\includegraphics[width=0.224\linewidth]{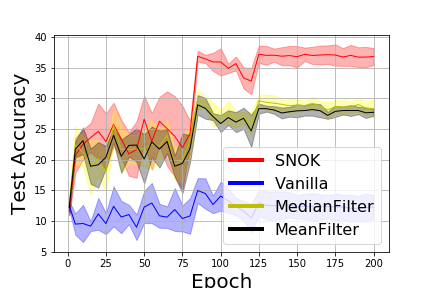}}
\subfigure[\scriptsize{DenseNet-Laplace-0.3}]{
\label{Densenet03CIFAR100L}
\includegraphics[width=0.224\linewidth]{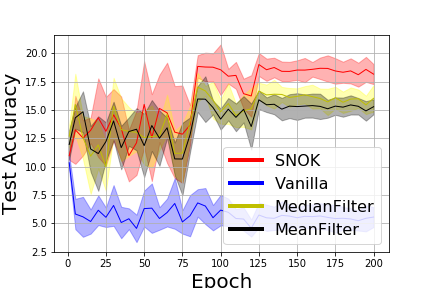}}
\subfigure[\scriptsize{ResNet-Clean}]{
\label{ResnetCleanCIFAR100L}
\includegraphics[width=0.224\linewidth]{./results/Resnet34_clean_cifar100.png}}
\subfigure[\scriptsize{ResNet-Laplace-0.1}]{
\label{Resnet01CIFAR100L}
\includegraphics[width=0.224\linewidth]{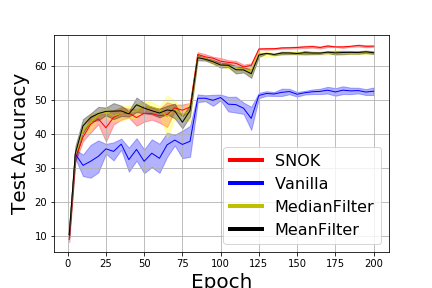}}
\subfigure[\scriptsize{ResNet-Laplace-0.2}]{
\label{Resnet02CIFAR100L}
\includegraphics[width=0.224\linewidth]{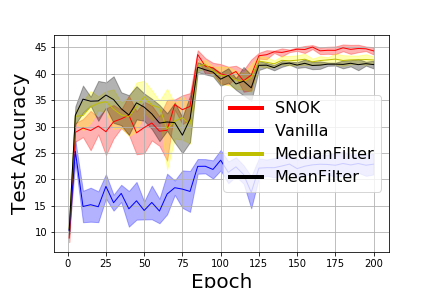}}
\subfigure[\scriptsize{ResNet-Laplace-0.3}]{
\label{Resnet03CIFAR100L}
\includegraphics[width=0.224\linewidth]{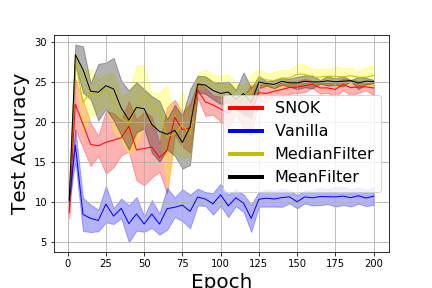}}
\caption{  Mean test accuracy $\pm$ std over 5 independent  runs on  CIFAR100 dataset with Laplace noise for DenseNet and ResNet backbone }
\label{CIFAR100Laplace}
\end{figure*}

\end{document}